%% file: main.tex
\documentclass[runningheads]{llncs}

\usepackage{eccv}
\usepackage{eccvabbrv}

\usepackage{graphicx}
\usepackage{booktabs}
\usepackage{color}
\usepackage{adjustbox}
\usepackage{caption}
\usepackage{amsmath}
\usepackage{amssymb}
\usepackage{tikz}
\usepackage{comment}
\usepackage{multicol}
\usepackage{multirow}
\usepackage{makecell}
\usepackage{subcaption}
\usepackage{microtype}
\usepackage{enumitem}
\usepackage{float}
\usepackage{svg}
\usepackage{rotating}
\usepackage[accsupp]{axessibility}
\usepackage{hyperref}
\usepackage{orcidlink}
\usepackage[capitalize]{cleveref}
\usepackage{tablefootnote}

\crefname{section}{Sec.}{Secs.}
\Crefname{section}{Section}{Sections}
\Crefname{table}{Table}{Tables}
\crefname{table}{Tab.}{Tabs.}

\newcommand{\optimizedfigspacing}{\vspace{-15pt}}
\newcommand{\OURS}{UPAL}
\newcommand{\OURSLONG}{Unified Efficient Points and Lines}
\newcommand{\fst}[1]{\textbf{#1}}
\newcommand{\snd}[1]{\underline{#1}}

\hypersetup{hidelinks}

\begin{document}

\title{Unified and Efficient Point-Line Local Features}

\author{
Fran\c{c}ois Costa\inst{1,}\thanks{Fran\c{c}ois Costa and Raphael Kreft contributed equally to this work.}
\and
Raphael Kreft\inst{1,*}
\and
Eckhard Goedeke\inst{1}
\and
Felix M\"oller\inst{1}
\and
Hardik Shah\inst{1}
\and
Ramanathan Rajaraman\inst{1}
\and
Shaohui Liu\inst{1}
\and
R\'emi Pautrat\inst{2}
\and
Marc Pollefeys\inst{1,2}
}

\authorrunning{Costa et al.}

\institute{
Department of Computer Science, ETH Zurich, Switzerland
\and
Microsoft Spatial AI Lab
}
\maketitle

\begin{abstract}
  \input{sec/0_abstract}
\end{abstract}

\input{sec/1_intro}
\input{sec/2_related_work}

\input{sec/3_method}
\input{sec/4_experiments}

\input{sec/5_conclusion}

\bibliographystyle{splncs04}
\bibliography{main}

\appendix
\input{sec/suppl}

\end{document}

%% file: sec/0_abstract.tex
Multi-view computer vision pipelines typically rely on accurate sparse keypoints and robust descriptors. While incorporating line features has shown clear benefits for matching and pose estimation, existing point-line approaches remain inefficient: they detect points and lines separately, use increasingly heavy networks, and depend on CPU-bound heuristics that hinder real-time performance.
We introduce a \OURSLONG\;(\OURS) feature extractor that jointly extracts keypoints, line segments, and feature descriptors within a single lightweight architecture. A shared backbone provides common representations that feed different branches for point and line features. Line segments are recovered through an accelerated post-processing stage, an enhanced and highly efficient variant of the LSD algorithm.
\OURS\ matches or exceeds state-of-the-art performance in both point and line applications while significantly reducing computational cost, achieving, for instance, a 4× speedup and 10× smaller memory footprint over the ALIKED + DeepLSD pipeline. Code is publicly available at
\url{https://github.com/francois141/upal}.

%% file: sec/1_intro.tex
\section{Introduction}
\label{sec:intro}

Extracting salient points and their corresponding descriptors is a fundamental step in many vision tasks involving multi-view geometry. Beyond 2D points, line segments are another type of feature that can complement points due to their prevalence in human-made structures, resulting in more informative image representations for geometric tasks. Accurately localizing line segments in images remains an active field of research due to several challenges, such as their extended size across images, their susceptibility to occlusion, and their sensitivity to perspective changes.
In spite of these challenges, recent methods combining points and lines have demonstrated state-of-the-art results in multiple fields, such as feature matching~\cite{ma_2021_wglsm,guo_2021_hdpl,pautrat_suarez_2023_gluestick,ubingazhibov_2025}, 3D reconstruction~\cite{Liu_2023_LIMAP,Liu2024RobustIS}, visual localization~\cite{ramalingam2011pose,vakhitov2016accurate,zhou2018stable,agostinho2019cvxpnpl,zhou2020fast,gao2022pose,Liu_2023_LIMAP}, relative pose estimation~\cite{Hrub2023HandbookOL}, and Simultaneous Localization and Mapping (SLAM)~\cite{zuo2017robust,pumarola2017pl,gomez2019pl,fu2020plvins,lim2022uv,pljslam,eplf-vins,Li2023IDLSID,xu2024airslam}.

Despite their effectiveness, such vision pipelines using both points and lines suffer from notable computational drawbacks. They often require two separate extractors: one for keypoint detection and description, and another for line localization.
This approach is inefficient, as the two tasks are strongly related and could share computations, since points and lines are both localized in regions with strong image gradients.
Moreover, recent network architectures tend to grow in size, making both point and line extraction computationally expensive, requiring high-end GPUs to achieve satisfactory throughput. While recent line detectors have pushed performance for feature extraction and matching, little effort has been invested in reducing the size of the networks and making them efficient. Furthermore, some state-of-the-art deep line detectors reuse classical algorithms such as LSD~\cite{Gioi2010} to extract their line segments~\cite{teplyakov2022,Pautrat_2023_DeepLSD,DEALLSD}. However, LSD is a CPU-only algorithm that contains operations that could be implemented more efficiently on GPU. In addition, it suffers from quadratic complexity with respect to the size of the image.

\input{figures/new_teaser}

Motivated by these observations, we introduce \emph{\OURSLONG\ (\OURS)} local features, a novel lightweight feature extractor that extracts keypoints, line segments, and feature descriptors jointly in a single forward pass. \OURS~leverages the architecture of a point feature extractor and adds a lightweight line detection branch on top of it. Without resorting to costly learned line descriptors or matchers~\cite{syoon_2021_linetr,pautrat_suarez_2023_gluestick,ubingazhibov_2025}, we argue that lines can be efficiently matched by first matching their endpoints and then finding the optimal assignment, thus requiring no extra line descriptors.
While we are not the first to propose joint point-line feature extraction~\cite{xu2024airslam,ferre2025wireframe}, previous attempts failed to reach state-of-the-art results for both features, despite maintaining high throughput.
With \OURS, we show how to reach such goals with a streamlined design.
Through distillation, \OURS~combines the best of both worlds: it leverages an efficient and lightweight neural network while being supervised by powerful state-of-the-art models. We carefully reviewed the best point and line extractors to date, and selected the best-performing combination to distill their knowledge into our architecture. By eliminating redundant computations in the extraction of the two kinds of features (points and lines), \OURS~preserves state-of-the-art performance while improving efficiency and memory footprint. In particular, we demonstrate that \textbf{state-of-the-art results in line detection can be achieved by adding only three convolutional layers to an existing point extractor.}

Inspired by DeepLSD~\cite{Pautrat_2023_DeepLSD}, we leverage a post-processing step based on the LSD~\cite{Gioi2010} algorithm to extract the final line segments. We introduce three major improvements over the original LSD and DeepLSD implementations. First, we show that discarding the branch predicting an angle field can simplify the architecture and at the same time improve the performance. Second, we move most of LSD’s image processing to the GPU and leave only the line-growing stage on CPU. Lastly, we propose a simple but effective heuristic to reduce the number of seed points in LSD, increasing the efficiency of the detector without any loss in performance.

We evaluate \OURS~on both low-level metrics and geometric downstream tasks, demonstrating its effectiveness across multiple challenging datasets. Combined with our optimized line extraction module, \OURS~achieves a 4$\times$ speedup over the current state of the art, with similar or better performance. Moreover, approaches with comparable performance have a memory footprint that is at least one order of magnitude higher.

In summary, our contributions are as follows:
\begin{enumerate}
\item We demonstrate how to distill state-of-the-art point and line detectors and descriptors into a single lightweight network, preserving, or even improving, their original accuracy. This design substantially simplifies existing pipelines, requiring only a few additional convolutions for line extraction.
\item We introduce an enhanced variant of LSD that provides significantly higher efficiency through GPU-accelerated processing.
\item \OURS~achieves state-of-the-art performance for both point and line features, across low-level metrics and downstream tasks, while running much faster than most prior approaches and being significantly lighter, making it more suitable for embedded applications.
\end{enumerate}

%% file: figures/new_teaser.tex
\begin{figure}[tb]
    \centering
    \includegraphics[width=0.98\linewidth]{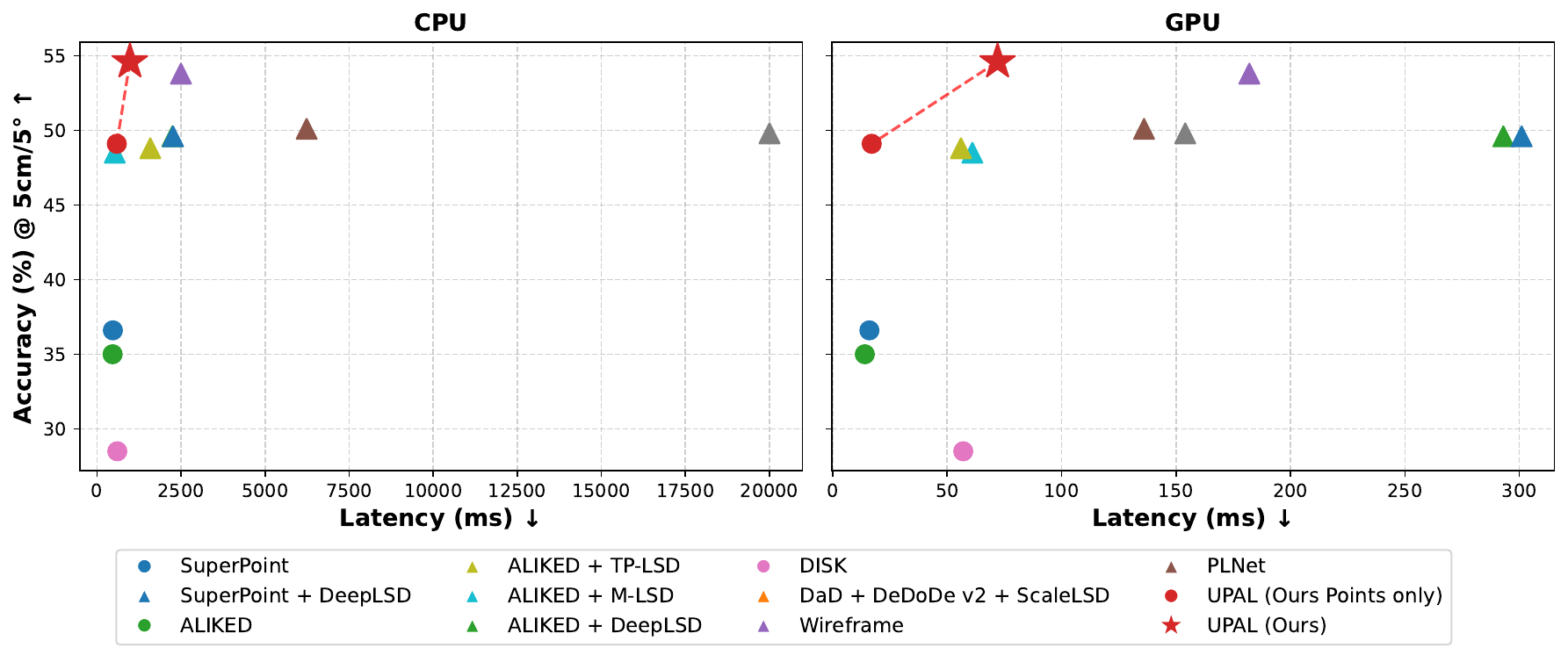}
    \caption{\textbf{In this paper, we propose \OURS, a joint point-line feature extractor} that outperforms most existing methods (here, on hybrid point-line visual localization with pose accuracy under 5cm / 5$^\circ$) \textbf{while being substantially more lightweight}. This is enabled by our compact architecture, joint point-line training with distilled supervision, and an efficient variant of the LSD algorithm. Results are shown on the \textit{Stairs} scene of the 7Scenes dataset~\cite{7scenes} with the setup described in \cref{sec:efficiency}.}
    \label{fig:teaser}
    \optimizedfigspacing
\end{figure}

%% file: sec/2_related_work.tex
\section{Related work}
\label{sec:related_work}

\textbf{Point Feature Extractors.}
Feature points are traditionally detected and described using patterns in the image gradient~\cite{Harris1988,lowe_2004,bay2008,orb}.
The deep learning era has subsequently introduced novel methods to jointly detect and describe feature points within a single network.
The pioneering work SuperPoint~\cite{DeTone2018} introduced the process of homography adaptation to create a pseudo ground truth of corner-like keypoints.
D2-Net~\cite{Dusmanu2019CVPR} showed how to describe first, then extract keypoints from the features.
R2D2~\cite{revaud2019} introduced the concept of reliability to only keep the most relevant keypoints.
DISK~\cite{tyszkiewicz2020disk} revisited how to learn reliable keypoints through reinforcement learning.
ALIKED~\cite{Zhao2023} was proposed as a lightweight network and remains among the top performers, which motivated our decision to follow the same architecture.
SiLK~\cite{silk} demonstrated that a simple setup can already lead to a strong performance.
Progress has also been made in terms of efficiency with XFeat~\cite{potje2024cvpr}, which made it possible to run feature extraction even on embedded devices.
The DeDoDe work~\cite{edstedt2024dedode} advocates for using independent networks for detection and description to maximize performance.
Finally, a recent keypoint detector, DaD~\cite{edstedt2025dad}, revisited the reinforcement learning objective to learn more robust points.
In this work, we reuse the architecture of ALIKED including its sparse descriptors branch, and leverage both the SuperPoint and DaD keypoints to supervise our keypoint detector.

\input{figures/overview}

~\\
\noindent
\textbf{Line Feature Extractors.}
Line features have recently emerged as an alternative to point features. As with points, traditional line detectors rely on image gradients to group pixels with strong, coherent gradient magnitude and orientation~\cite{Gioi2010,Akinlar2011,salaun_2016,Elder2020,zhang2021,suarez2022elsed}. With the rise of deep learning, wireframe extraction, detecting interconnected lines that capture the main structure of indoor scenes, has become a popular task~\cite{Huang2018}, leading to numerous wireframe-detection networks~\cite{Huang2018,Zhou2019,zhang2019ppgnet,afm,hawp,Meng2020,HAWP-journal}. In parallel, more general deep line detectors have been proposed to detect all kinds of lines, not only structural ones~\cite{Cai2019,huang2020tplsd,Xu2021,Gu2021,Pautrat_2023_DeepLSD,teplyakov2022,ScaleLSD,DEALLSD}. Following a trend similar to point features, joint line detection and description has also been explored~\cite{Pautrat_Lin_2021_CVPR,Zhang_2021_ICCV,Abdelbaki2022}. Among these methods, DeepLSD~\cite{Pautrat_2023_DeepLSD} and ScaleLSD~\cite{ScaleLSD} stand out for geometric tasks: the former combines a learned image gradient with LSD~\cite{Gioi2010}, while the latter scales network capacity and training data. In this work, we follow the DeepLSD design by predicting a deep image gradient and using LSD to extract segments, but introduce multiple performance and efficiency improvements to both the learned and handcrafted components.

~\\
\noindent
\textbf{Joint Point-Line Applications.}
Points and lines have long been shown to be complementary across a wide range of applications, consistently outperforming approaches relying on points or lines alone.
PLNet~\cite{xu2024airslam} and Wireframe~\cite{ferre2025wireframe} pioneered the joint detection and description of points and lines within a single network. However, this joint prediction came at the cost of lower performance compared to individual best-in-class baselines.
Joint point-line matching is a key component in many multi-view pipelines and has recently been addressed using graph neural networks~\cite{ma_2021_wglsm,guo_2021_hdpl,pautrat_suarez_2023_gluestick,ubingazhibov_2025}.
This complementarity is particularly valuable in SLAM, where the longer spatial extent of lines improves feature tracking and reduces drift~\cite{zuo2017robust,pumarola2017pl,gomez2019pl,fu2020plvins,lim2022uv,pljslam,eplf-vins,Li2023IDLSID,xu2024airslam}.
Lines also provide additional geometric structure beneficial for 3D reconstruction, and recent work has shown that combining points and lines improves both accuracy and completeness~\cite{Liu_2023_LIMAP,Liu2024RobustIS}.
Although lines alone do not fully constrain relative pose between two views, they offer useful cues through endpoints and vanishing points, effectively complementing points~\cite{Hrub2023HandbookOL}.
Finally, point-line combinations have been widely explored for visual localization~\cite{ramalingam2011pose,vakhitov2016accurate,zhou2018stable,agostinho2019cvxpnpl,zhou2020fast,gao2022pose}, where lines are advantageous in low-texture scenes, an established failure mode for point-based methods.

%% file: figures/overview.tex
\begin{figure*}[tb]
    \centering
    \includegraphics[width=\linewidth, height=192pt]{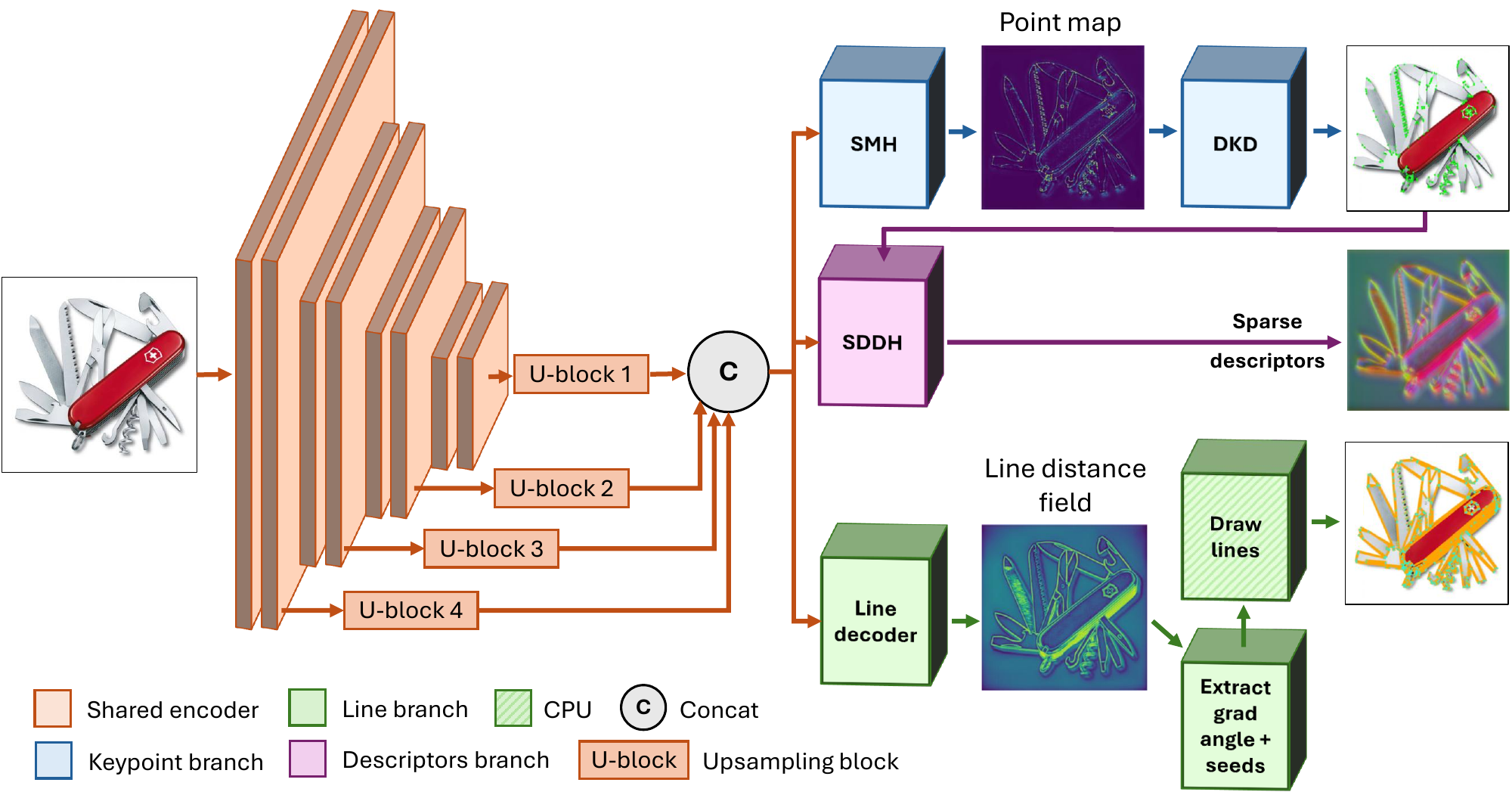}
    \caption{\textbf{Overview of our network architecture.} We use a shared convolutional encoder to generate a multi-scale feature map (orange). We then use three decoder branches: a keypoint detection branch (blue) with a Score Map Head (SMH) followed by Differentiable Keypoint Detection (DKD), a Sparse Deformable Descriptor Head (SDDH) to obtain descriptors (purple), and a branch predicting a distance field (green). The latter is then passed to our fast LSD to extract the line segments.}
    \label{fig:architecture}
    \optimizedfigspacing
\end{figure*}

%% file: sec/3_method.tex
\section{Method}

We present a unified pipeline that jointly predicts points, lines, and feature descriptors using a single neural network, reducing inference time and memory footprint while preserving or exceeding the performance of separate detectors. The model is trained in a student-teacher framework that distills multiple state-of-the-art keypoint and line detectors. An overview of the architecture is provided in \cref{fig:architecture}.

\noindent
\textbf{Design Choices. }
Our objective is to match the performance of independent feature extractors while distilling their capabilities into a single lightweight network for faster inference and a lower memory footprint. As dense matching remains computationally costly, we focus on the classical paradigm of detecting and describing features in a single image. We extensively evaluated point detectors, line detectors, and feature descriptors in isolation and identified the following strong performers among existing methods: SuperPoint~\cite{DeTone2018} and DaD~\cite{edstedt2025dad} for keypoints, ALIKED~\cite{Zhao2023} for descriptors, and DeepLSD~\cite{Pautrat_2023_DeepLSD} for line detection. To unify all methods, we adopt the architecture of ALIKED~\cite{Zhao2023} for its low latency and memory footprint and extend it with additional heads for the new feature types.

\subsection{Encoder}

The encoder follows ALIKED~\cite{Zhao2023} and consists of four convolutional blocks with progressively reduced resolution; the last two employ Deformable Convolutions (DCN)~\cite{dcn} to improve geometric invariance. Each block’s output is passed through a $1 \times 1$ convolution, then upsampled to the original resolution and finally concatenated with the other blocks' outputs to form the bottleneck $\mathbf{F}$. This representation captures both low- and high-level cues and supports accurate point detection, descriptor computation, and distance-field prediction. Figure~\ref{fig:shared_backbone} illustrates a keypoint score map, a line distance field, and a PCA visualization of $\mathbf{F}$, showing that the bottleneck can encode information for both points and lines, as keypoints frequently lie on line segments.

\input{figures/heatmap_visualization}

\subsection{Point Extraction}

Starting from the bottleneck features $\mathbf{F}$, we use the Score Map Head (SMH) from ALIKED~\cite{Zhao2023}, a lightweight convolutional head, to extract a keypoint probability heatmap $\mathbf{S} \in [0, 1]^{H \times W}$. This map is then fed into a Differentiable Keypoint Detection (DKD) module~\cite{Zhao2022ALIKE}, which applies Non-Maximum Suppression (NMS) thresholding to obtain a set of N keypoints $\{\mathbf{p_i}\}, i \in \{1, \dots, N\}$, and softargmax within a $3 \times 3$ window to obtain subpixel-accurate coordinates.

As in ALIKED~\cite{Zhao2023}, we use the Sparse Deformable Descriptor Head (SDDH). For each predicted keypoint $\mathbf{p_i}$, a $K \times K$ window is extracted around $p_i$ from the bottleneck $\mathbf{F}$, and a small module predicts M sampling positions. These sampling positions are then used to sample the bottleneck features with bilinear interpolation, followed by a $1 \times 1$ convolution and SELU activation. The sampled features are then aggregated into a single descriptor $\mathbf{d_i} \in \mathcal{R}^D$ through a weighted sum with learned weights. This produces a deformable descriptor that can incorporate information from various parts of the image and thus easily adapt to geometric deformations.

\subsection{Distance Field Prediction}

State-of-the-art line detectors such as DeepLSD~\cite{Pautrat_2023_DeepLSD} and ScaleLSD~\cite{ScaleLSD} predict an attraction field before extracting line segments from it. DeepLSD regresses in particular both a distance and an angle field, corresponding to dense maps where each pixel value encodes the distance to the nearest line and the orientation of that line. While experimenting with DeepLSD, we discovered that the angle field predicted by DeepLSD is detrimental, as will be shown in Table \ref{tab:ablation}. Instead of predicting an angle field, we observed that directly using image-gradient orientation yielded better performance, while reducing the complexity of the network. Thus, the line decoder branch of our architecture only predicts a line distance field and we compute the angle field using a GPU-based image gradient angle computation.

Furthermore, DeepLSD relies on a deep and computationally expensive U-Net backbone, which contains roughly 10× more learnable parameters than ALIKED. Since predicting a distance field is a relatively low-level task, we argue and demonstrate that a smaller-scale backbone is sufficient to produce equivalent results. In \cref{sec:experiments} we show that our lighter backbone network, followed by the same line decoder branch as in DeepLSD, can obtain results similar to or better than those of DeepLSD, while reducing the number of learnable parameters by a factor of 10. This line decoder consists only of 3 convolutions with batch normalization and ReLU activation, making it extremely lightweight.
Overall, this makes the network architecture very lightweight and efficient.

\subsection{Line Extraction}

Similar to DeepLSD~\cite{Pautrat_2023_DeepLSD}, our network predicts the distance field and delegates line extraction to the handcrafted LSD detector~\cite{Gioi2010}. LSD is an expensive, two-stage, CPU-only algorithm. It first computes the image gradient (or receives it from a learned detector), and performs an approximate sorting of the gradient magnitude of the pixels. Then, it iterates over the nearly sorted points to start drawing lines, starting from the most promising ones with the highest gradient magnitudes. This sorting-and-seeding step is expensive on CPU and the current LSD implementation has a quadratic cost with respect to the image side lengths, as it is proportional to the number of pixels in the image. 

In this work, we propose an accelerated variant of LSD, which removes the expensive approximate sorting by moving the extraction of useful seed points directly to the GPU. Following the observation that most points with low gradient values never contribute to lines and can therefore be discarded, we reduce the number of seed points. We empirically determined that downsampling the grid of seed points with a stride of 2 and keeping the top 20\% of pixels with the lowest distance-field values has no impact on the performance of the detector, while improving the efficiency by a factor of 3.

\subsection{Losses}

The keypoint heatmap \( \mathbf{S} \) is supervised using the teacher prediction \( \mathbf{\hat{S}} \) with a weighted binary cross-entropy loss:
\begin{equation}
    \mathcal{L}_{\text{wbce}}(\mathbf{S}, \mathbf{\hat{S}}) 
    = -\lambda\, \mathbf{\hat{S}}\,\log(\mathbf{S})\;-\;(1 - \mathbf{\hat{S}}) \log(1 - \mathbf{S}).
\end{equation}
The weighting factor \( \lambda \) compensates for the imbalance between positive keypoint pixels and the predominantly near-zero background values in the pseudo-ground truth. Motivated by our empirical analysis of keypoint detectors, we construct the teacher heatmap by taking the element-wise maximum of the predictions from SuperPoint~\cite{DeTone2018} and DaD~\cite{edstedt2025dad}.

The line distance field \( \mathbf{D} \) is supervised with an L1 loss against the ground-truth map \( \mathbf{\hat{D}} \) generated by DeepLSD~\cite{Pautrat_2023_DeepLSD}. Following DeepLSD, we restrict this loss to pixels within a radius \( r = 5 \) of each ground-truth line, which improves convergence and accuracy by focusing the supervision on the most informative regions. We also normalize both the prediction and the ground truth to yield more meaningful gradients in the vicinity of lines:
\begin{equation}
    \mathcal{L}_{D} = \left\lVert \mathbf{\hat{D_n}} - \mathbf{D_n} \right\rVert,
    \qquad
    \mathbf{D_n} = -\log\!\left(\frac{\mathbf{D}}{r}\right).
\end{equation}

Local descriptors \( \mathbf{d_i} \) are supervised using ground-truth descriptors \( \mathbf{\hat{d_i}} \) with an L1 loss averaged over the top \( n = 1000 \) detected keypoints:
\begin{equation}
    \mathcal{L}_{\text{desc}} = \frac{1}{n} \sum_{i=1}^{n} \left\lVert \mathbf{\hat{d_i}} - \mathbf{d_i} \right\rVert.
\end{equation}
The ground-truth descriptors are obtained by querying ALIKED-n32~\cite{Zhao2023} at the predicted keypoint locations.

The final training objective is the sum of all components:
\begin{equation}
    \mathcal{L} = \mathcal{L}_{\text{wbce}} + \mathcal{L}_{D} + \mathcal{L}_{\text{desc}}.
\end{equation}

\input{figures/detection_examples}

\subsection{Implementation Details}
Our encoder and keypoint branch follow the ALIKED-n16 architecture~\cite{Zhao2023}, while the distance-field branch is adapted from DeepLSD~\cite{Pautrat_2023_DeepLSD}. We train on 10k distractor images from the Oxford-Paris dataset~\cite{radenovic2018} at a resolution of $800 \times 800$. This dataset contains a mix of indoor and outdoor scenes, making our model suitable for both conditions. We set $K=3$ in the SDDH, use 128-dimensional descriptors, apply a keypoint loss weight of $\lambda = 200$, and optimize with Adam at a learning rate of $1\mathrm{e}{-4}$.

Training proceeds in two stages: we first learn the distance-field branch for about one hour, using early stopping after a single epoch, then freeze the encoder and distance-field module and train the remaining components for 60 epochs (about 12 hours). ALIKED-derived components are initialized with their pretrained weights. Training is performed on four NVIDIA TITAN GPUs (24 GB each) with a batch size of 4 per GPU.

%% file: figures/heatmap_visualization.tex
\begin{figure}[tb]
    \centering
    \scriptsize
    \setlength{\tabcolsep}{2pt}
    \begin{tabular}{cccc}
        \includegraphics[width=0.24\columnwidth]{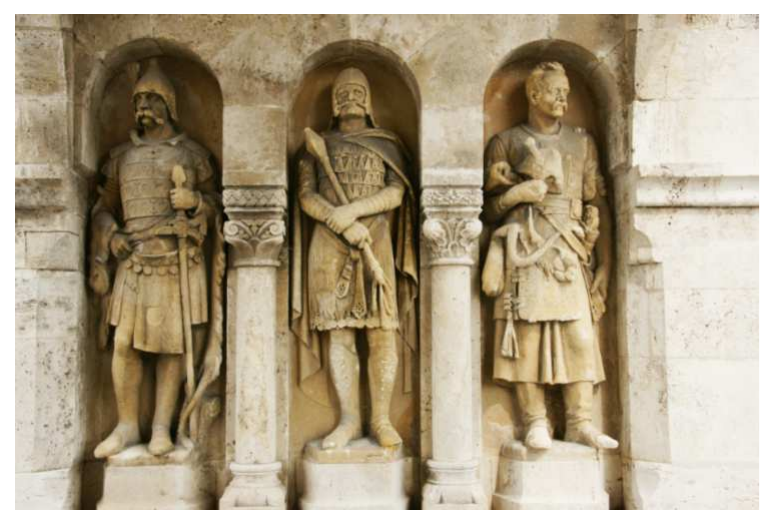} &
        \includegraphics[width=0.24\columnwidth]{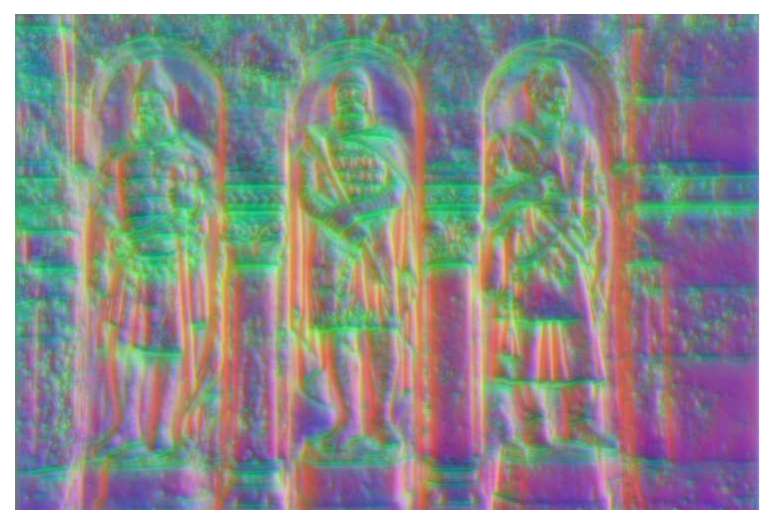} &
        \includegraphics[width=0.24\columnwidth]{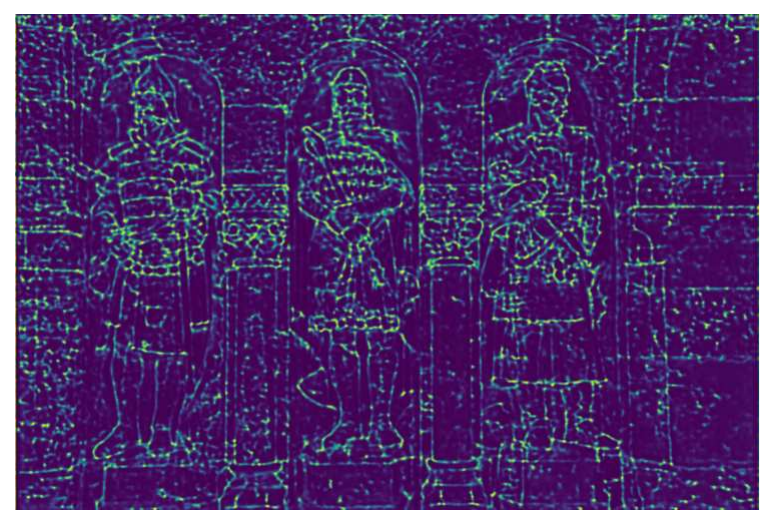} &
        \includegraphics[width=0.24\columnwidth]{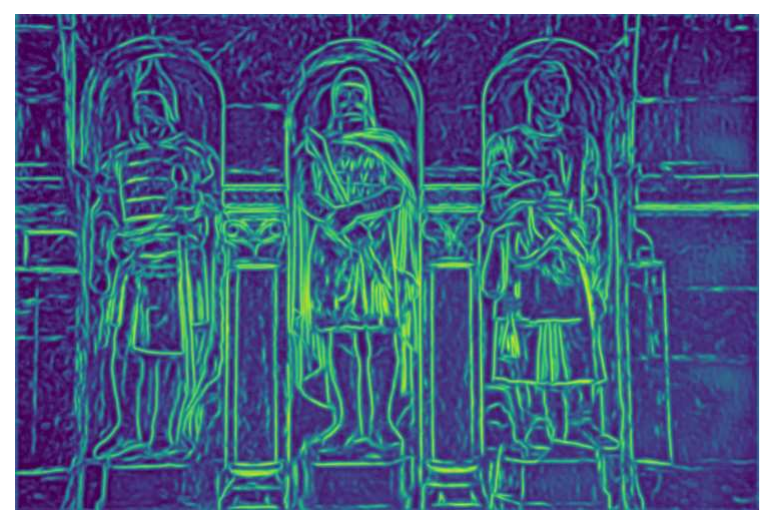} \\
        (a) Input image & (b) Bottleneck features &
        (c) Keypoint score map & (d) Line distance field \\
    \end{tabular}
    \caption{\textbf{Visualization of the outputs of the network.} The shared backbone receives an input image (a) and encodes it into bottleneck features (b). We observe that the predicted keypoint score map (c) and line distance field (d) show some similarity (e.g. keypoints are often line endpoints or intersections), justifying their joint prediction.}
    \label{fig:shared_backbone}
    \optimizedfigspacing
\end{figure}

%% file: figures/detection_examples.tex
\begin{figure*}[tb]
    \centering
    \scriptsize
    \setlength{\tabcolsep}{2pt}
    \begin{tabular}{ccccc}
        LSD~\cite{Gioi2010} & ScaleLSD~\cite{ScaleLSD} & Wireframe~\cite{ferre2025wireframe} & DeepLSD~\cite{Pautrat_2023_DeepLSD} & \OURS \ (Ours) \\
        \includegraphics[width=0.19\linewidth]{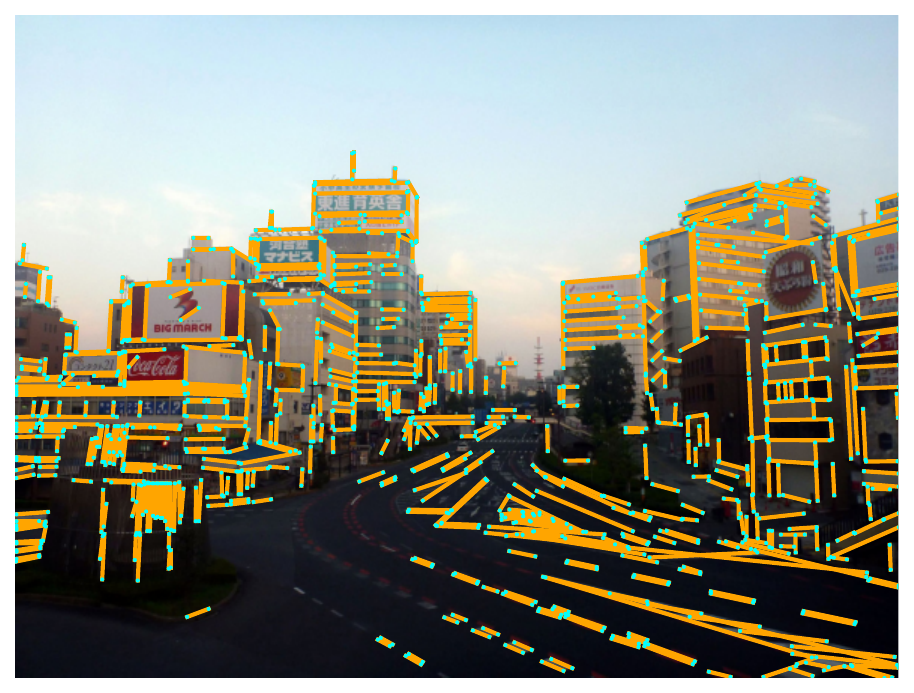} &
        \includegraphics[width=0.19\linewidth]{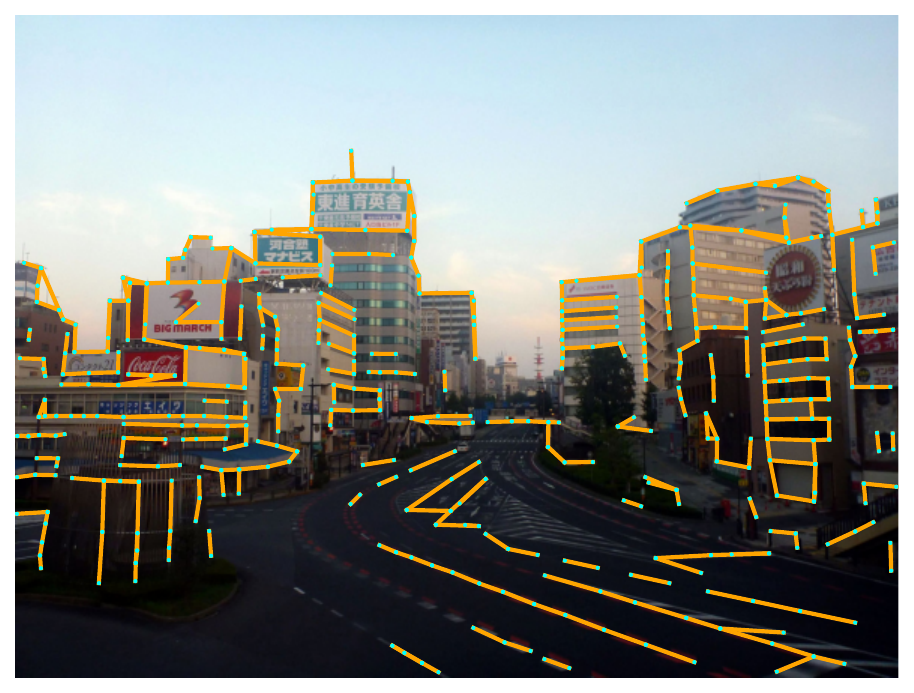} &
        \includegraphics[width=0.19\linewidth]{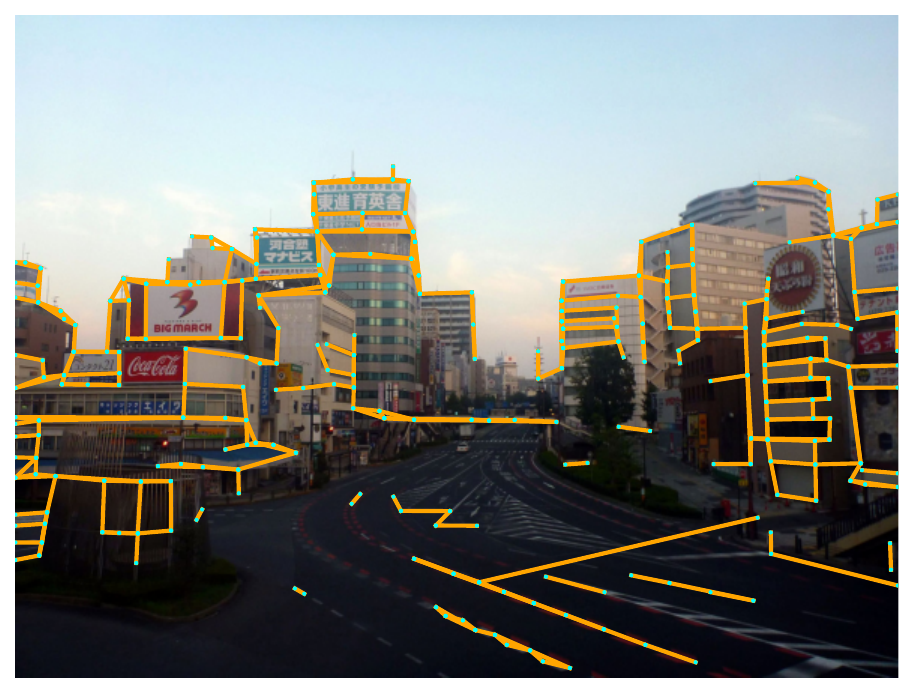} &
        \includegraphics[width=0.19\linewidth]{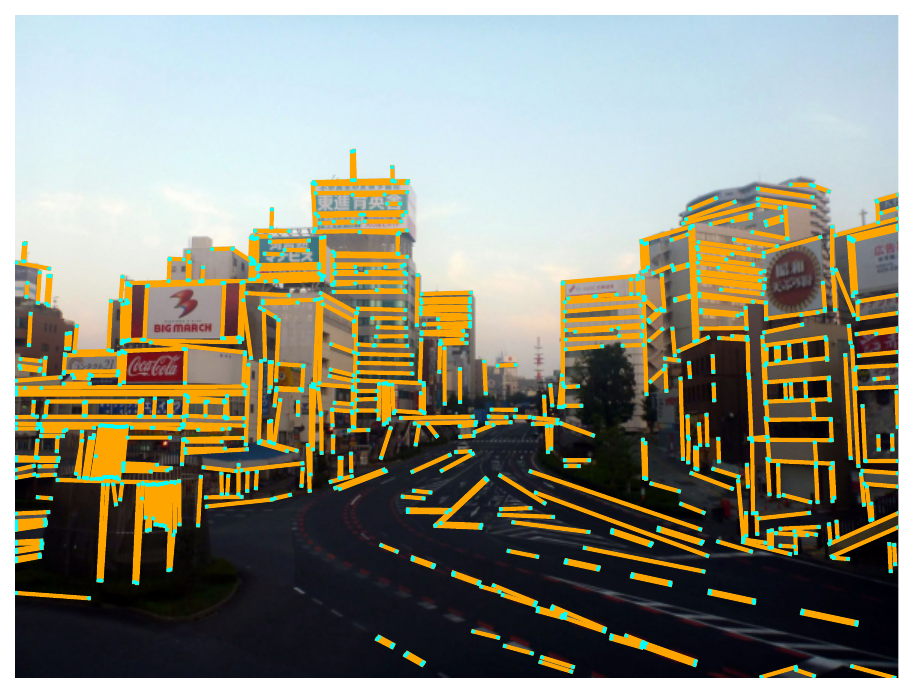} &
        \includegraphics[width=0.19\linewidth]{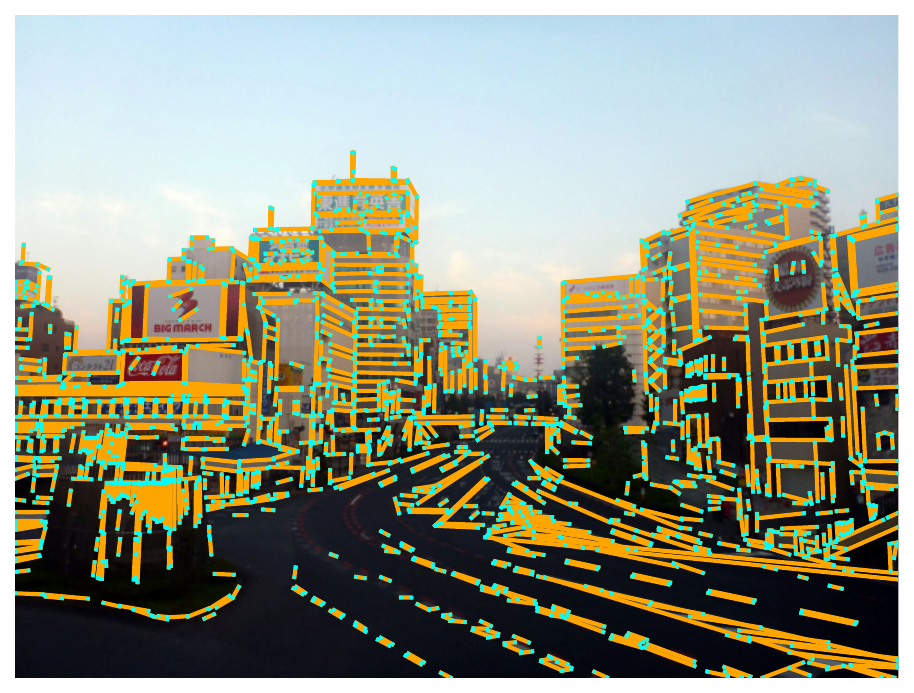}
        \\
        \includegraphics[width=0.19\linewidth]{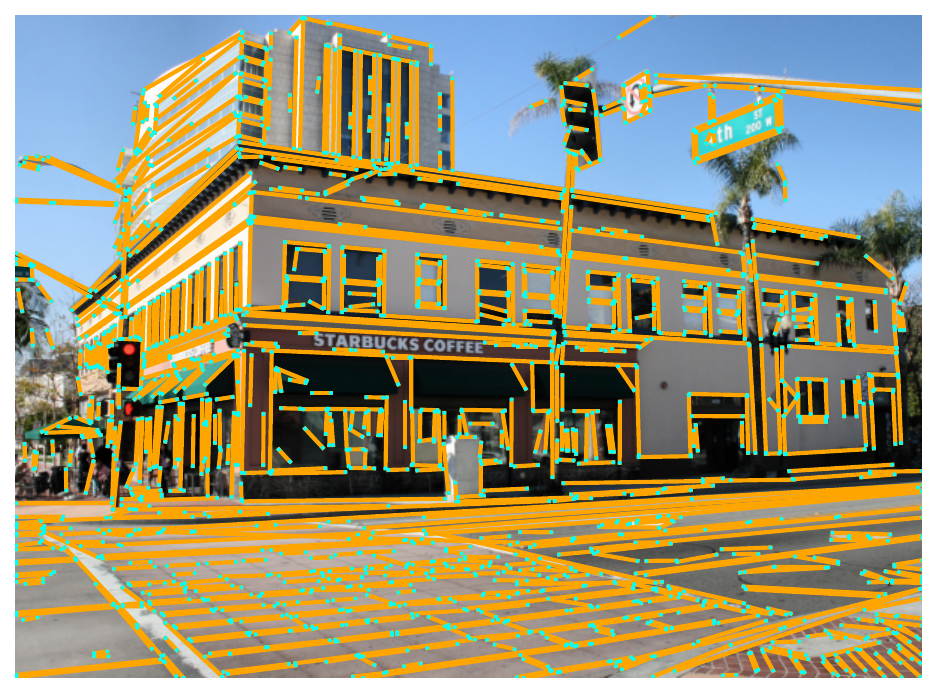} &
        \includegraphics[width=0.19\linewidth]{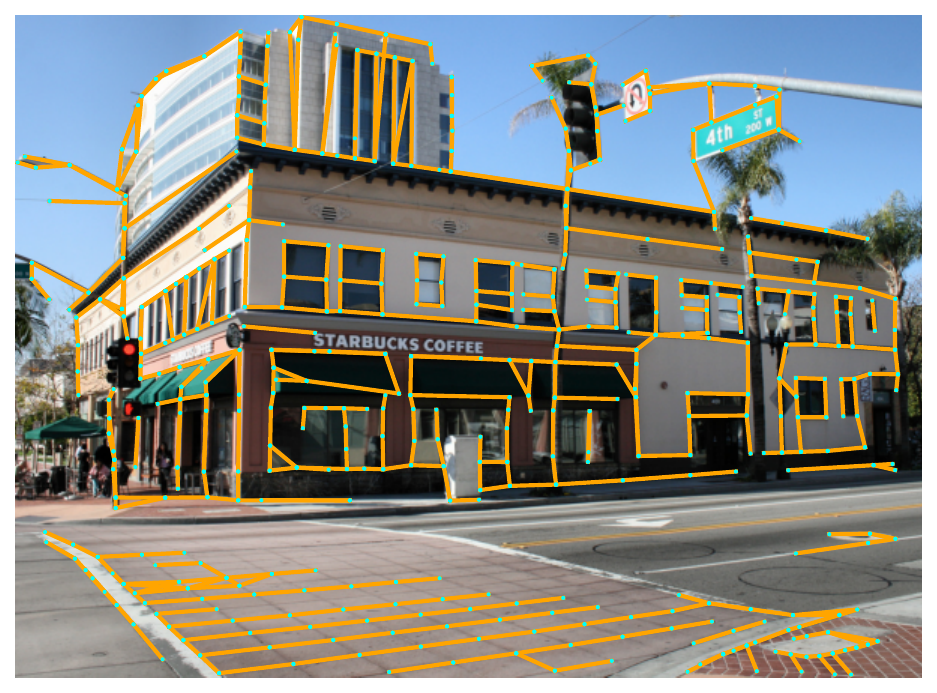} &
        \includegraphics[width=0.19\linewidth]{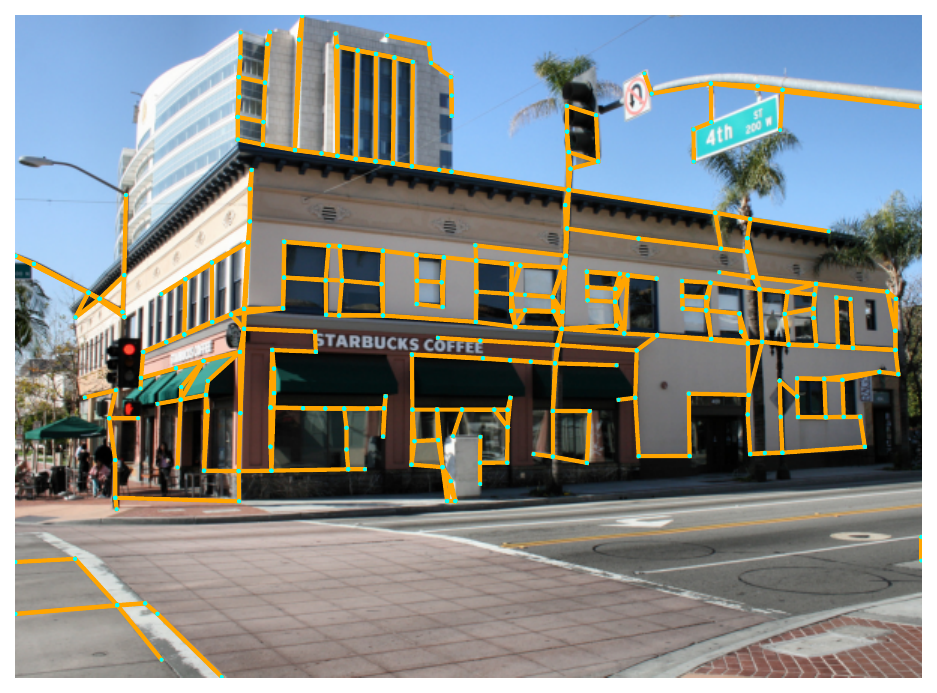} &
        \includegraphics[width=0.19\linewidth]{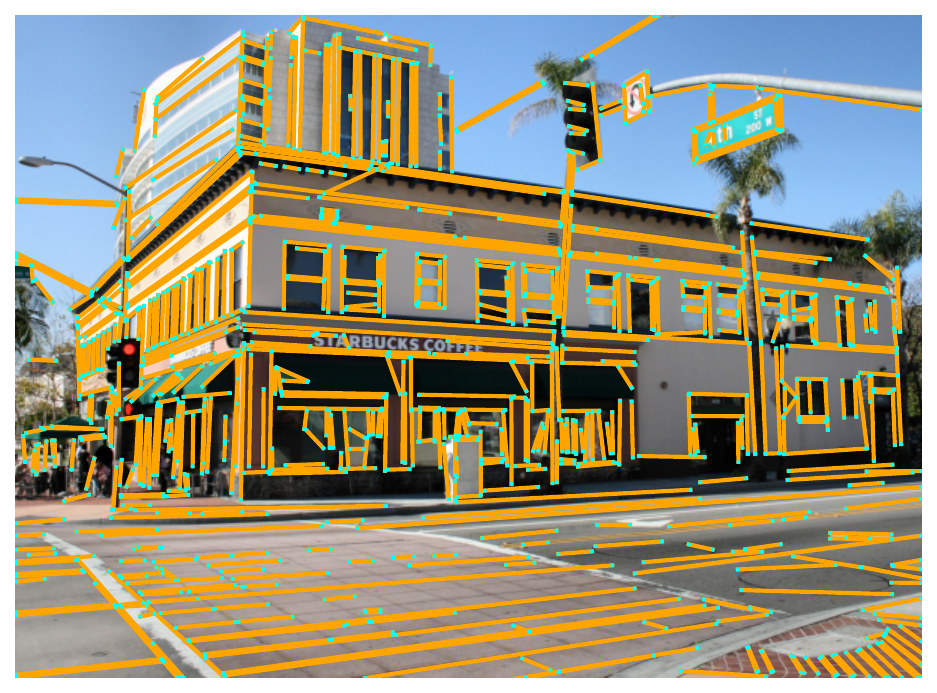} &
        \includegraphics[width=0.19\linewidth]{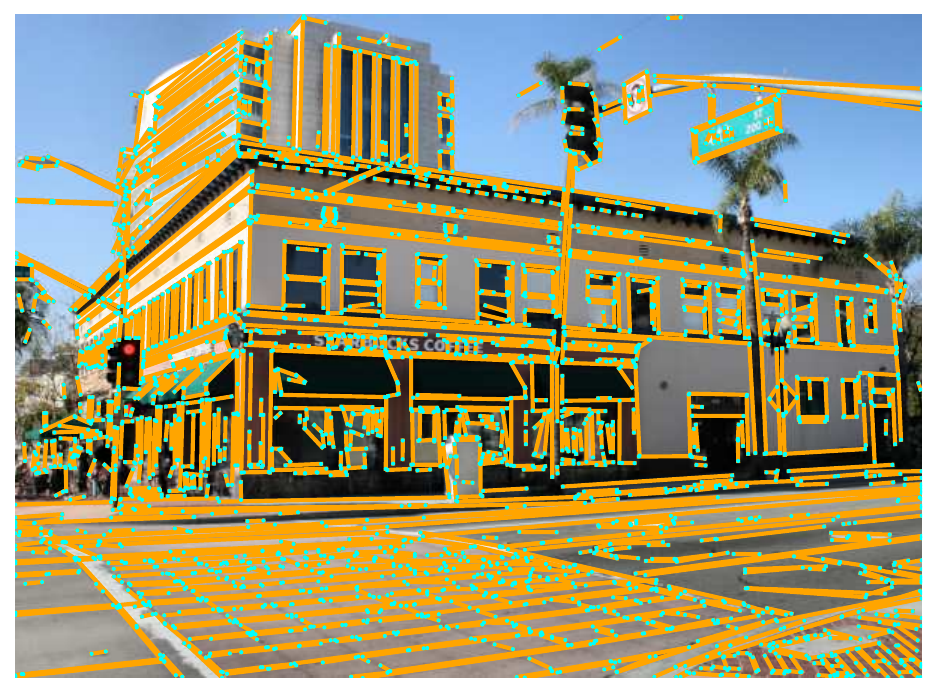}
    \end{tabular}
    \caption{\textbf{Line detection examples.} In spite of its efficiency, \OURS~outputs generic lines with quality similar to more expensive methods.}
    \label{fig:example_detection}
    \optimizedfigspacing
\end{figure*}

%% file: sec/4_experiments.tex
% =============================
% HPatches (sorted)
% =============================

\input{tables/results_point_hpatches}

\section{Experiments}
\label{sec:experiments}
We evaluate \OURS, first showing that it matches or exceeds the state of the art in point features, line segment detection, and point-line applications. We then highlight its efficiency benefits in terms of inference speed and memory footprint and provide an ablation study of our design choices. A visual comparison of predicted lines is shown in Figure~\ref{fig:example_detection}.

%------------------------------------------------------------------------------
\subsection{Point Evaluation}
%------------------------------------------------------------------------------

We begin by assessing \OURS\ against state-of-the-art point extractors, verifying that adding the line-detection branch does not degrade point performance. We use three benchmarks that evaluate point matching, jointly measuring keypoint and descriptor quality. For each benchmark, point features are extracted from image pairs and matched via mutual nearest neighbors. We compare \OURS\ to the following lightweight methods: SIFT~\cite{lowe_2004}, SuperPoint~\cite{DeTone2018}, DISK~\cite{tyszkiewicz2020disk}, ALIKED-n16~\cite{Zhao2023} (sharing the same architecture as ours), XFeat~\cite{potje2024cvpr}, the Wireframe point detector and descriptor~\cite{ferre2025wireframe}, and PLNet~\cite{xu2024airslam}. We also add more heavyweight baselines (more than 10M params) as a comparison: DeDoDe v2~\cite{edstedt2024dedodev2}, DaD~\cite{edstedt2025dad} keypoints paired with DeDoDe v2 descriptors, Ripe~\cite{kunzel2025ripereinforcementlearningunlabeled}, and RDD~\cite{rdd}.

~\\
\noindent
\textbf{Homography Estimation. }
The HPatches~\cite{Balntas2017} dataset contains 580 image pairs, with either illumination or viewpoint variations.
We resize the longest side of the images to 800 pixels while keeping the aspect ratio, and extract the top 1024 keypoints per image.
The benchmark evaluates the estimation of the dominant planar homography from keypoint correspondences.
After matching, we use PoseLib~\cite{Larsson2020} to estimate the homography, and evaluate the Area Under the Curve (AUC) of the reprojection error of the four image corners at thresholds of 1, 3, and 5 pixels, as in \cite{sarlin20superglue}. Results can be found in \cref{tab:hptaches_points}.

~\\
\noindent
\textbf{Outdoor Relative Pose Estimation.}  
The MegaDepth~\cite{Zheng2023} benchmark for feature matching contains 1,500 image pairs sampled with various levels of overlap, taken from two outdoor landmark scenes of the original MegaDepth dataset. For this benchmark, we resize images to 1600 pixels on the long side and use the top 2048 keypoints for each extractor. After matching, we leverage PoseLib~\cite{Larsson2020} to estimate the relative translation and rotation from the correspondences, using the 5-point algorithm~\cite{Nister03} to obtain a robust estimate. The results are returned as the AUC of the pose error (maximum of the angular error in both translation and rotation) at three error thresholds. Results are shown in \cref{tab:megadepth_points}.

\input{tables/results_point_megadepth}

~\\
\noindent
\textbf{Indoor Relative Pose Estimation.} 
\label{sec:scannet}
ScanNet~\cite{dai2017scannet} is a large-scale RGB-D indoor dataset, from which \cite{sarlin20superglue} sampled 1,500 overlapping image pairs with ground-truth poses. Indoor scenes are particularly challenging due to repetitive structures and texture-less regions, although line features are abundant and can effectively complement point features. The evaluation protocol is identical to that of the previous section, and \cref{tab:scannet_points} shows the results.

\input{tables/results_point_scannet}

~\\
\noindent
\textbf{Discussion of the Point Evaluation.}
While \OURS~is mainly distilled from existing methods for the point branch, it consistently ranks in the top three among all these benchmarks. The first reason is that we explicitly selected the best teacher for each component. For example, while ALIKED~\cite{Zhao2023} has generally the best descriptors overall, we noticed that its points underperform in indoor scenarios, as can be seen in \cref{tab:scannet_points}, probably because of its unsupervised training and lack of indoor training data. In contrast, SuperPoint~\cite{DeTone2018} keypoints are supervised as corners and naturally work better in indoor scenarios where corners are abundant. We also empirically found that mixing SuperPoint with DaD~\cite{edstedt2025dad} keypoints could further boost the overall performance. A second reason for the increased performance of our approach is that jointly detecting points and lines benefits the network by helping it identify more stable keypoints along line structures, which are naturally abundant in indoor environments.

%------------------------------------------------------------------------------
\subsection{Line Evaluation}
%------------------------------------------------------------------------------
%

We evaluate our line detection on two standard datasets to test the robustness of the methods: HPatches~\cite{Balntas2017} and RDNIM~\cite{Pautrat_2020_ECCV}.
The latter consists of 1,722 image pairs related by a homography and with challenging day-night variations.

Following~\cite{Pautrat_Lin_2021_CVPR,Pautrat_2023_DeepLSD}, we evaluate the repeatability and localization error metrics. Since our teacher's lines are semi-supervised and generic~\cite{Pautrat_2023_DeepLSD}, evaluating on a biased set of lines such as in the Wireframe dataset~\cite{Huang2018} would not provide a fair comparison for generic line detection, and we instead report quality metrics of our lines for downstream tasks.
Given a pair of images with ground-truth homography, we detect line segments in both images and establish one-to-one correspondences by warping lines between frames. To decide which pairs should be considered as a match, we compute the symmetric orthogonal distance: for each pair of lines, we calculate the average distance from both endpoints of a segment to the other line, compute this bidirectionally, and take the mean. Repeatability measures the ratio of lines whose match has an error below 1, 3, and 5 pixels, and the localization error returns the average distance of the 10, 50, and 300 most accurate matches. We also compute a homography estimation score, as in~\cite{DeTone2018,Pautrat_2023_DeepLSD}. Reusing the previous line matches, we sample minimal sets of 4 line matches and run LO-RANSAC~\cite{Lebeda2012loransac} with the orthogonal line distance as reprojection error to estimate a homography.

\input{tables/results_line_evaluation}

\cref{tab:rdnim_hpatches_horizontal} compares \OURS~to two classical detectors: LSD~\cite{Gioi2010} and ELSED~\cite{suarez2022elsed}; the recent learned baselines TP-LSD~\cite{huang2020tplsd}, SOLD2~\cite{Pautrat_Lin_2021_CVPR}, M-LSD tiny~\cite{Gu2021}, DeepLSD~\cite{Pautrat_2023_DeepLSD}, Linea~\cite{janampa2025linea}, and ScaleLSD~\cite{ScaleLSD}; and the joint point-line extractors Wireframe~\cite{ferre2025wireframe} and PLNet~\cite{xu2024airslam}. 
For LSD, we use the implementation available at \url{https://github.com/iago-suarez/pytlsd} and otherwise use the authors' implementations with default parameters for all the other models. Despite its light backbone, \OURS~has the highest repeatability across the board and often ranks second in terms of localization error and homography estimation. 
This shows that line detection is a low-level task that can be solved with as few as 780k parameters. Notably, \OURS~improves over its teacher, DeepLSD~\cite{Pautrat_2023_DeepLSD}, on the RDNIM dataset. This is due to the removal of the angle field prediction, which yields inaccurate predictions on such challenging images.

\input{tables/results_localization}

%------------------------------------------------------------------------------
\subsection{Point-Line Applications}
%------------------------------------------------------------------------------
%
\noindent
\textbf{Visual Localization.}
We evaluate the combination of points and lines for the task of visual localization using lightweight detectors. We use the 7Scenes dataset~\cite{7scenes}, consisting of 7 indoor scenes with RGB images and ground-truth poses. We focus here on the scene with the most lines, Stairs, and report the other scenes in the supplementary. We do not use depth information in this experiment. We run the point SfM on database images and find the camera poses of query images using the hloc~\cite{sarlin2019coarse,hloc} framework. We then add line segments to the 3D model and perform pose estimation of the query images with a combination of points and lines using the LIMAP framework~\cite{Liu_2023_LIMAP}. Line matching is performed as originally proposed in \cite{pautrat_suarez_2023_gluestick}: we extract descriptors at line endpoints, create an assignment matrix between the lines of both images by summing the matching scores of both endpoints, and finally solve for the best assignment with the Sinkhorn algorithm~\cite{sinkhorn_1967_concerning,cuturi_2013_sinkhorn}. For each image, we extract a maximum of 2048 keypoints and 3000 lines, and resize the maximum side of the image to 1024 pixels. Camera poses are estimated using the solvers in PoseLib~\cite{Larsson2020}. We report the median translation and rotation error on the retrieved camera pose, as well as the pose estimation accuracy at a 5cm / 5$^\circ$ threshold.

In \cref{tab:visual_localization_7scenes} we compare the results of point-only pipelines with point-line approaches and first observe that \OURS~achieves the best results among all point baselines. This aligns with \cref{sec:scannet}, where \OURS~excels in indoor scenarios. We attribute this to joint training with lines, encouraging keypoints to be better localized on lines, which are abundant in indoor scenarios. Second, the addition of lines improves results over points in isolation, especially in scenes that are challenging for keypoints but rich in lines, like the Stairs scene. This aligns with existing works~\cite{Liu_2023_LIMAP,pautrat_suarez_2023_gluestick} which demonstrated the complementarity of lines with respect to points. Finally, \OURS~with points and lines has the best overall results, showcasing the synergy of jointly learned features.

~\\
\noindent

\input{tables/eval_eth3d_triangulation}
\textbf{3D Reconstruction.}
We further evaluate 3D point-cloud reconstruction from multi-view posed images. We evaluate the accuracy and completeness of the point triangulation \cite{schoenberger2016sfm} on ETH3D \cite{schops2017multi} following existing practices \cite{dusmanu2020multi,lindenberger2021pixel}. All detectors are tested with mutual nearest neighbor matching with a maximum number of keypoints of 2048. \cref{tab:eth3d_triangulation} shows that we achieve strong performance both in terms of accuracy and completeness, consistently exceeding SuperPoint \cite{DeTone2018} and ALIKED \cite{Zhao2023}, while falling behind DISK \cite{tyszkiewicz2020disk}, which performs particularly well under this setup. This further proves the applicability of \OURS\ for accurate 3D reconstruction with a compact, unified architecture.

%------------------------------------------------------------------------------
\subsection{Efficiency}
\label{sec:efficiency}
%------------------------------------------------------------------------------
%

We benchmark efficiency against combinations of state-of-the-art models on an NVIDIA RTX 2080 Ti GPU (11 GB) as well as on an Intel Core i9-13900K CPU, on 500 images from the Oxford-Paris dataset~\cite{radenovic2018} with a batch size of one. Independent point and line extractors are run sequentially. Images are resized to $800 \times 800$. As shown in \cref{tab:efficiency}, \OURS\ achieves a mean GPU inference time of 70 ms, providing roughly a $4\times$ speedup over the SOTA ALIKED \cite{Zhao2023} + DeepLSD \cite{Pautrat_2023_DeepLSD}. This efficiency stems from joint inference, a smaller network architecture, and an optimized line post-processing pipeline. \OURS\ is also competitive with lightweight approaches such as ALIKED~\cite{Zhao2023} + M-LSD~\cite{Gu2021}, while obtaining better performance (e.g. in \cref{tab:rdnim_hpatches_horizontal}). Furthermore, our number of parameters is lower by an order of magnitude compared to previous joint extractors (Wireframe~\cite{ferre2025wireframe} and PLNet~\cite{xu2024airslam}), and GPU latency is $37\%$ lower.

\input{tables/runtime_evaluation}

%------------------------------------------------------------------------------
\subsection{Ablation study}
%------------------------------------------------------------------------------
%

We validate our design choices for the line detection implementation on the HPatches dataset~\cite{Balntas2017}. In \cref{tab:ablation} we compare our final model to various versions of our line post-processing. First, we show that DeepLSD without an angle field prediction (2) obtains a higher repeatability and lower localization error than the vanilla DeepLSD (1). Next, we show the benefits of our enhancements over the vanilla LSD (3). For each modification, computing the gradient angle in parallel (4), sampling seed points with a stride of 2 (5), selecting only a subset of the seed points (6), and offloading the image preprocessing to the GPU (7), we observe consistent improvements in latency, with no degradation in performance. 

\input{tables/runtime_ablation}

%% file: tables/results_point_hpatches.tex
\begin{table}[tb]
\centering
\scriptsize
\caption{\textbf{Homography estimation on HPatches~\cite{Balntas2017}.} We report the latency in ms and the Area Under the Curve (AUC) of correctly predicted homographies for various pixel error thresholds. Methods are grouped into lightweight and heavyweight categories. Within each category, the best result is shown in \textbf{bold}, and the second best is \underline{underlined}.}
\setlength{\tabcolsep}{4.5pt}
\renewcommand{\arraystretch}{1}
\begin{tabular}{lcccc}
\toprule
Method & AUC @1px $\uparrow$ & AUC @3px $\uparrow$ & AUC @5px $\uparrow$ & Latency (ms) $\downarrow$ \\
\midrule

\multicolumn{5}{l}{\textbf{Lightweight Methods}}\\
\midrule
SIFT~\cite{lowe_2004}        & 35.1 & 65.0 & 75.3 & 237\\
SuperPoint~\cite{DeTone2018} & 35.6 & 65.2 & 75.7 & \snd{50}\\
DISK~\cite{tyszkiewicz2020disk} & 29.9 & 61.8 & 73.2 & 100\\
XFeat~\cite{potje2024cvpr}   & 34.4 & 62.2 & 73.3 & \fst{44}\\
Wireframe~\cite{ferre2025wireframe} & \underline{38.3} & 66.4 & 76.1 & 308\\
PLNet~\cite{xu2024airslam}   & 35.5 & \underline{67.0} & \underline{77.4} & 158 \\
\OURS~(Ours)                 & \textbf{39.2} & \textbf{67.9} & \textbf{77.7} & 62\\

\midrule
\multicolumn{5}{l}{\textbf{Heavyweight Methods}}\\
\midrule
DeDoDe v2~\cite{edstedt2024dedodev2}
& \textbf{40.9} & \textbf{69.1} & 78.4 & 348\\
DaD~\cite{edstedt2025dad} + DeDoDe v2~\cite{edstedt2024dedodev2}
& \underline{39.7} & \underline{68.6} & \textbf{78.6} & \textbf{164}\\
Ripe~\cite{kunzel2025ripereinforcementlearningunlabeled}                         & 37.5 & 58.9 & 69.3 & \underline{193}\\
RDD~\cite{rdd}
& 34.6 & 60.0 & 68.4 & 522 \\

\bottomrule
\end{tabular}
\label{tab:hptaches_points}
\optimizedfigspacing
\end{table}

%% file: tables/results_point_megadepth.tex
\begin{table}[tb]
\centering
\scriptsize
\caption{\textbf{Relative pose estimation on MegaDepth~\cite{Zheng2023}.} The latency and AUC of correctly retrieved poses under various angular error thresholds are reported. Methods are grouped into lightweight and heavyweight categories. Within each category, the best result is highlighted in \textbf{bold} and the second best is \underline{underlined}. Results reported in parentheses are not considered for ranking due to overlap between the training data and the MegaDepth benchmark test set. }
\setlength{\tabcolsep}{5pt}
\renewcommand{\arraystretch}{1}
\begin{tabular}{lcccc}
\toprule
Method & AUC @5$^\circ$ $\uparrow$ & AUC @10$^\circ$ $\uparrow$ & AUC @20$^\circ$ $\uparrow$ & Latency (ms) $\downarrow$ \\
\midrule

\multicolumn{5}{l}{\textbf{Lightweight Methods}}\\
\midrule
SIFT~\cite{lowe_2004} & 40.1 & 53.6 & 64.4 & 729\\
SuperPoint~\cite{DeTone2018} & 51.3 & 64.4 & 73.7 & \underline{78} \\
DISK~\cite{tyszkiewicz2020disk} & 53.9 & 66.1 & 75.3 & 335 \\
ALIKED~\cite{Zhao2023} & \textbf{59.8} & \textbf{72.0} & \textbf{81.1} & 79 \\
XFeat~\cite{potje2024cvpr} & 44.6 & 59.0 & 70.2 & \textbf{28} \\
Wireframe~\cite{ferre2025wireframe} & 47.7 & 60.7 & 70.8 & 307 \\
PLNet~\cite{xu2024airslam} & 37.1 & 51.6 & 63.9 & 288 \\
\OURS~(Ours) & \underline{58.2} & \underline{70.8} & \underline{79.4} & 80\\
\midrule

\multicolumn{5}{l}{\textbf{Heavyweight Methods}}\\
\midrule
Ripe~\cite{kunzel2025ripereinforcementlearningunlabeled} & \underline{58.2} & \underline{70.7} & \underline{79.8} & \underline{751} \\
DeDoDe v2~\cite{edstedt2024dedodev2}
& 57.2 & 69.6 & 78.8 & 1142 \\
DaD~\cite{edstedt2025dad}+DeDoDe v2~\cite{edstedt2024dedodev2}
& \textbf{60.3} & \textbf{72.8} & \textbf{81.6} & \textbf{485} \\
RDD~\cite{rdd}\tablefootnote{RDD is reported separately since its training data overlaps with the MegaDepth benchmark test set, making these results not directly comparable to the other methods.}
& \textit{(64.8)} & \textit{(77.2)} & \textit{(85.8)} & $\infty$\tablefootnote{RTX 2080 Ti ran out of memory} \\
\bottomrule
\end{tabular}
\label{tab:megadepth_points}
\end{table}

%% file: tables/results_point_scannet.tex
\begin{table}[tb]
\centering
\scriptsize
\caption{\textbf{Relative pose estimation on ScanNet~\cite{dai2017scannet}.} The latency and AUC of correctly estimated poses under various angular error thresholds are reported. Methods are grouped into lightweight and heavyweight categories. Within each category, the best result is highlighted in \textbf{bold} and the second best is \underline{underlined}.}
\setlength{\tabcolsep}{5pt}
\renewcommand{\arraystretch}{1}
\begin{tabular}{lcccc}
\toprule
Method & AUC @5$^\circ$ $\uparrow$ & AUC @10$^\circ$ $\uparrow$ & AUC @20$^\circ$ $\uparrow$ & Latency (ms) \\
\midrule

\multicolumn{5}{l}{\textbf{Lightweight Methods}}\\
\midrule
SIFT~\cite{lowe_2004} & 6.7 & 14.9 & 24.2 & 742\\
SuperPoint~\cite{DeTone2018} & 15.5 & \underline{29.8} & \underline{43.4} & 83\\
DISK~\cite{tyszkiewicz2020disk} & 10.1 & 19.2 & 29.5 & 348\\
ALIKED~\cite{Zhao2023} & 6.9 & 13.3 & 20.1 & \underline{82}\\
XFeat~\cite{potje2024cvpr} & \textbf{16.4} & \textbf{31.2} & \textbf{45.7} & \textbf{27}\\
Wireframe~\cite{ferre2025wireframe} & 11.5 & 22.3 & 34.5 & 293\\
PLNet~\cite{xu2024airslam} & 9.5 & 20.7 & 34.8 & 427\\
\OURS~(Ours) & \underline{15.6} & 28.9 & 41.4 & 83\\

\midrule
\multicolumn{5}{l}{\textbf{Heavyweight Methods}}\\
\midrule
Ripe~\cite{kunzel2025ripereinforcementlearningunlabeled} & 5.5 & 11.6 & 18.9 & 782\\
DeDoDe v2~\cite{edstedt2024dedodev2}
& 9.6 & 18.5 & 28.4 & \underline{1135}\\
DaD~\cite{edstedt2025dad}+DeDoDe v2~\cite{edstedt2024dedodev2}
& \textbf{14.6} & \underline{27.9} & \underline{41.6} & \textbf{462}\\
RDD~\cite{rdd}
& \underline{14.2} & \textbf{28.2} & \textbf{42.7} & $\infty$\tablefootnote{RTX 2080 Ti ran out of memory}\\

\bottomrule
\end{tabular}
\label{tab:scannet_points}
\optimizedfigspacing
\end{table}

%% file: tables/results_line_evaluation.tex
\begin{table*}[tb]
\centering
\tiny
\caption{\textbf{Line detection evaluation on HPatches~\cite{Balntas2017} and RDNIM~\cite{Pautrat_2020_ECCV}.} We compare the line localization error (Loc Err) on a fixed set of l lines, homography estimation (H Estim), and line repeatability (Rep) at various pixel thresholds. Horizontal lines separate Traditional / Learned / Joint methods.}
\setlength{\tabcolsep}{1.4pt}

\begin{tabular}{lcccccccccccccccccc}
\toprule
& \multicolumn{9}{c}{HPatches} & \multicolumn{9}{c}{RDNIM} \\
\cmidrule(lr){2-10} \cmidrule(lr){11-19}

& \multicolumn{2}{c}{Loc Err} & \multicolumn{3}{c}{H Estim} & \multicolumn{3}{c}{Rep} & Time
& \multicolumn{2}{c}{Loc Err} & \multicolumn{3}{c}{H Estim} & \multicolumn{3}{c}{Rep} & Time \\
\cmidrule(lr){2-3} \cmidrule(lr){4-6} \cmidrule(lr){7-9} \cmidrule(lr){11-12} \cmidrule(lr){13-15} \cmidrule(lr){16-18}

 & 50l & 300l & 1px & 3px & 5px & 1px & 3px & 5px & (ms)
 & 50l & 300l & 1px & 3px & 5px & 1px & 3px & 5px & (ms) \\
\midrule

% --- Traditional ---
LSD~\cite{Gioi2010}
& \snd{0.50} & \snd{1.42}
& \fst{58.3} & \fst{88.3} & \snd{90.9}
& \snd{26.4} & \snd{53.6} & 61.6 & 150
& 1.78 & 1.85
& 7.4 & 44.8 & 57.6
& 9.2 & 30.0 & 37.7 & 85 \\

ELSED~\cite{suarez2022elsed}
& 0.71 & 1.63
& 40.7 & 65.8 & 67.9
& 18.8 & 41.8 & 50.8 & 67
& 1.92 & 1.96
& 5.9 & 30.8 & 40.9
& 8.9 & 25.9 & 30.9 & \fst{40} \\

\midrule
% --- Learned ---
TP-LSD~\cite{huang2020tplsd}
& 1.49 & 1.61
& 10.9 & 34.6 & 42.6
& 11.6 & 42.9 & 57.2 & \snd{65}
& 1.93 & 1.93
& 0.2 & 6.3 & 14.2
& 4.9 & \snd{33.3} & \fst{47.5} & 52 \\

SOLD2~\cite{Pautrat_Lin_2021_CVPR}
& 1.09 & \fst{1.29}
& 23.7 & 48.5 & 55.4
& 19.8 & 44.9 & 51.5 & 745
& \snd{1.39} & \fst{1.41}
& 0.6 & 9.4 & 16.0
& 8.8 & 20.0 & 24.0 & 546 \\

M-LSD~\cite{Gu2021}
& 2.10 & 2.21
& 15.0 & 54.1 & 63
& 8.9 & 35.6 & 51.2 & 91
& 2.41 & 2.41
& 0.5 & 13.4 & 26.0
& 4.3 & 22.6 & 34.9 & 70 \\

DeepLSD~\cite{Pautrat_2023_DeepLSD}
& 0.55 & 1.43
& \snd{58.1} & 87.4 & 90.6
& \fst{27.2} & \fst{54.1} & \snd{61.8} & 473
& 1.73 & 1.81
& 9.4 & 45.3 & 60.5
& \snd{10.1} & 31.8 & 39.5 & 294 \\

Linea~\cite{janampa2025linea}
& 1.81 & 2.03
& 15.9 & 46.5 & 53.0
& 8.3 & 35.3 & 50.0 & \fst{63}
& 2.25 & 2.26
& 0.3 & 9.0 & 17.1
& 2.6 & 25.0 & 40.0 & \snd{44} \\

ScaleLSD~\cite{ScaleLSD}
& 0.57 & 1.56
& 53.3 & 87.6 & \fst{91.1}
& 20.2 & 46.5 & 55.2 & 177
& 1.71 & 1.83
& \fst{9.8} & \snd{47.4} & \snd{63.4}
& 9.1 & 25.0 & 31.4 & 166 \\

\midrule
% --- Joint ---
Wireframe~\cite{ferre2025wireframe}
& 1.05 & 1.67
& 33.7 & 72.8 & 79.4
& 16.7 & 43.7 & 52.3 & 294
& 1.88 & 1.90
& 2.0 & 20.4 & 31.4
& 7.3 & 25.0 & 31.9 & 291 \\

PLNet~\cite{xu2024airslam}
& \fst{0.39} & 1.57
& 41.9 & 67.0 & 71.7
& 11.5 & 30.4 & 40.0 & 240
& \fst{0.99} & 2.35
& 8.3 & \fst{49.4} & \fst{66.8}
& 4.0 & 16.1 & 24.4 & 199 \\

\OURS\ (Ours)
& 0.51 & \snd{1.42}
& 57.0 & \snd{88.1} & \snd{90.9}
& \fst{27.2} & \fst{54.1} & \fst{61.9} & 141
& 1.62 & \snd{1.69}
& \snd{9.6} & 45.3 & 58.0
& \fst{13.3} & \fst{35.9} & \snd{43.6} & 83 \\

\bottomrule
\end{tabular}
\label{tab:rdnim_hpatches_horizontal}
\end{table*}

%% file: tables/results_localization.tex
\begin{table}[tb]
\centering
\scriptsize
\caption{\textbf{Visual localization on Stairs scene of the 7Scenes dataset~\cite{7scenes}.} We report the median translation / rotation errors (cm / deg) and pose accuracy at 5 cm / 5° threshold. Please refer to our supplementary material for results on all scenes.}
\setlength{\tabcolsep}{8pt}
\begin{tabular}{l l c c}
\toprule
\textbf{} & Method & T / R err. $\downarrow$ & Acc. $\uparrow$ \\
\midrule
\multirow{6}{*}{\makecell{\textbf{Points}}}
& SuperPoint~\cite{DeTone2018} & \phantom{0}6.2 / 1.65 & 36.6 \\
& DISK~\cite{tyszkiewicz2020disk} & \phantom{0}7.2 / 2.16 & 28.5 \\
& ALIKED~\cite{Zhao2023} & \phantom{0}7.8 / 1.98 & 35.0 \\
% & DaD~\cite{edstedt2025dad}+DeDoDe v2~\cite{edstedt2024dedodev2} & 13.8 / 3.16 & 14.6 \\
& PLNet~\cite{xu2024airslam} & \phantom{0}6.2 / 1.68 & 38.2 \\
& \OURS~(Ours) & \fst{\phantom{0}5.1 / 1.39} & \fst{49.1} \\
\midrule
\multirow{8}{*}{\makecell{\textbf{Points} \\ \textbf{+ Lines}}}
& SuperPoint~\cite{DeTone2018}+DeepLSD~\cite{Pautrat_2023_DeepLSD} & \phantom{0}5.0 / 1.33 & 49.6 \\
& ALIKED~\cite{Zhao2023}+TP-LSD~\cite{huang2020tplsd} & \phantom{0}5.2 / 1.48 & 48.8 \\
& ALIKED~\cite{Zhao2023}+M-LSD~\cite{Gu2021} & \phantom{0}5.2 / 1.52 & 48.5 \\
& ALIKED~\cite{Zhao2023}+DeepLSD~\cite{Pautrat_2023_DeepLSD} & \phantom{0}5.1 / 1.48 & 49.6 \\
% & SuperPoint~\cite{DeTone2018}+ScaleLSD~\cite{ScaleLSD} & 5.1 / 1.33 & 49.5 \\
% & ALIKED~\cite{Zhao2023}+ScaleLSD~\cite{ScaleLSD} & 5.1 / 1.50 & 49.4 \\
% & DaD~\cite{edstedt2025dad}+ScaleLSD~\cite{ScaleLSD} & \phantom{0}5.0 / 1.31 & 49.8 \\
& Wireframe~\cite{ferre2025wireframe} & \phantom{0}4.6 / 1.30 & 53.8 \\
& PLNet~\cite{xu2024airslam} & \phantom{0}5.0 / 1.36 & 50.1 \\
& \OURS~(Ours) & \fst{\phantom{0}4.4 / 1.23} & \fst{54.6} \\
\bottomrule
\end{tabular}
\label{tab:visual_localization_7scenes}
\optimizedfigspacing
\end{table}

%% file: tables/eval_eth3d_triangulation.tex
\begin{table}[tb]
\centering
\scriptsize
\caption{\textbf{Evaluation on multi-view point triangulation on ETH3D \cite{schops2017multi}.}}
\setlength{\tabcolsep}{4.5pt}
\renewcommand{\arraystretch}{1}
\begin{tabular}{lcccccccc}
\toprule
\multirow{2}{*}{Method} & \multicolumn{3}{c}{\textbf{Completeness (\%)}} & & \multicolumn{3}{c}{\textbf{Accuracy (\%)}} \\
& 1cm & 2cm & 5cm & & 1cm & 2cm & 5cm \\
\midrule
SuperPoint~\cite{DeTone2018} & 0.17 & 0.81 & 4.34 & & 49.52 & 64.73 & 80.01 \\
DISK~\cite{tyszkiewicz2020disk} & \textbf{0.33} & \textbf{1.18} & 4.41 & & \textbf{69.22} & \textbf{81.10} & \textbf{91.29} \\
ALIKED~\cite{Zhao2023} & 0.13 & 0.56 & 2.76 & & 56.61 & 70.78 & 85.18 \\
Wireframe~\cite{ferre2025wireframe} & 0.19 & 0.98 & \textbf{5.41} & & 50.98 & 66.44 & 82.50 \\
\OURS ~(Ours) & \snd{0.30} & \snd{1.09} & \snd{4.45} & & \snd{64.80} & \snd{77.57} & \snd{88.87} \\
\bottomrule
\end{tabular}
\label{tab:eth3d_triangulation}
\end{table}

%% file: tables/runtime_evaluation.tex
\begin{table}[tb]
    \centering
    \scriptsize
    \caption{\textbf{Runtime evaluation.} Parameter counts are reported, and latencies are averaged over 500 sequential detections on an NVIDIA RTX 2080 Ti GPU and Intel Core i9-13900K CPU.} 
    \setlength{\tabcolsep}{5pt}
    \renewcommand{\arraystretch}{1}
    \begin{tabular}{lccc}
    \toprule
      & \multirow{2}{*}{Nb params $\downarrow$} & \multicolumn{2}{c}{Latency (ms) $\downarrow$} \\
      \cmidrule(lr){3-4}
      & & GPU & CPU \\
     \midrule
     SuperPoint~\cite{DeTone2018} + DeepLSD~\cite{Pautrat_2023_DeepLSD} & 9.8M & 293 & 2259 \\
     ALIKED~\cite{Zhao2023} + DeepLSD~\cite{Pautrat_2023_DeepLSD} & 9.2M & 286 & 2239 \\
     ALIKED~\cite{Zhao2023} + M-LSD~\cite{Gu2021} & \snd{1.3M} & \fst{54} & \fst{525} \\
     ALIKED~\cite{Zhao2023} + TP-LSD~\cite{huang2020tplsd} & 25M & \snd{56} & 1582 \\
     DaD~\cite{edstedt2025dad} + DeDoDe v2~\cite{edstedt2024dedodev2} + ScaleLSD~\cite{ScaleLSD} & 120M & 147 & >200K\\
     Wireframe~\cite{ferre2025wireframe} & 5.8M & 266 & 2493 \\
     PLNet~\cite{xu2024airslam} & 7.5M & 112 & 6233 \\
     \OURS~(Ours) & \fst{0.78M} & 70 & \snd{976} \\
     \bottomrule
     \end{tabular}
     \label{tab:efficiency}
\end{table}

%% file: tables/runtime_ablation.tex
\begin{table}[tb]
    \centering
    \scriptsize
    \caption{\textbf{Ablation study on the HPatches dataset~\cite{Balntas2017}.} We compare the localization error, homography estimation, repeatability, and latency for variants of our approach. }
    \setlength{\tabcolsep}{3pt}
    \renewcommand{\arraystretch}{1}
    \begin{tabular}{clcccc}
    \toprule
      & & Loc Error $\downarrow$  & H estimation $\uparrow$ & Repeatability $\uparrow$ & Latency (ms) $\downarrow$ \\
     \midrule
     \textit{(1)} & DeepLSD~\cite{Pautrat_2023_DeepLSD} & 1.432 & 90.6 & 27.2 & 233 \\ 
     \textit{(2)} & DeepLSD no AF & \fst{1.341} & 90.6  & \snd{28.2} & 218 \\ 
     \textit{(3)} & LSD~\cite{Gioi2010} & 1.414 & \fst{91.1} & 26.3 &  36 \\ 
     \textit{(4)} & (3) + parall. ang. comp.  & 1.414 & \fst{91.1} & 26.3 & 27 \\ 
     \textit{(5)} & (4) + stride 2  & 1.455 & 90.6 & 26.3 &  19 \\
     \textit{(6)} & (5) + subset  & 1.476  & 90.6 & 26.3 &  \snd{17} \\ 
     \textit{(7)} & (6) + GPU & \snd{1.343} & \snd{90.7} & \fst{28.9} &  \fst{11} \\ 
     \bottomrule
     \end{tabular}
     \label{tab:ablation}
     \optimizedfigspacing
\end{table}

%% file: sec/5_conclusion.tex
\section{Conclusion}
\label{sec:conclusion}
In this work, we introduced \OURS, a unified architecture capable of jointly extracting keypoints, line segments, and feature descriptors, while maintaining high efficiency and preserving the performance of each individual component. Our method achieves a 4$\times$ speedup compared to combining existing standalone components, while reducing the memory footprint by at least one order of magnitude. Looking ahead, we plan to integrate \OURS\ with a lightweight, jointly learned matcher to enable end-to-end point-line image matching. We believe that our efficient, joint feature extractor marks an important step toward broadening access to advanced geometric perception, ultimately facilitating the deployment of point-line applications on embedded platforms and in low-compute environments.

\subsubsection{Acknowledgments: }

We would like to thank Viktor Larsson for valuable discussions regarding the LSD line detector. During our efforts to develop a faster variant of LSD, he suggested precomputing the distance field on the GPU. This insight ultimately helped shape the accelerated version of LSD presented in this work. We also thank Olivia Buset for her help during this project.

%% file: sec/suppl.tex
\clearpage
\setcounter{page}{1}

\title{Supplementary Material}
\author{}
\institute{}
\maketitle

We provide additional analyses of our proposed \OURS\ feature extractor in the following. \cref{sec:supp_distillation} provides additional information about the distillation process. \cref{sec:supp_joint_training} justifies the joint point and line training. \cref{sec:supp_bad_gpu} provides another perspective on the inference cost, by comparing baselines on a low-end GPU, simulating an embedded GPU. \cref{sec:supp_visloc} presents the evaluation of point-line visual localization on the full 7Scenes dataset. \cref{sec:supp_line_detect_ex} displays additional visualizations of the predicted lines. Finally, \cref{sec:supp_matching} shows qualitative examples of point and line matching.

\section{Score map distillation}
\label{sec:supp_distillation}

We use the raw score map predicted by DaD~\cite{edstedt2025dad} when fusing its keypoint heatmap with the SuperPoint~\cite{DeTone2018} output. This score map can be interpreted as a prior probability indicating whether a pixel corresponds to a keypoint, similarly to the formulation used in SuperPoint. Therefore, taking the maximum response between the two maps is a meaningful fusion strategy.

The choice of $\lambda = 200$ for balancing positive and negative keypoints follows the setting introduced in MTLDesc~\cite{wang2022mtldesclookingwiderbetter}. For images of size $800 \times 800$, this corresponds to approximately 3,200 keypoints.

\section{Joint point and line training}
\label{sec:supp_joint_training}

Beyond computational efficiency, joint point-and-line training also improves the quality of the learned point features. As shown in~\cref{tab:joint_training_ablation}, the jointly trained model consistently outperforms its point-only counterpart on MegaDepth relative pose estimation, demonstrating that line supervision provides complementary geometric information during training. Lines encode structural cues such as edges, boundaries, and intersections, which help the network localize keypoints more accurately and learn features that are more geometrically consistent across views. This additional supervision acts as a form of regularization, encouraging the shared backbone to capture both local appearance and higher-level scene geometry. 

\input{tables/results_ablation}

\section{Efficiency on constrained resources}
\label{sec:supp_bad_gpu}

\input{tables/supp_evaluation_old_nvidia}
~\\
\noindent
To assess efficiency in low-resource environments, representative of embedded applications or consumer devices, we additionally benchmark efficiency using combinations of state-of-the-art models with large backbones on an NVIDIA GeForce GTX 1050 Ti GPU with 4 GB of VRAM. Similar to the setup of Table~\ref{tab:efficiency} of the main paper, we use 500 images from the Oxford-Paris dataset~\cite{radenovic2018} with a batch size of 1 to measure latency. Independent point and line extractors are executed sequentially, and all images are resized to $800 \times 800$.

As shown in Table~\ref{tab:old_nvidia}, \OURS{} achieves a mean inference time of 186\,ms, yielding roughly a $4\times$ speedup over ALIKED + DeepLSD. More importantly, we observe a significant speed difference between \OURS{} and DaD + ScaleLSD with up to $18\times$ speedup. The reason is that DaD + ScaleLSD, and especially ScaleLSD, rely on modern powerful GPUs to be efficient and perform poorly on limited hardware due to their large model size. In contrast, \OURS{} performs well in both settings due to its lightweight configuration.

\section{Visual localization}
\label{sec:supp_visloc}

~\\
\noindent
\cref{tab:visual_localization_7scenes} of the main paper reports the results for visual localization on the Stairs scene of the 7Scenes~\cite{7scenes} dataset for both points-only and points+lines jointly. In \cref{tab:visual_localization_7scenes_all}, we give the results for all seven scenes of the dataset.

\input{tables/supp_localization}

The results demonstrate that our method achieves highly competitive performance across visual localization tasks on the 7Scenes dataset, consistently outperforming or matching strong baselines in both point-only and points+lines configurations. In the point-only category, \OURS{} secures the best accuracy for five of the seven scenes. While SuperPoint and PLNet are the strongest competing baselines, DISK and especially ALIKED fall behind. In general, translation and rotation errors show less variance. Similar to the accuracy results, \OURS{}, SuperPoint, and PLNet consistently achieve the best results.

Looking at points+lines performance, we observe that SuperPoint+DeepLSD and PLNet represent the strongest competing baselines and outperform \OURS{} on Chess, Heads, and Office, with SuperPoint+DeepLSD showing especially strong results on Chess, while PLNet scores best in the Heads and Office scenes, where long structural lines are abundant. However, our method demonstrates clear advantages on scenes with complex textures and varying illumination conditions such as Fire, Pumpkin, and RedKitchen, where we outperform these baselines in accuracy while maintaining comparable or better translation/rotation errors. We see the strong points-only performance of \OURS{}, SuperPoint, and PLNet translating to the points+lines scenario as well. Overall, \OURS{} performs best on most scenes for points-only and points+lines localization on 7Scenes while being significantly faster and more lightweight than the baselines.

\section{Qualitative comparison of line detection results}
\label{sec:supp_line_detect_ex}

\input{figures/supp_detection_ex}

Similar to \cref{fig:example_detection}, \cref{fig:supp_example_detection} depicts additional line detection results. All detectors are run with their default settings.
The qualitative detection results clearly show that ScaleLSD and Wireframe consistently detect significantly fewer lines in the images. They focus on long structural lines and miss many generic valid lines that could be useful for some applications (e.g. visual localization). LSD, DeepLSD and \OURS{} show a higher recall of line segments, even in structural regions. The difference between LSD, DeepLSD, and \OURS{} is less pronounced; nonetheless, we observe that DeepLSD and \OURS{} can recall more line segments in areas of subtle image gradients (see second row of \cref{fig:supp_example_detection}). We explain this with the strong learned line distance field informing the LSD algorithm in low gradient areas. 

Overall, \OURS{} produces qualitative results superior to Wireframe and ScaleLSD in particular, while showing DeepLSD-level detection quality. Notably, we achieve this while being significantly faster and substantially more lightweight, as shown in \cref{tab:efficiency}. This advantage is even more pronounced on lower-tier hardware as shown in Section \ref{sec:supp_bad_gpu}.

\section{Point and line matching qualitative examples}
\label{sec:supp_matching}

We provide qualitative examples of point and line matching with \OURS\ on image pairs from HPatches~\cite{Balntas2017}, with both illumination-varying and viewpoint-varying pairs. Features are extracted after resizing each image so that its longer side is 800 pixels while preserving the aspect ratio, as in our HPatches evaluation (\cref{sec:experiments}). Keypoints are matched by mutual nearest neighbors over their descriptors, and line segments likewise over their two endpoint descriptors. For better visibility, each pair is shown at its original aspect ratio, and only the top-300 matches are drawn, ranked by detector scores for points and by segment length for lines. Correct matches are drawn in yellow (points) and green (lines), and incorrect ones in red, where a match is deemed correct when its reprojection error (points) or orthogonal line distance (lines) under the ground-truth homography is below 3 pixels. \cref{fig:supp_point_matching} shows point correspondences and \cref{fig:supp_line_matching} shows line correspondences, including both favorable and challenging cases.

Notably, the same viewpoint-varying pairs (artwork, sculpture, and stained glass) that are matched reliably by points are almost entirely mismatched as lines. This follows from our design: \OURS\ has no dedicated line descriptor and matches segments only through the descriptors at their two endpoints~\cite{pautrat_suarez_2023_gluestick} (\cref{sec:experiments}). As these endpoints are LSD segment terminations rather than repeatable keypoints, a perspective change shifts them along the segment and alters their surrounding appearance, breaking the endpoint correspondences; the point branch is unaffected, as it matches repeatable keypoints directly. The simple mutual-nearest-neighbor matcher used here further limits robustness, and we expect a lightweight learned point-line matcher, which we leave to future work (\cref{sec:conclusion}), to improve these results.

\input{figures/supp_matching_ex}

%% file: tables/results_ablation.tex
\setcounter{table}{8}
\begin{table}[tb]
    \centering
    \scriptsize
    \caption{\textbf{Impact of joint point and line training on MegaDepth~\cite{Zheng2023} relative pose estimation.}}
    \setlength{\tabcolsep}{4pt}
    \renewcommand{\arraystretch}{1}
    \begin{tabular}{lccc}
    \toprule
    Method & AUC@5$^\circ$ $\uparrow$ & AUC@10$^\circ$ $\uparrow$ & AUC@20$^\circ$ $\uparrow$ \\
    \midrule
    Ours (points only) & 55.2 & 67.6 & 76.4 \\
    Ours (joint training) & \textbf{58.2} & \textbf{70.8} & \textbf{79.4} \\
    \midrule
    Improvement & +3.0 & +3.2 & +3.0 \\
    \bottomrule
    \end{tabular}
    \label{tab:joint_training_ablation}
\end{table}

%% file: tables/supp_evaluation_old_nvidia.tex
\setcounter{table}{9}
\begin{table}[h]
    \centering
    \scriptsize
    \setlength{\tabcolsep}{5pt}
    \renewcommand{\arraystretch}{1}
    \caption{\textbf{Runtime evaluation.} Parameter counts and latencies (ms) are averaged over 500 sequential detections on an NVIDIA GeForce GTX 1050 Ti with 4 GB of VRAM. \OURS\ runs in 186 ms, about 18$\times$ faster than DaD + ScaleLSD.} 
    \begin{tabular}{lccc}
    \toprule
      & Nb params $\downarrow$ & Latency $\downarrow$ \\ %& Throughput $\uparrow$ \\
     \midrule
     SuperPoint~\cite{DeTone2018} + DeepLSD~\cite{Pautrat_2023_DeepLSD} & $9.8 \times 10^6$ & 
     707 \\
     ALIKED~\cite{Zhao2023} + DeepLSD~\cite{Pautrat_2023_DeepLSD} & $9.2 \times 10^6$ & 716 \\ %& 3.6 im/sec \\ 
    %ALIKED~\cite{Zhao2023} + M-LSD~\cite{Gu2021} & $1.3\times 10^6$ & 193ms  \\
     %ALIKED~\cite{Zhao2023} + TP-LSD~\cite{Huang2020} & $2.5\times 10^7$ & 878ms  \\
     DaD~\cite{edstedt2025dad} + DeDoDe v2~\cite{edstedt2024dedodev2} + ScaleLSD~\cite{ScaleLSD} & $1.2\times 10^8$ & 3474 \\ %& 7.2 im/sec \\ 
     %Wireframe~\cite{ferre2025wireframe} & $5.8\times 10^6$ & 653ms \\ %& 5.5 im/sec \\ 
    %PLNet~\cite{xu2024airslam} & $7.5\times 10^6$ & 824 ms \\
     \OURS~(Ours) & \textbf{$7.8\times10^5$} & \textbf{186} \\ %& \textbf{11.9 im/sec} \\ 
     \bottomrule
     \end{tabular}
     \label{tab:old_nvidia}
\end{table}

%% file: tables/supp_localization.tex
\begin{sidewaystable}[ht]
\centering
\tiny
\caption{\textbf{Visual localization on the full 7Scenes dataset~\cite{7scenes}.} We report the median translation / rotation errors (cm / deg) and pose accuracy (\%) at 5cm / 5° threshold for each scene. We denote the best method for each scene and category (Points or Points+Lines) in \textbf{bold} and the second-best method is \underline{underlined}.}
\setlength{\tabcolsep}{1.5pt}
\begin{tabular}{llcccccccccccccccc}
\toprule
& & \multicolumn{2}{c}{\textbf{Chess}} & \multicolumn{2}{c}{\textbf{Fire}} & \multicolumn{2}{c}{\textbf{Heads}} & \multicolumn{2}{c}{\textbf{Office}} & \multicolumn{2}{c}{\textbf{Pumpkin}} & \multicolumn{2}{c}{\textbf{RedKitchen}} & \multicolumn{2}{c}{\textbf{Stairs}} & \multicolumn{2}{c}{\textbf{Total}} \\
\cmidrule(lr){3-4} \cmidrule(lr){5-6} \cmidrule(lr){7-8} \cmidrule(lr){9-10} \cmidrule(lr){11-12} \cmidrule(lr){13-14} \cmidrule(lr){15-16} \cmidrule(lr){17-18}
\textbf{} & Method & T/R $\downarrow$ & Acc. $\uparrow$ & T/R $\downarrow$ & Acc. $\uparrow$ & T/R $\downarrow$ & Acc. $\uparrow$ & T/R $\downarrow$ & Acc. $\uparrow$ & T/R $\downarrow$ & Acc. $\uparrow$ & T/R $\downarrow$ & Acc. $\uparrow$ & T/R $\downarrow$ & Acc. $\uparrow$ & T/R $\downarrow$ & Acc. $\uparrow$ \\
\midrule
\multirow{6}{*}{\rotatebox{90}{\textbf{Points}}}
& SuperPoint~\cite{DeTone2018} & \snd{2.5}/\snd{0.87} & \snd{92.2} & \snd{2.3}/\snd{0.90} & \snd{89.5} & \snd{1.1}/0.80 & \fst{95.6} & \snd{3.2}/\fst{0.93} & \snd{74.4} & \snd{5.2}/\snd{1.40} & \snd{48.2} & \snd{4.4}/\snd{1.43} & \snd{56.7} & \snd{6.2}/\snd{1.65} & 36.6 & 3.56/\snd{1.14} & \snd{70.5} \\
& DISK~\cite{tyszkiewicz2020disk} & \fst{2.4}/\snd{0.87} & \snd{92.2} & 2.4/0.96 & 86.4 & 1.2/0.86 & 89.9 & 3.8/1.06 & 64.9 & 5.4/1.47 & 46.0 & 4.6/1.49 & 55.2 & 7.2/2.16 & 28.5 & 3.86/1.27 & 66.2 \\
& ALIKED~\cite{Zhao2023} & 2.6/0.92 & 86.3 & 2.7/1.06 & 80.3 & 1.4/1.00 & 77.3 & 5.7/1.71 & 45.8 & 5.3/1.44 & 47.1 & 5.2/1.65 & 48.0 & 7.8/1.98 & 35.0 & 4.39/1.39 & 60.0 \\
%& DaD~\cite{edstedt2025dad}+DeDoDe v2~\cite{edstedt2024dedodev2} & \snd{2.5}/0.88 & 91.6 & \snd{2.3}/0.91 & 86.3 & 1.3/0.91 & 86.9 & 3.9/1.09 & 65.8 & 5.3/1.50 & 46.6 & 4.5/1.48 & 55.1 & 13.8/3.16 & 14.6 & 4.80/1.42 & 63.8 \\
& PLNet~\cite{xu2024airslam} & \snd{2.5}/\fst{0.86} & \snd{92.2} & \snd{2.3}/0.91 & 88.6 & \snd{1.1}/\snd{0.79} & \snd{95.2} & \fst{3.1}/\fst{0.93} & \fst{74.9} & \snd{5.2}/1.44 & 47.5 & \snd{4.4}/\fst{1.42} & 56.6 & \snd{6.2}/1.68 & \snd{38.2} & \snd{3.54}/1.15 & \snd{70.5} \\
& \OURS~(Ours) & \snd{2.5}/0.88 & \fst{93.4} & \fst{2.1}/\fst{0.85} & \fst{91.9} & \fst{1.0}/\fst{0.76} & 93.9 & 3.3/\snd{0.96} & 72.9 & \fst{4.9}/\fst{1.25} & \fst{51.2} & \fst{4.3}/1.44 & \fst{58.0} & \fst{5.1}/\fst{1.39} & \fst{49.1} & \textbf{3.31/1.08} & \textbf{72.9} \\
\midrule
\multirow{8}{*}{\rotatebox{90}{\textbf{Points+Lines}}}
& SuperPoint~\cite{DeTone2018}+DeepLSD~\cite{Pautrat_2023_DeepLSD} & \fst{2.5}/\snd{0.87} & \fst{92.5} & \snd{2.1}/0.85 & 93.2 & \snd{1.1}/\snd{0.79} & \snd{93.9} & \fst{3.1}/\fst{0.91} & \snd{76.0} & 4.9/\snd{1.31} & 51.4 & \snd{4.3}/1.43 & 57.5 & 5.0/1.33 & 49.6 & \snd{3.29}/\snd{1.07} & \snd{73.4} \\
& ALIKED~\cite{Zhao2023}+DeepLSD~\cite{Pautrat_2023_DeepLSD} & \snd{2.6}/0.90 & 88.0 & 2.4/0.95 & 85.2 & 1.4/1.00 & 74.5 & 5.3/1.62 & 47.8 & 5.0/1.38 & 49.8 & 5.1/1.61 & 49.1 & 5.1/1.48 & 49.6 & 3.84/1.28 & 63.4 \\
& ALIKED~\cite{Zhao2023}+TP-LSD~\cite{huang2020tplsd} & \snd{2.6}/0.89 & 88.2 & 2.4/0.96 & 84.3 & 1.4/0.98 & 74.3 & 5.3/1.64 & 48.1 & 5.2/1.38 & 48.5 & 5.1/1.61 & 48.8 & 5.2/1.48 & 48.8 & 3.89/1.28 & 63.0 \\
& ALIKED~\cite{Zhao2023}+M-LSD~\cite{Gu2021} & \snd{2.6}/0.89 & 88.2 & 2.5/0.96 & 84.2 & 1.4/0.99 & 73.9 & 5.3/1.62 & 47.6 & 5.2/1.37 & 48.5 & 5.1/1.6 & 48.9 & 5.2/1.52 & 48.5 & 3.90/1.28 & 62.8 \\
% & SuperPoint~\cite{DeTone2018}+ScaleLSD~\cite{ScaleLSD} & \fst{2.5}/0.88 & \fst{92.7} & \snd{2.1}/0.84 & \snd{93.8} & \snd{1.1}/0.81 & \snd{93.3} & \fst{3.1}/\fst{0.91} & \fst{76.0} & 4.9/\snd{1.31} & 51.7 & \snd{4.3}/\fst{1.41} & \snd{57.5} & 5.1/1.33 & 49.5 & -- & -- \\
% & ALIKED~\cite{Zhao2023}+ScaleLSD~\cite{ScaleLSD} & 2.6/0.9 & 87.9 & 2.4/0.95 & 84.8 & 1.4/1.00 & 74.1 & 5.3/1.64 & 47.6 & 5.0/1.38 & 49.9 & 5.1/1.62 & 49.0 & 5.1/1.50 & 49.4 & -- & -- \\
%& DaD~\cite{edstedt2025dad}+ScaleLSD~\cite{ScaleLSD} & \fst{2.5}/\snd{0.87} & 92.0 & \fst{2.0}/\snd{0.82} & \snd{93.5} & \snd{1.1}/0.84 & 88.5 & 3.4/1.00 & 71.1 & \snd{4.8}/1.35 & \snd{52.1} & \snd{4.3}/1.45 & 57.4 & 5.0/1.31 & 49.8 & 3.30/1.09 & 72.1 \\
& Wireframe~\cite{ferre2025wireframe} & \fst{2.5}/\fst{0.86} & 91.9 & 2.2/0.87 & 90.4 & \snd{1.1}/\snd{0.79} & 89.4 & 3.5/0.98 & 70.8 & 5.1/1.40 & 49.3 & 4.5/1.49 & 55.6 & \snd{4.6}/\snd{1.30} & \snd{53.8} & 3.36/1.10 & 71.6 \\
& PLNet~\cite{xu2024airslam} & \fst{2.5}/0.89 & \snd{92.3} & \snd{2.1}/0.85 & 92.4 & \snd{1.1}/0.80 & \fst{94.0} & \fst{3.1}/\fst{0.91} & \fst{76.6} & 4.9/1.36 & 50.7 & \snd{4.3}/\snd{1.42} & \snd{57.8} & 5.0/1.36 & 50.1 & \snd{3.29}/1.08 & \snd{73.4} \\
& \OURS~(Ours) & \textbf{2.5}/0.88 & 92.2 & \fst{2.0}/\fst{0.79} & \fst{94.9} & \textbf{1.0/0.77} & 92.8 & \snd{3.2}/\snd{0.93} & 75.8 & \fst{4.5}/\fst{1.17} & \fst{54.8} & \fst{4.2}/\fst{1.41} & \fst{59.6} & \fst{4.4}/\fst{1.23} & \fst{54.6} & \textbf{3.11/1.03} & \textbf{75.0} \\
\bottomrule
\end{tabular}
\label{tab:visual_localization_7scenes_all}
\end{sidewaystable}

%% file: figures/supp_detection_ex.tex
\setcounter{figure}{4}
\begin{figure*}[tb]
    \centering
    \scriptsize
    \setlength{\tabcolsep}{2pt}
    \begin{tabular}{ccccc}
        LSD~\cite{Gioi2010} & ScaleLSD~\cite{ScaleLSD} & Wireframe~\cite{ferre2025wireframe} & DeepLSD~\cite{Pautrat_2023_DeepLSD} & \OURS \ (Ours) \\
        \includegraphics[width=0.19\linewidth]{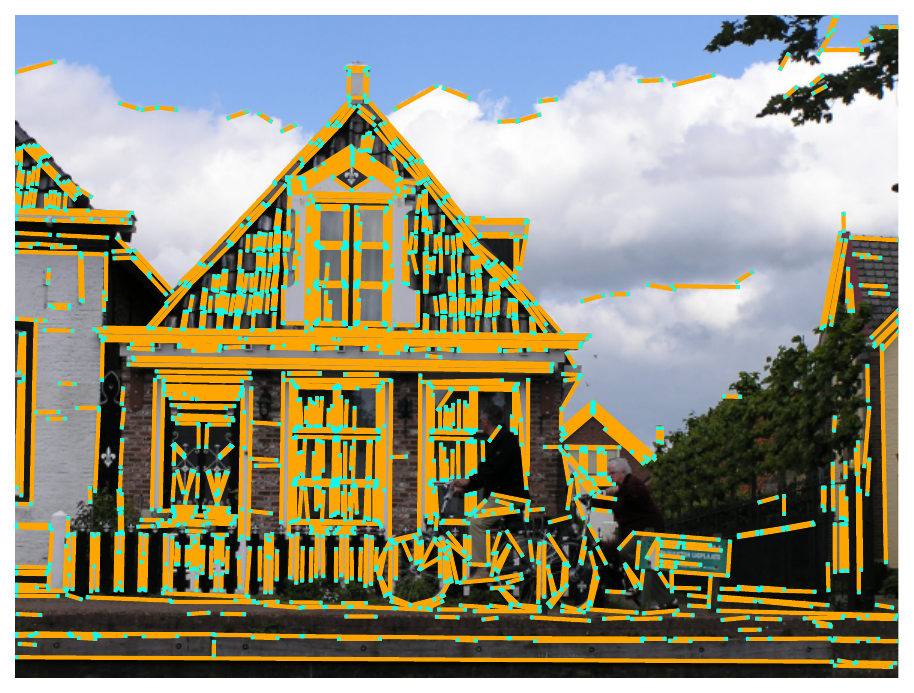} &
        \includegraphics[width=0.19\linewidth]{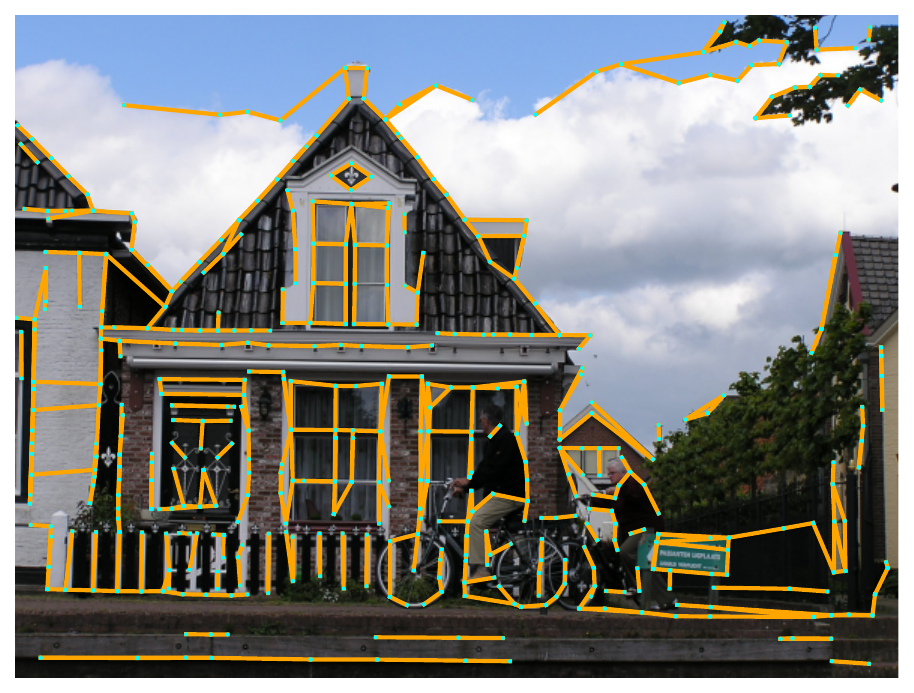} &
        \includegraphics[width=0.19\linewidth]{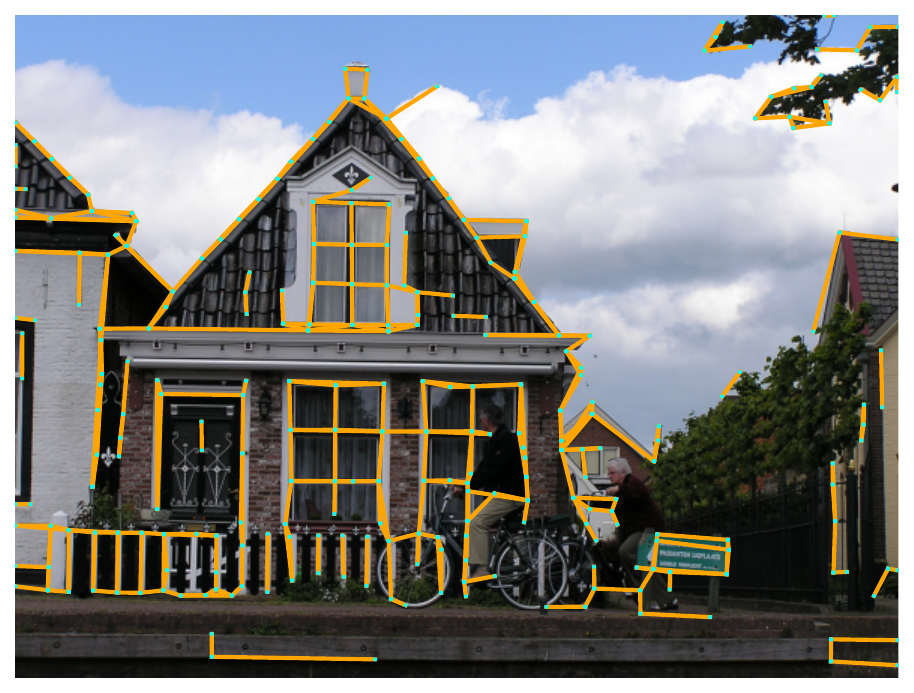} &
        \includegraphics[width=0.19\linewidth]{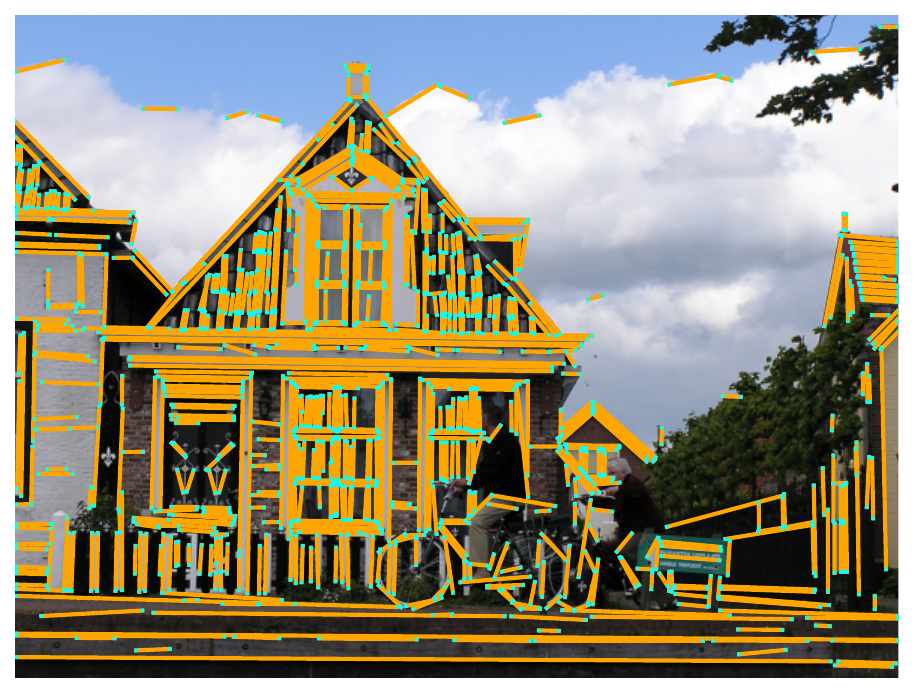} &
        \includegraphics[width=0.19\linewidth]{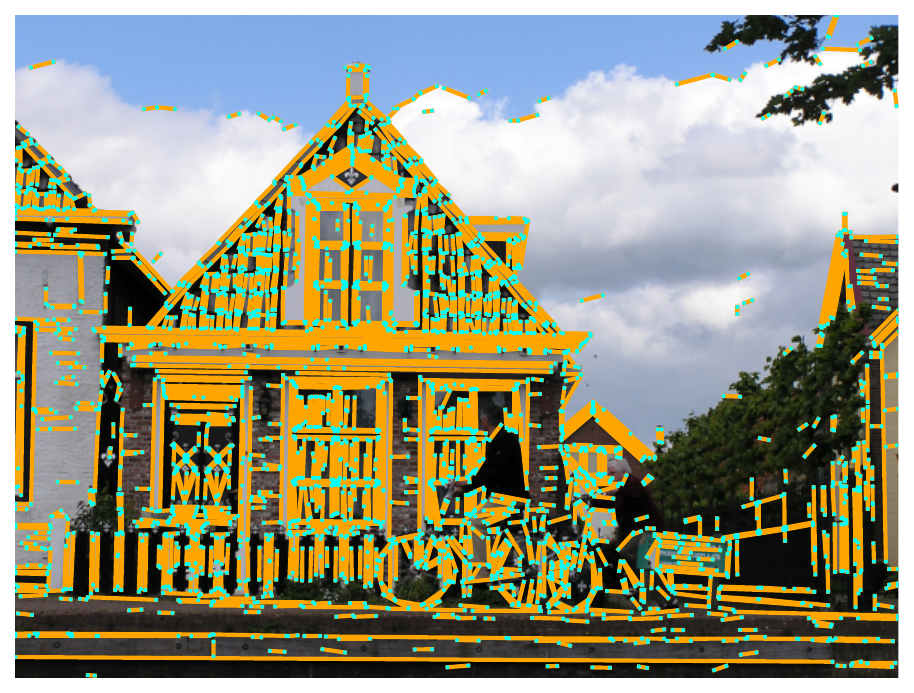}
        \\
        \includegraphics[width=0.19\linewidth]{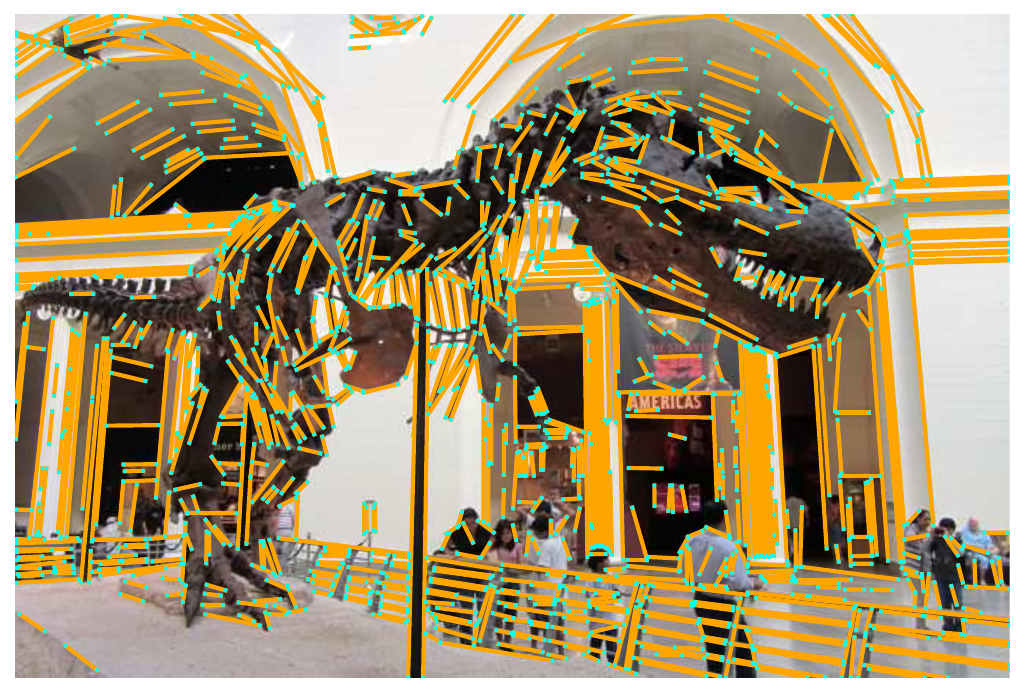} &
        \includegraphics[width=0.19\linewidth]{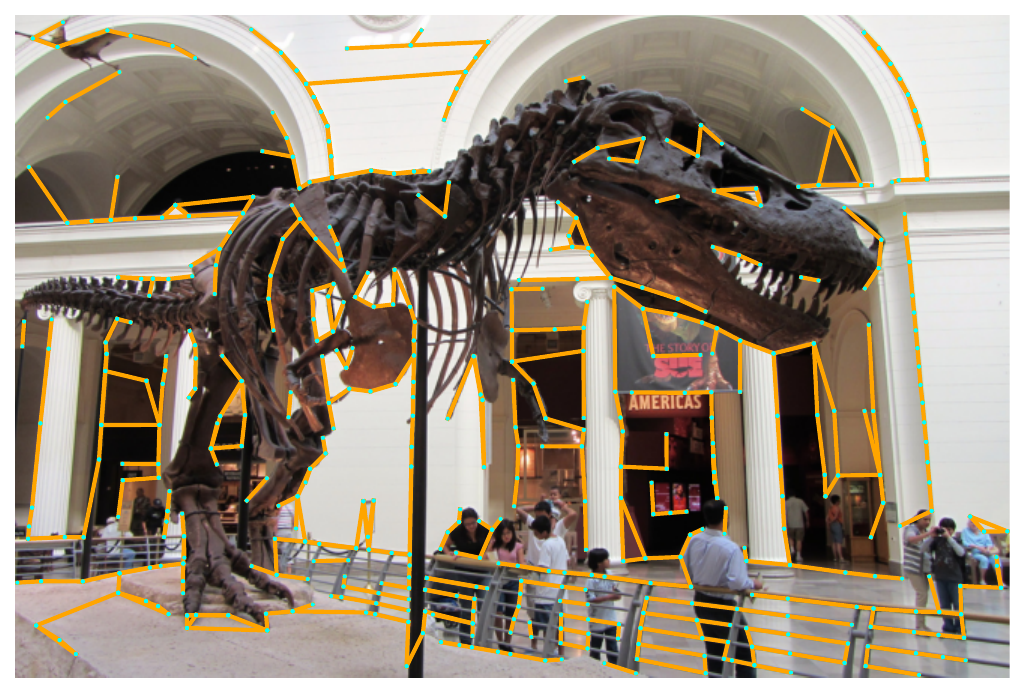} &
        \includegraphics[width=0.19\linewidth]{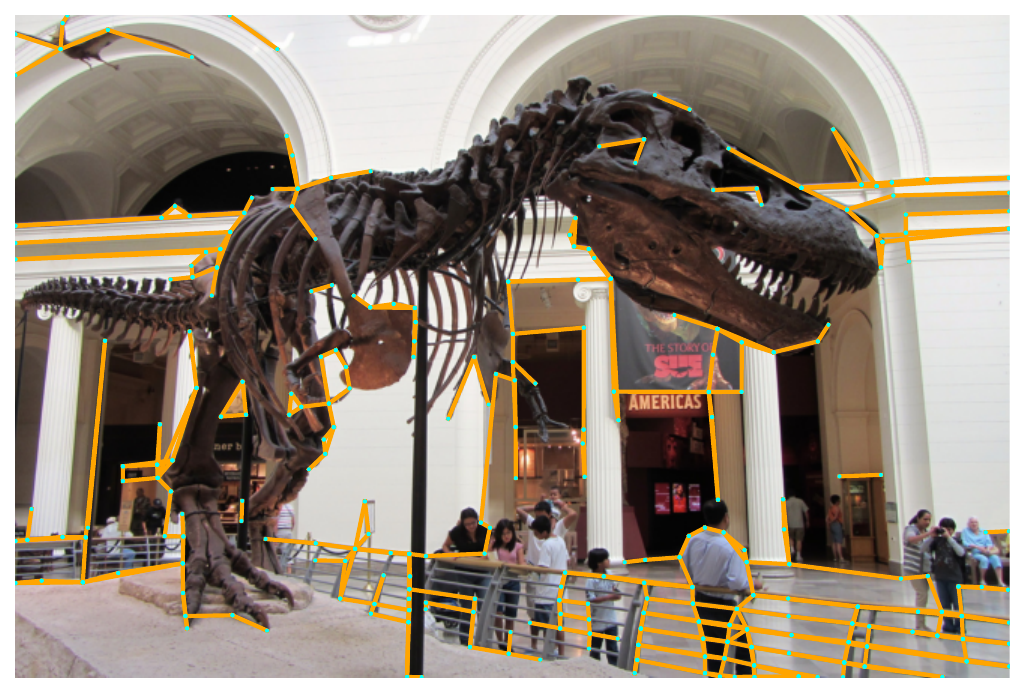} &
        \includegraphics[width=0.19\linewidth]{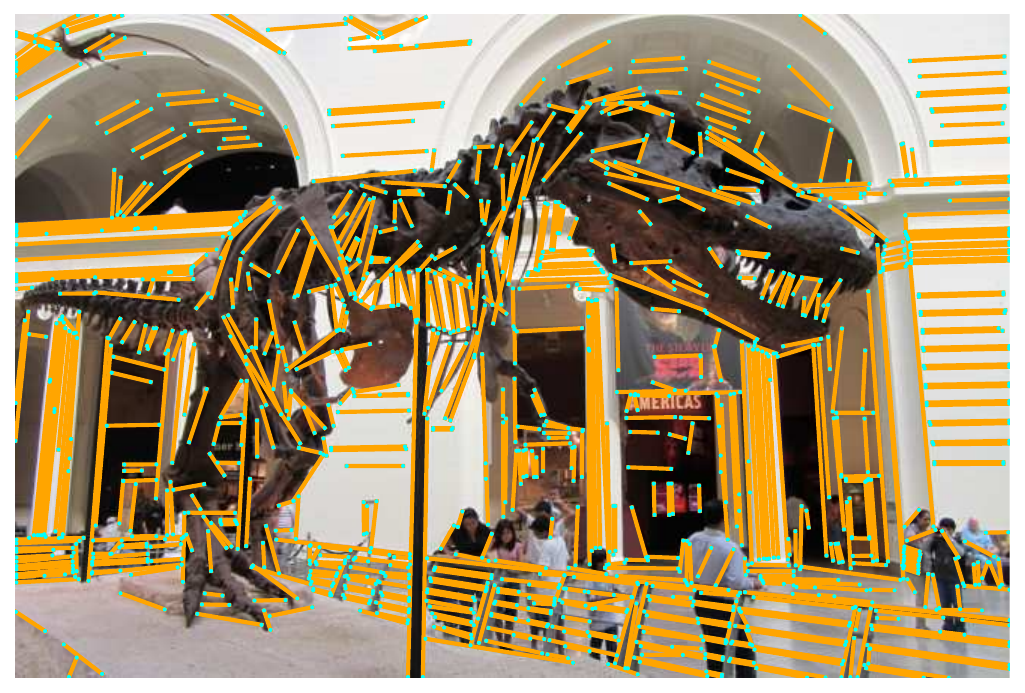} &
        \includegraphics[width=0.19\linewidth]{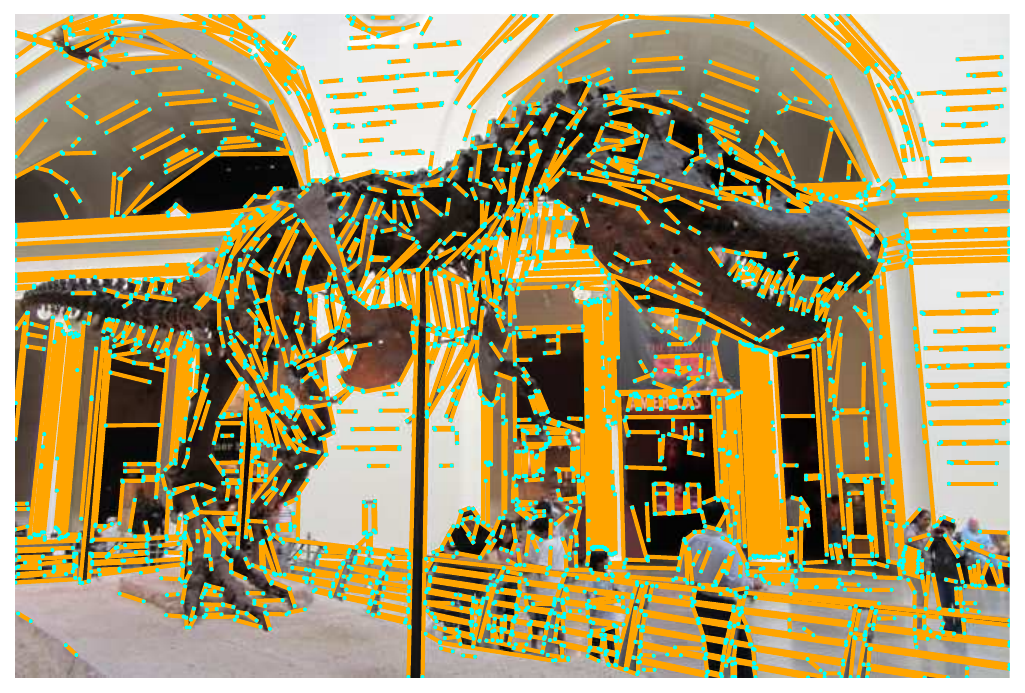}
        \\
        \includegraphics[width=0.19\linewidth]{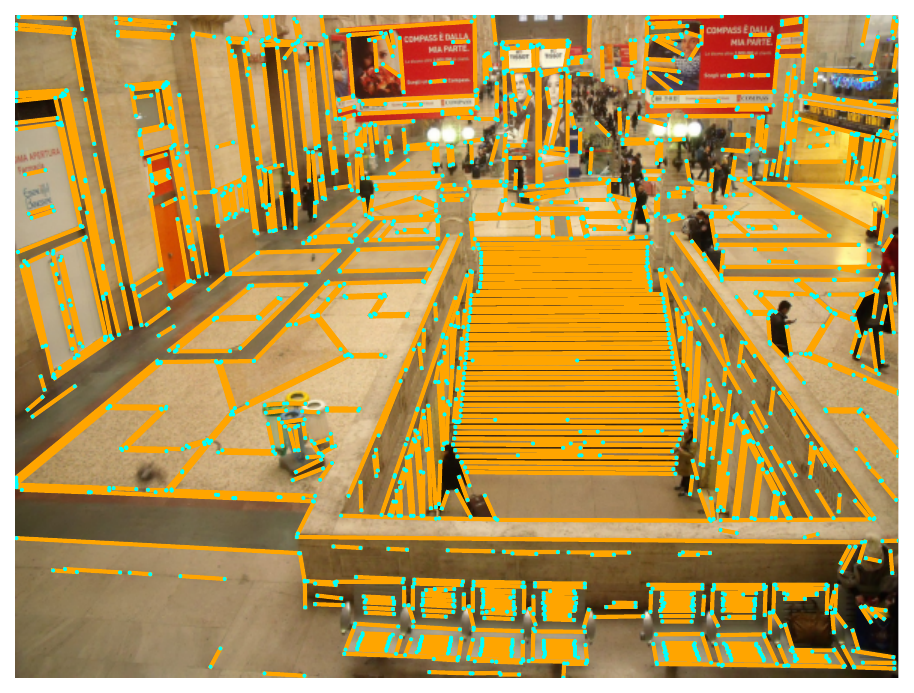} &
        \includegraphics[width=0.19\linewidth]{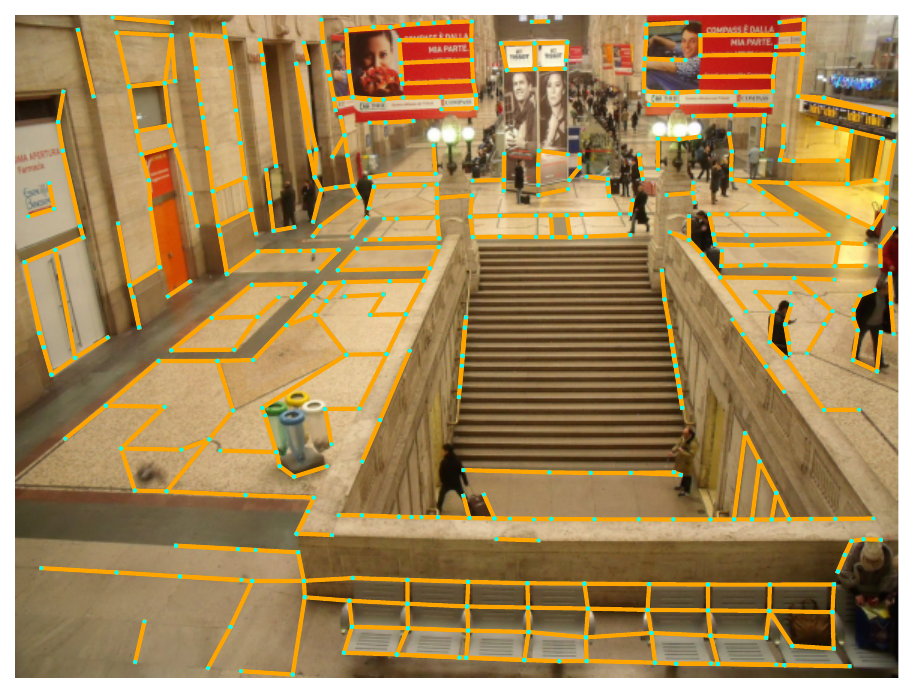} &
        \includegraphics[width=0.19\linewidth]{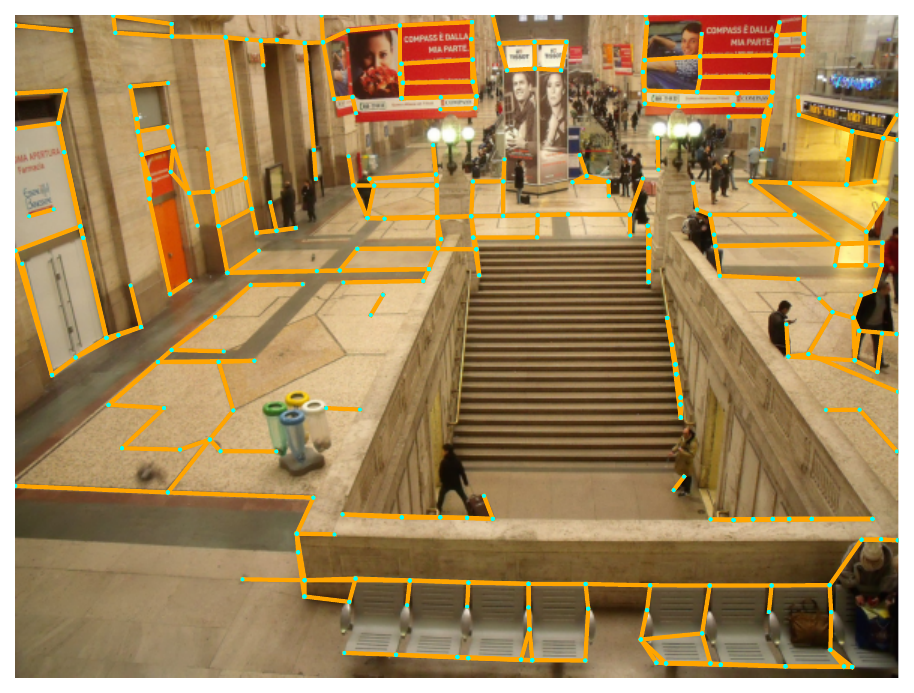} &
        \includegraphics[width=0.19\linewidth]{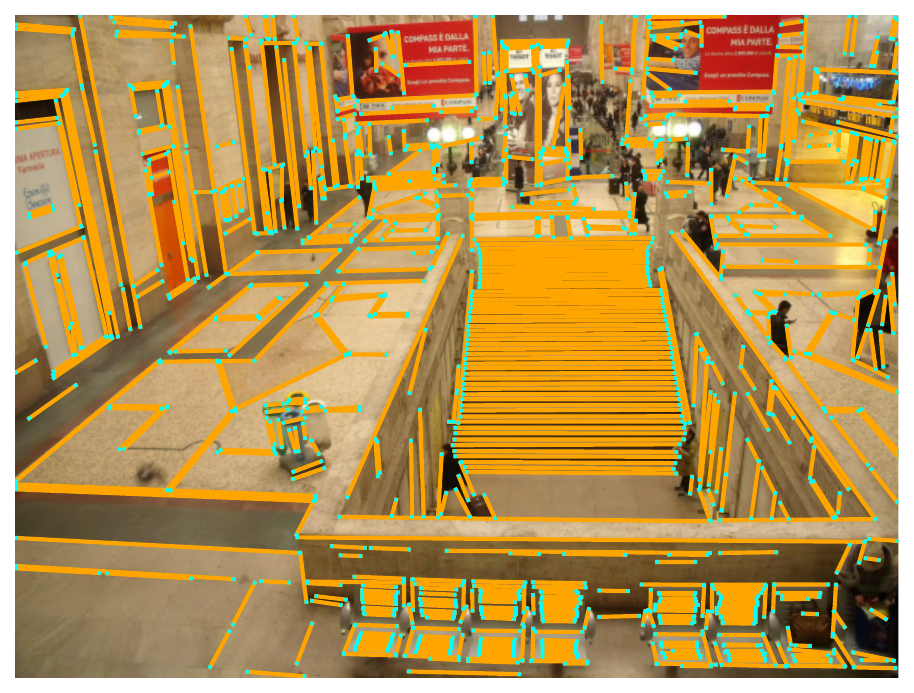} &
        \includegraphics[width=0.19\linewidth]{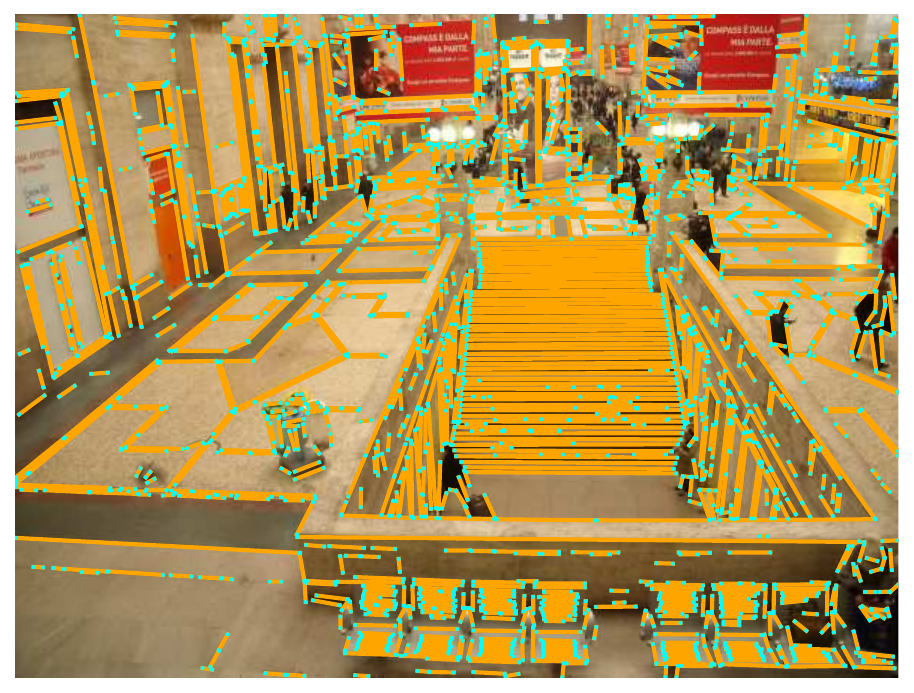}
        \\
        \includegraphics[width=0.19\linewidth]{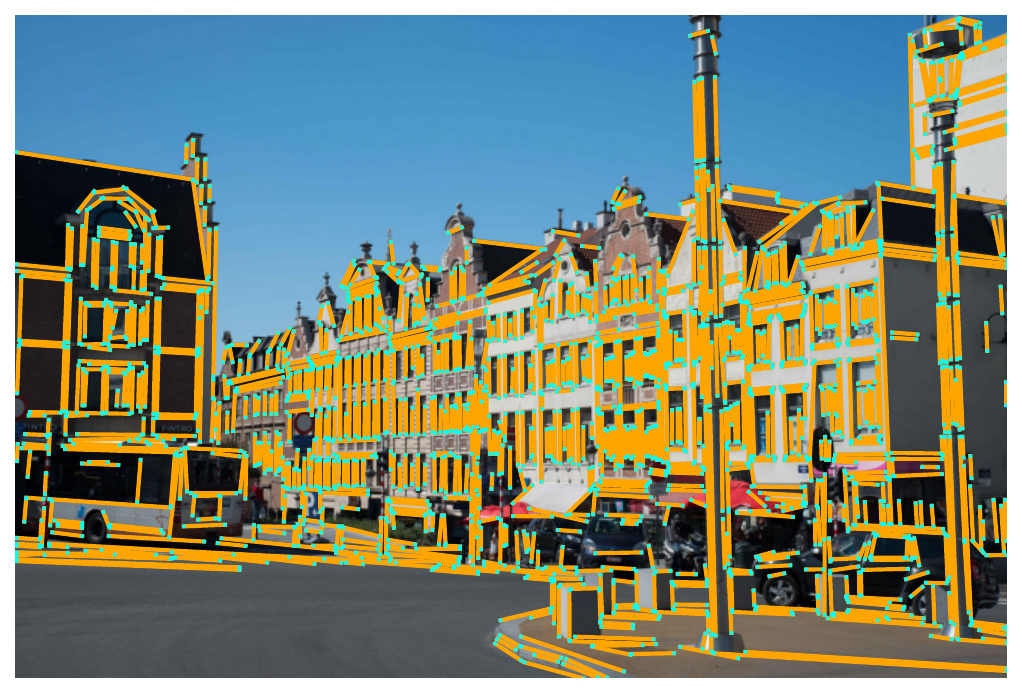} &
        \includegraphics[width=0.19\linewidth]{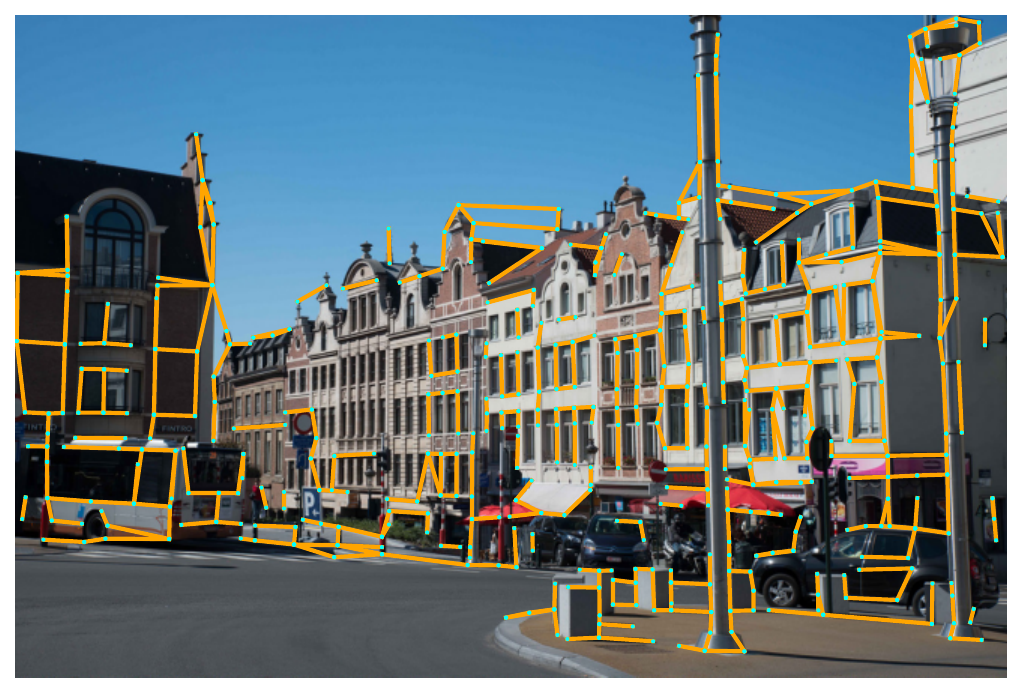} &
        \includegraphics[width=0.19\linewidth]{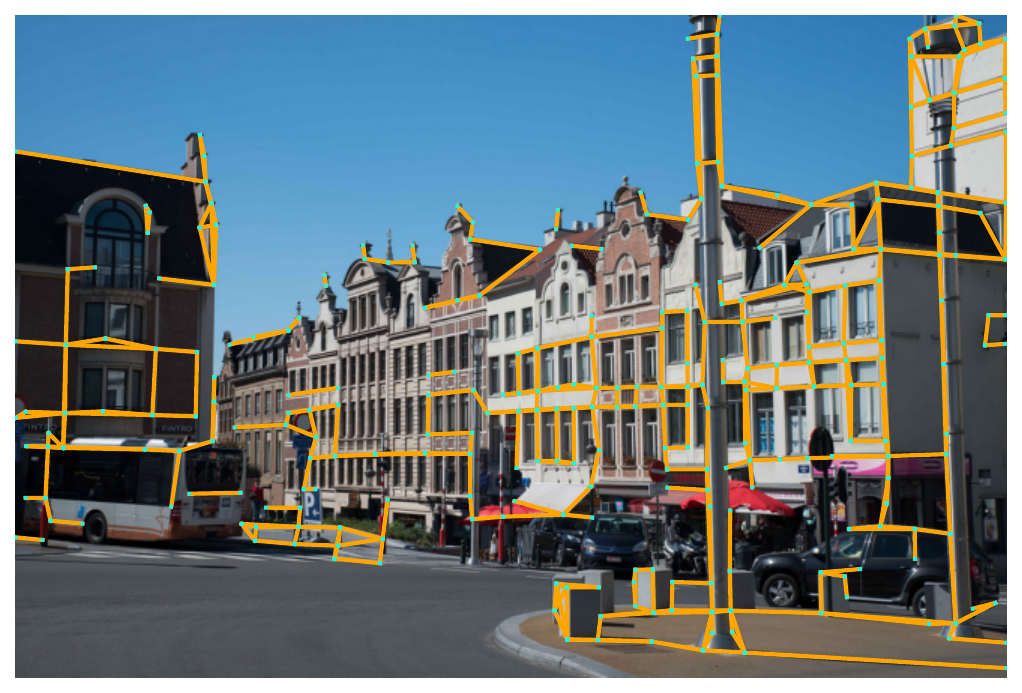} &
        \includegraphics[width=0.19\linewidth]{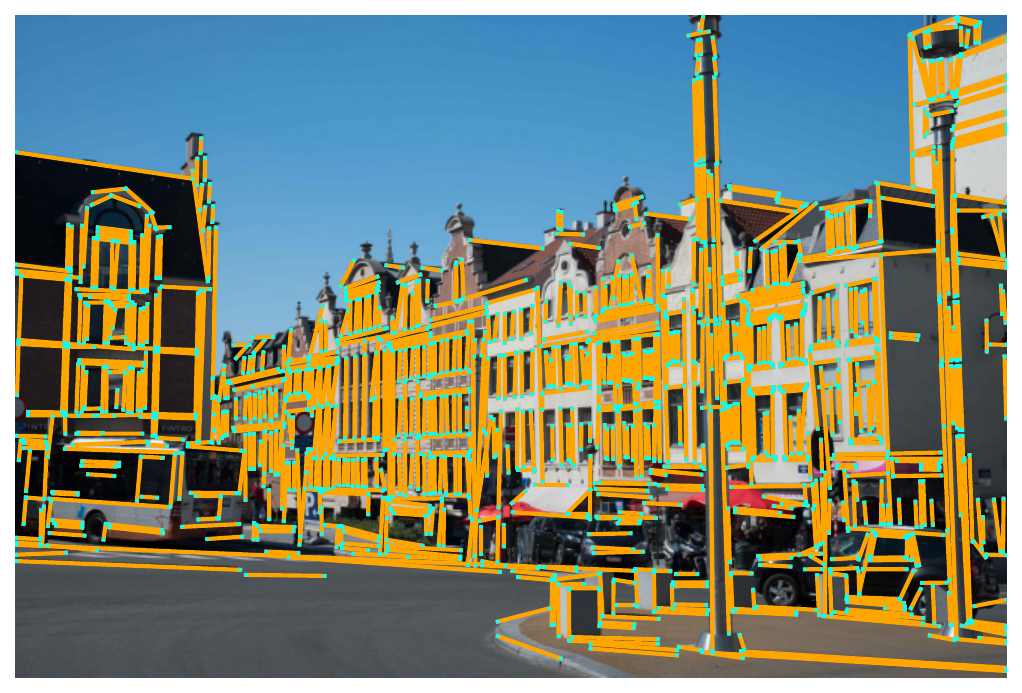} &
        \includegraphics[width=0.19\linewidth]{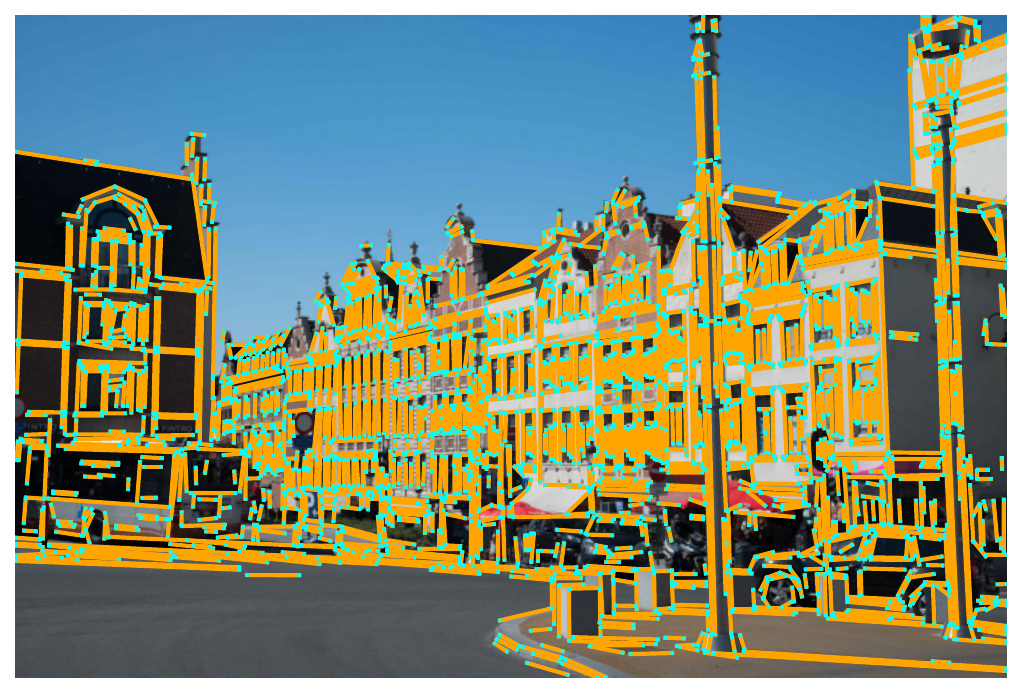}
        \\
        \includegraphics[width=0.19\linewidth]{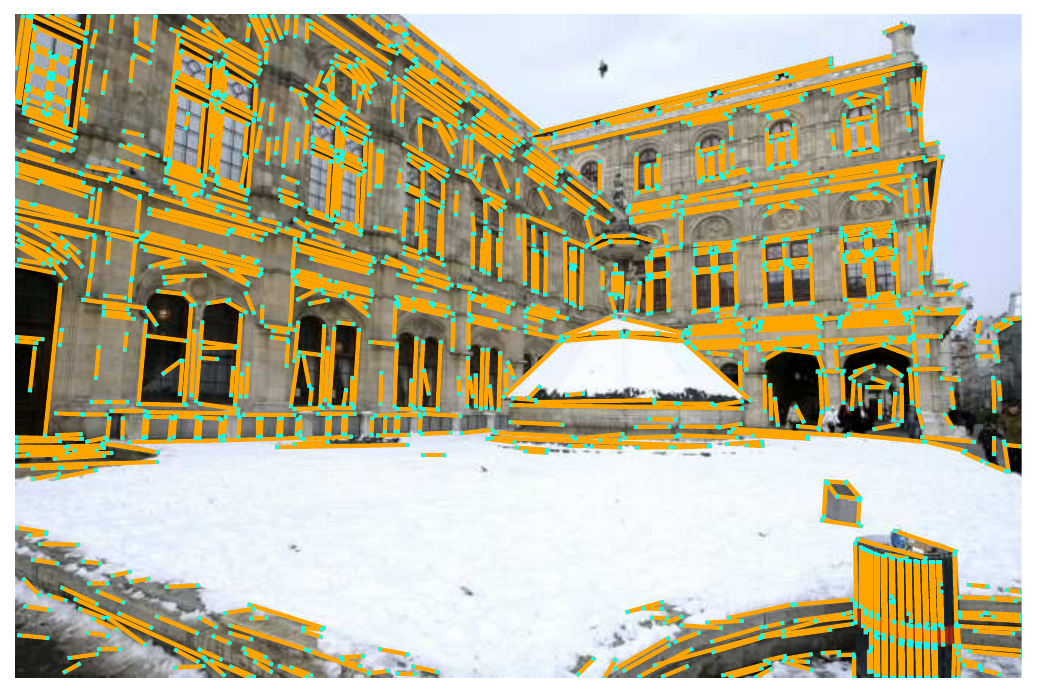} &
        \includegraphics[width=0.19\linewidth]{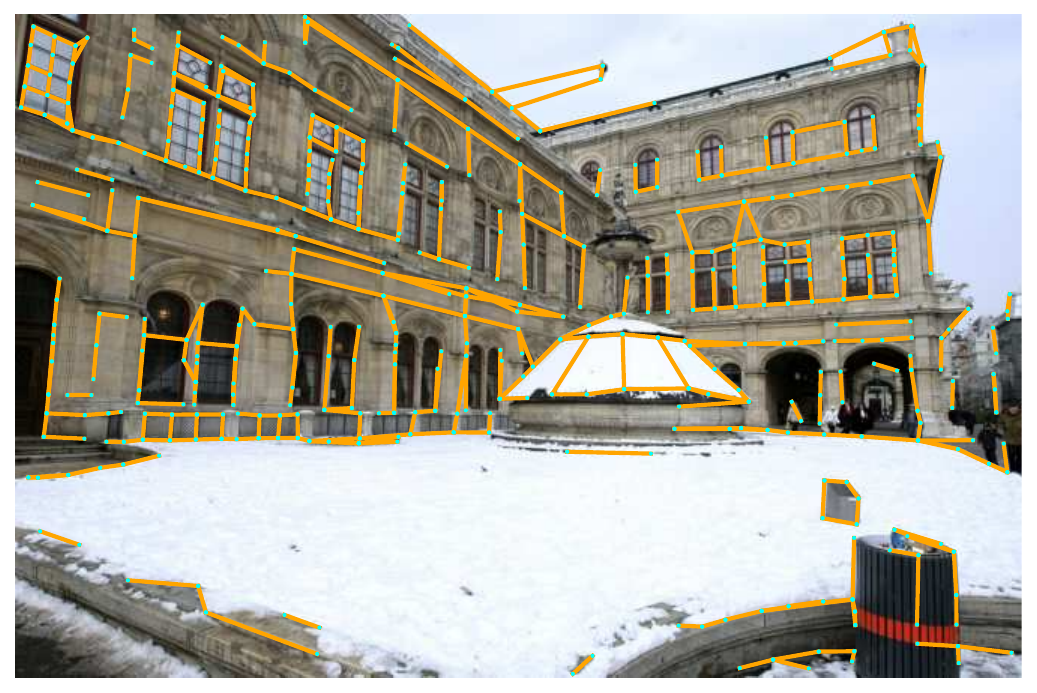} &
        \includegraphics[width=0.19\linewidth]{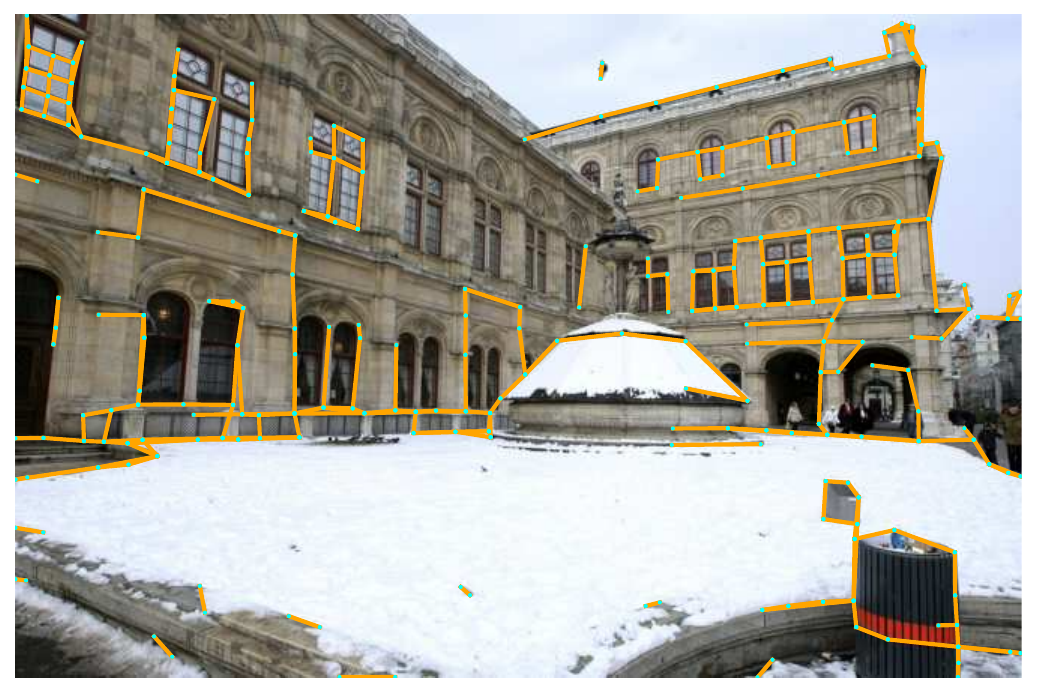} &
        \includegraphics[width=0.19\linewidth]{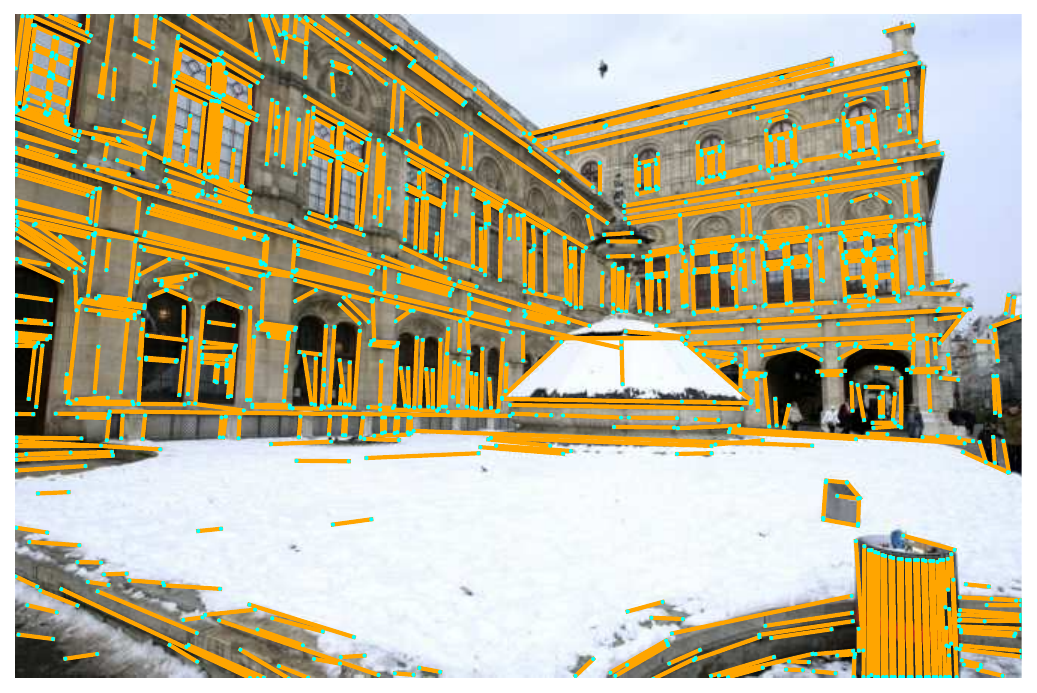} &
        \includegraphics[width=0.19\linewidth]{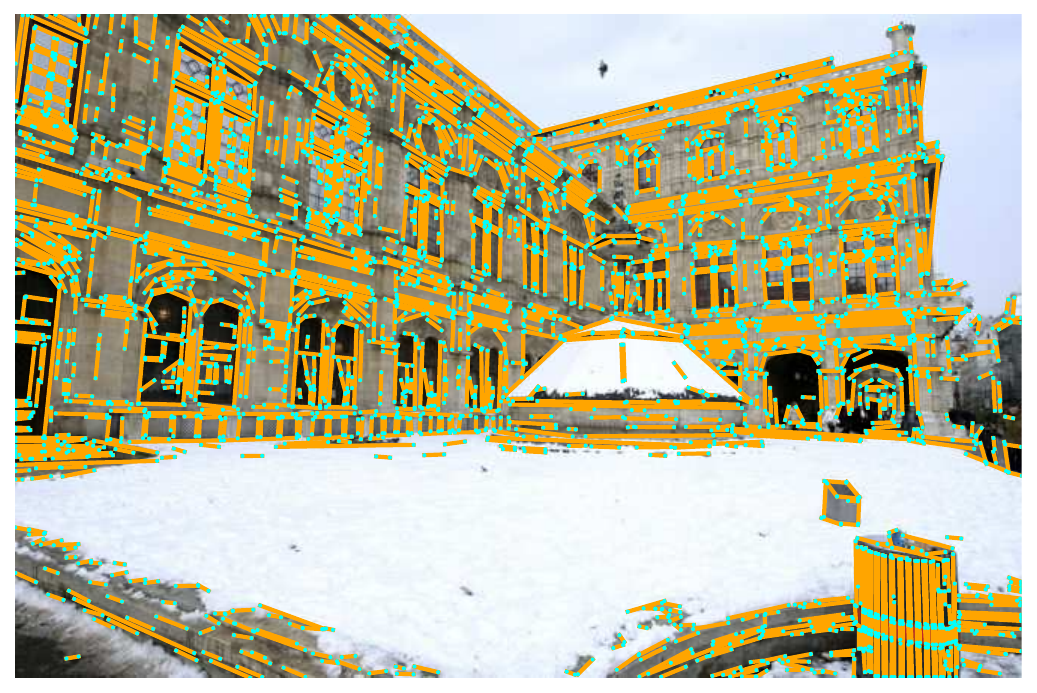}
        \\
        \includegraphics[width=0.19\linewidth]{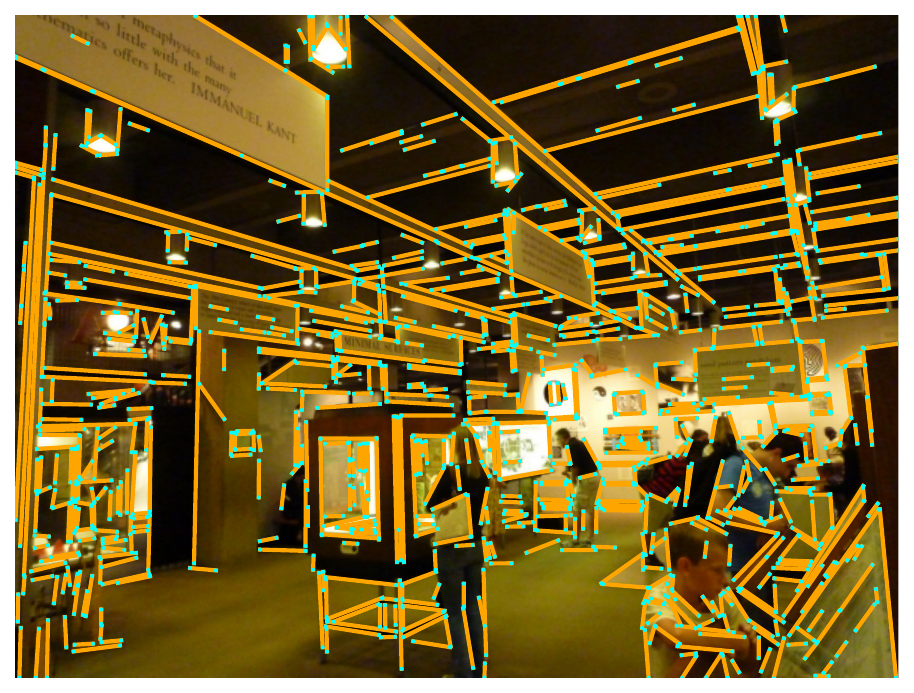} &
        \includegraphics[width=0.19\linewidth]{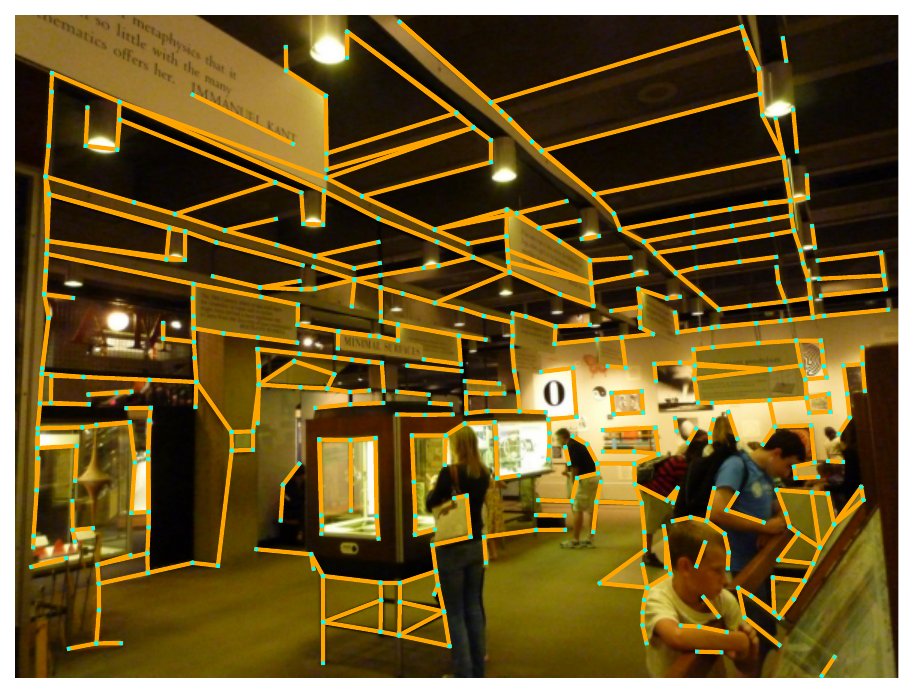} &
        \includegraphics[width=0.19\linewidth]{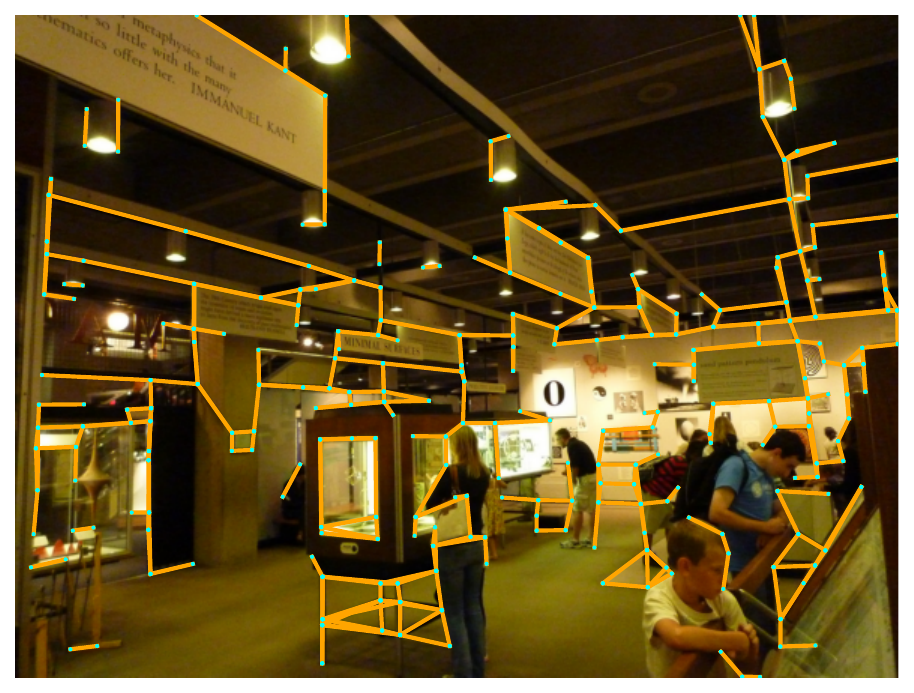} &
        \includegraphics[width=0.19\linewidth]{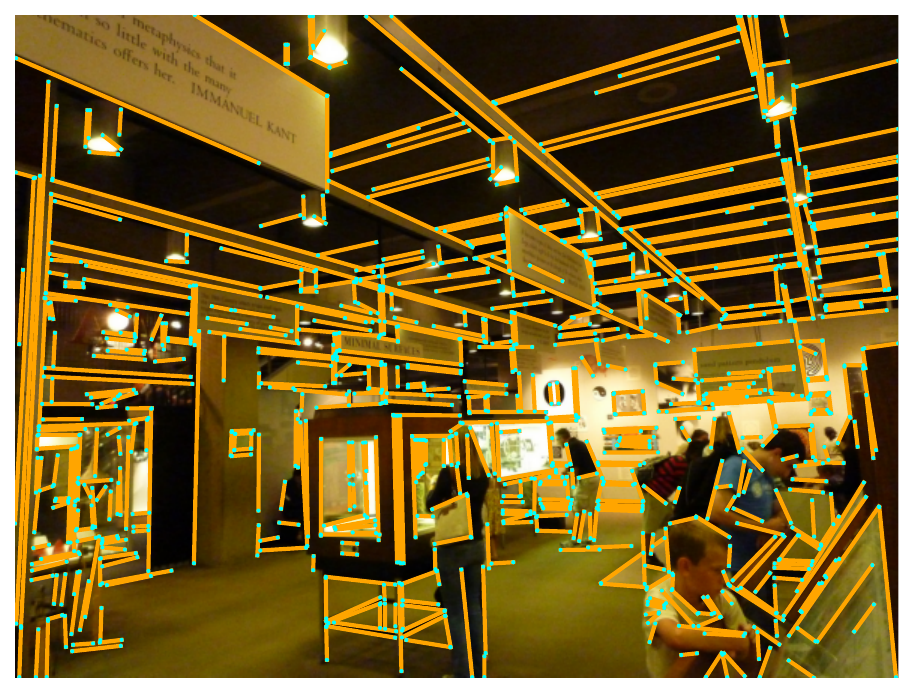} &
        \includegraphics[width=0.19\linewidth]{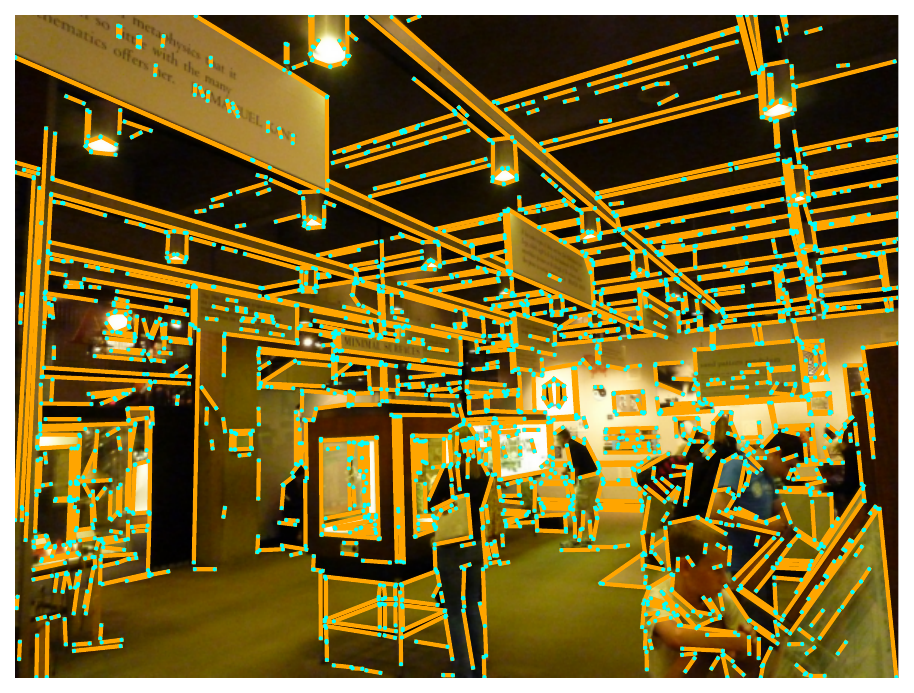}

        \\
        \includegraphics[width=0.19\linewidth]{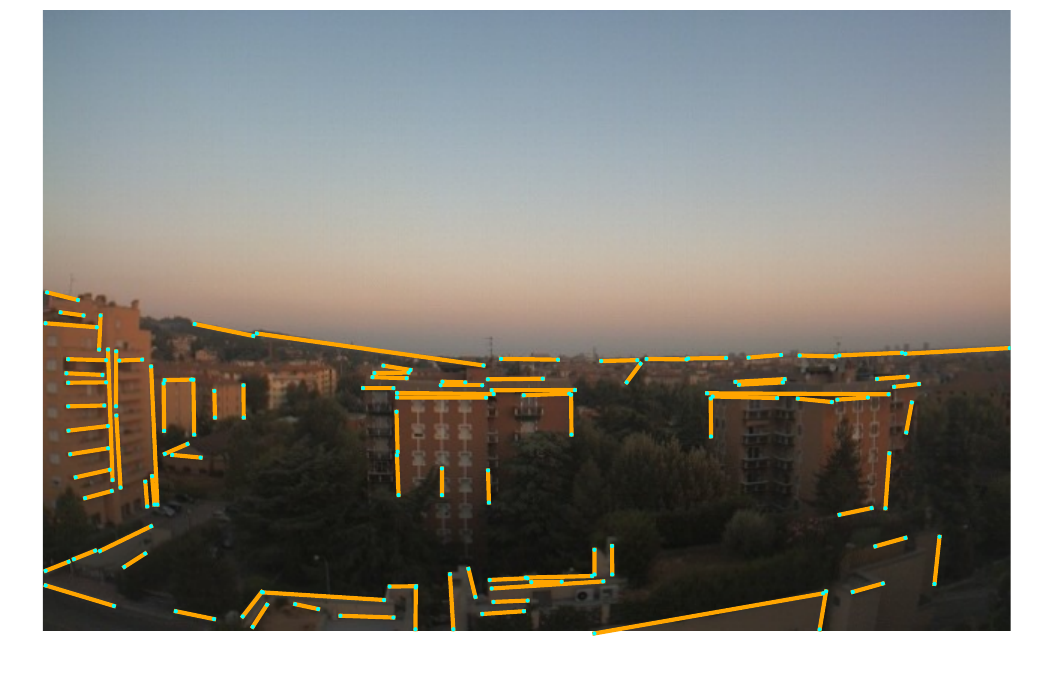} &
        \includegraphics[width=0.19\linewidth]{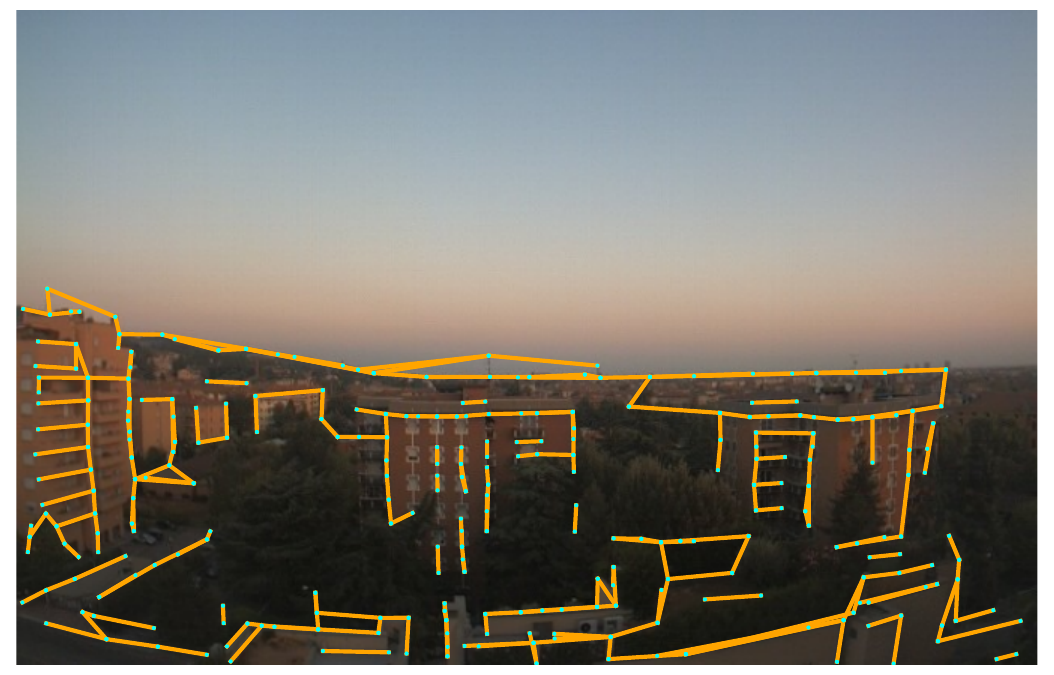} &
        \includegraphics[width=0.19\linewidth]{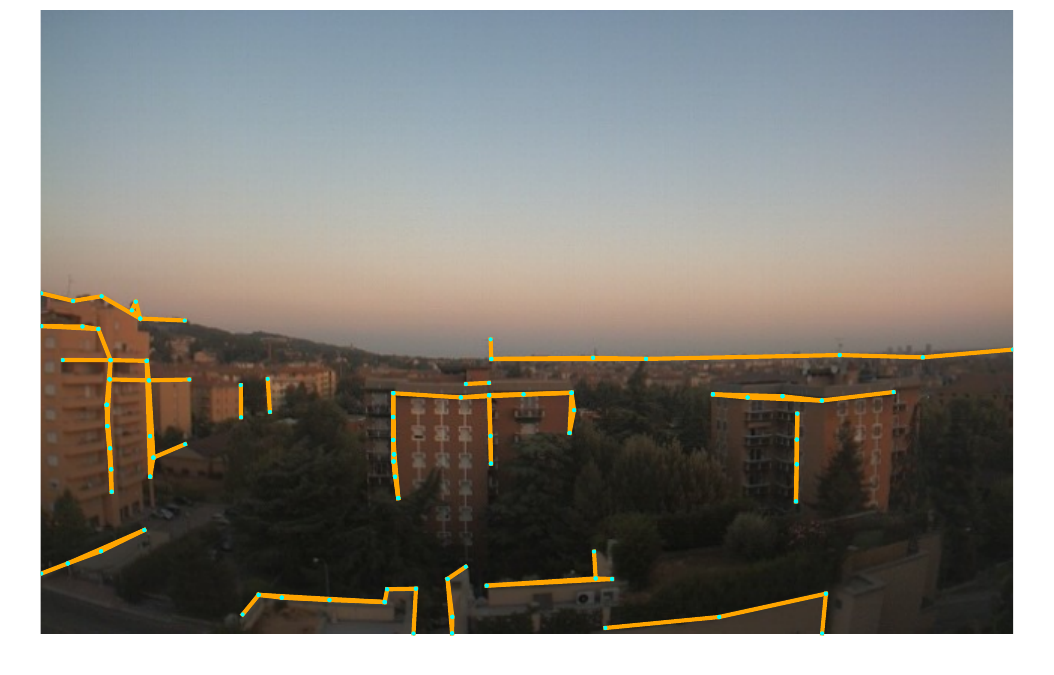} &
        \includegraphics[width=0.19\linewidth]{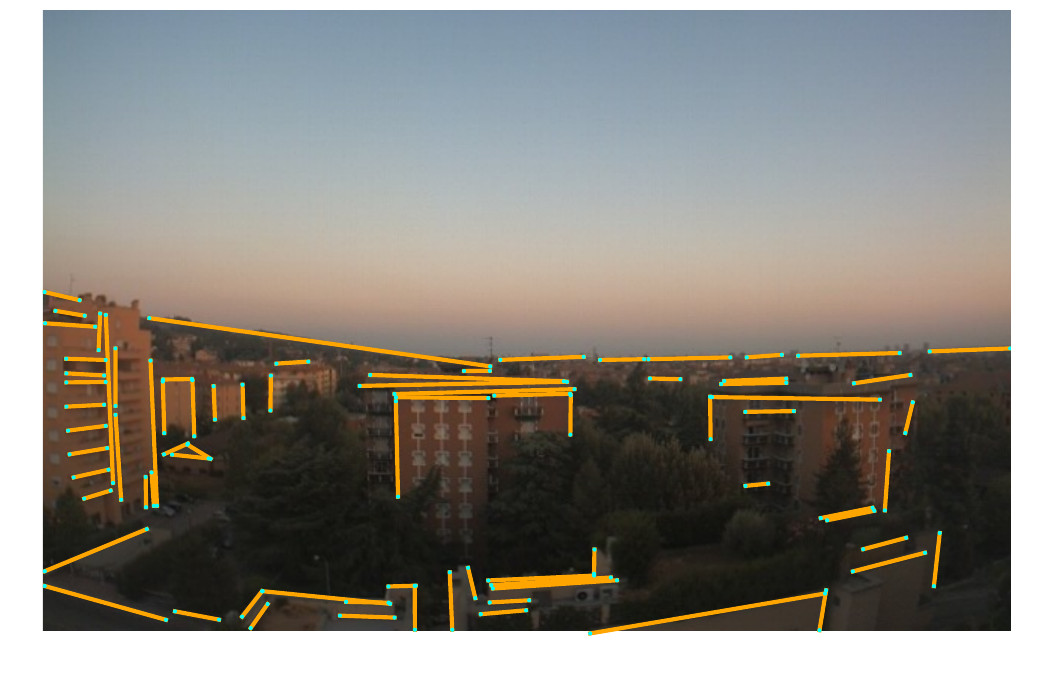} &
        \includegraphics[width=0.19\linewidth]{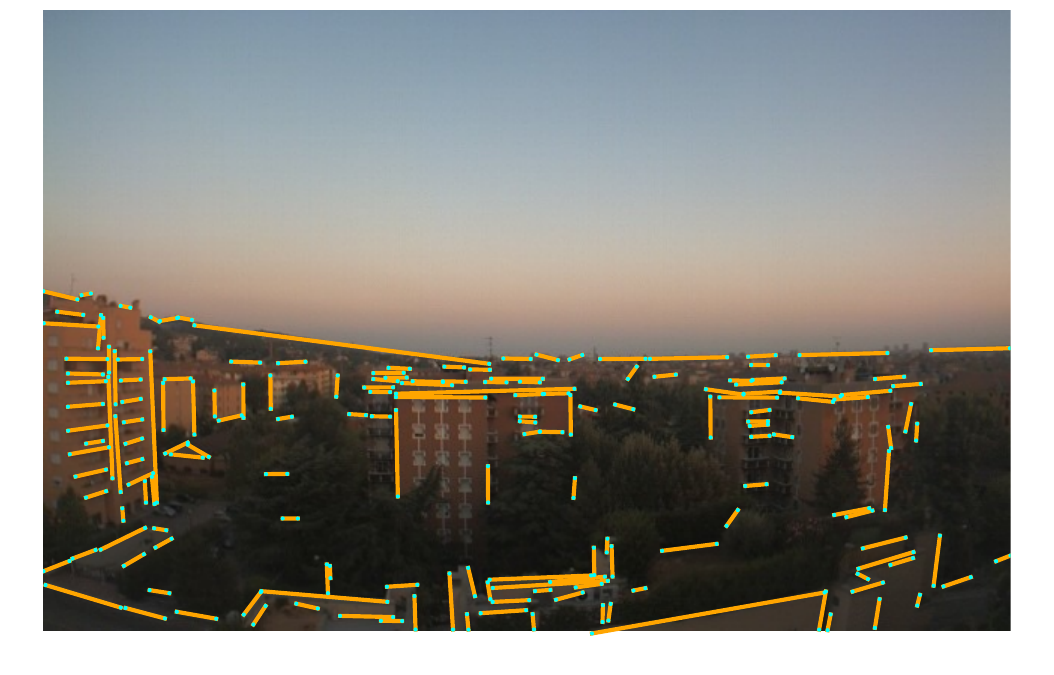}
        \\
        \includegraphics[width=0.19\linewidth]{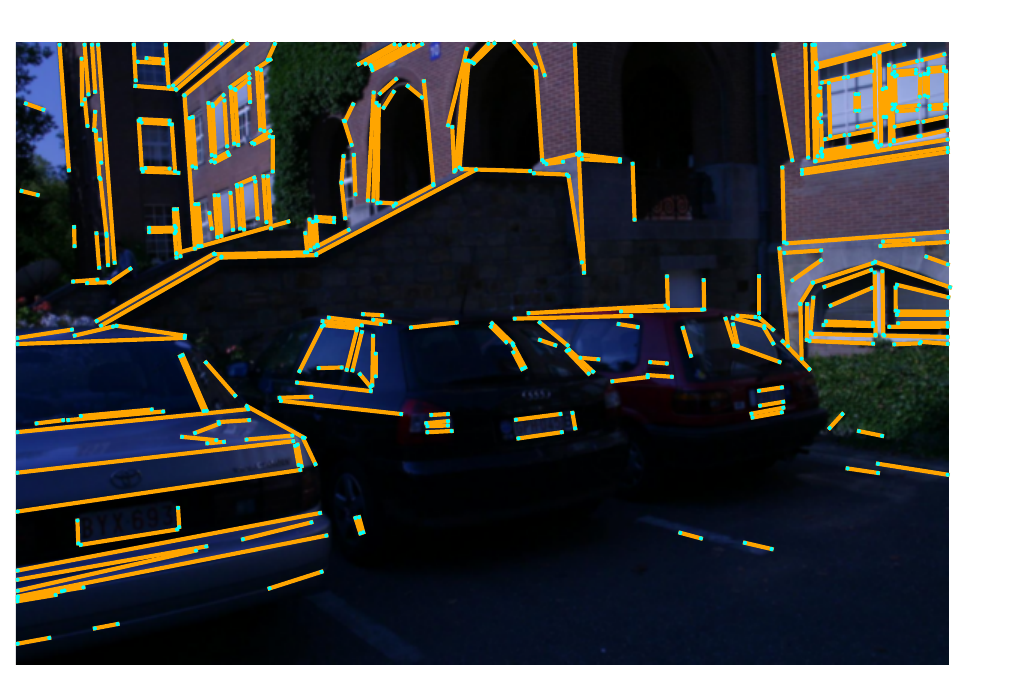} &
        \includegraphics[width=0.19\linewidth]{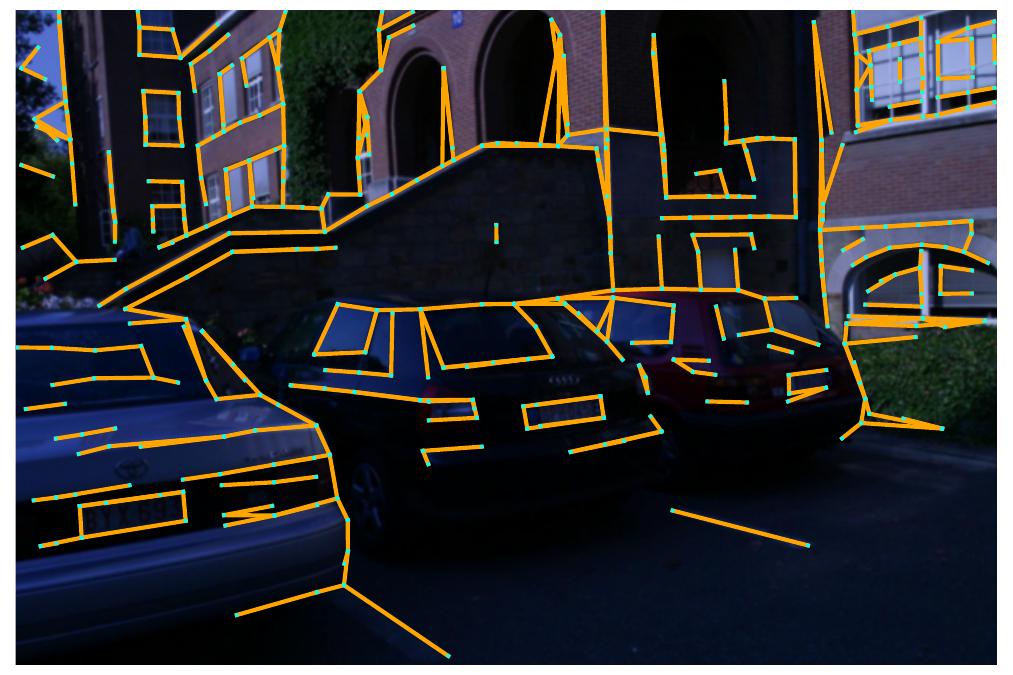} &
        \includegraphics[width=0.19\linewidth]{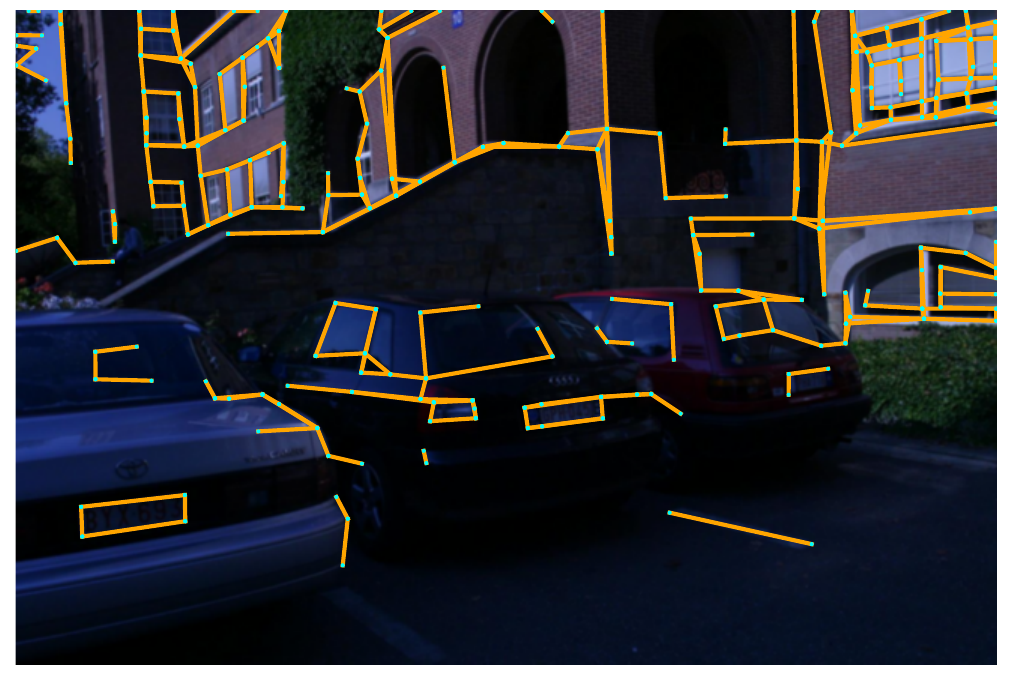} &
        \includegraphics[width=0.19\linewidth]{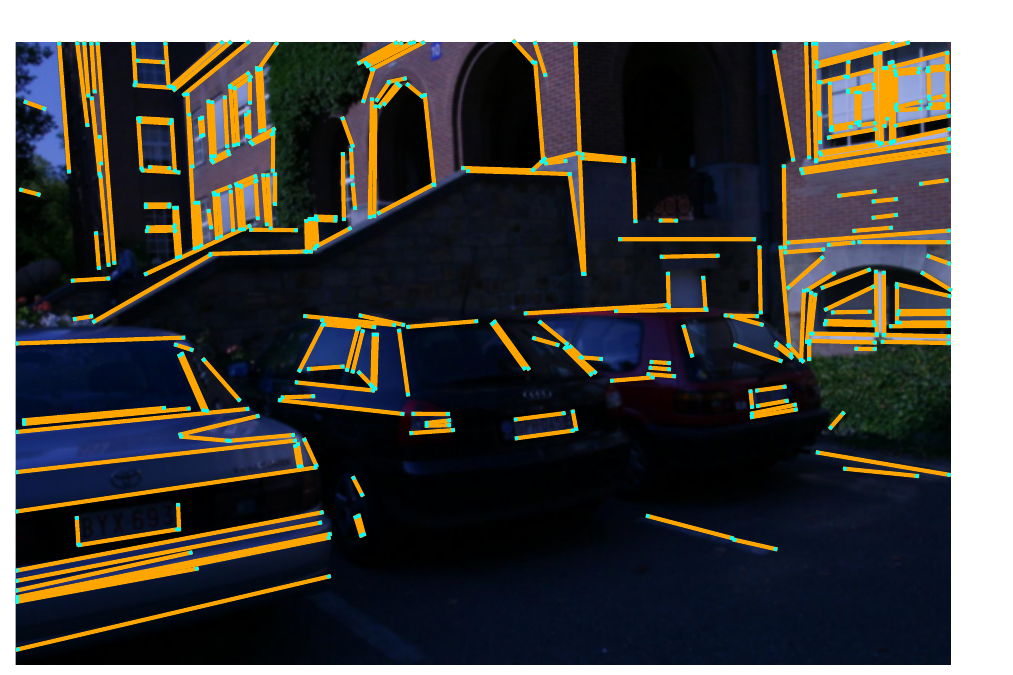} &
        \includegraphics[width=0.19\linewidth]{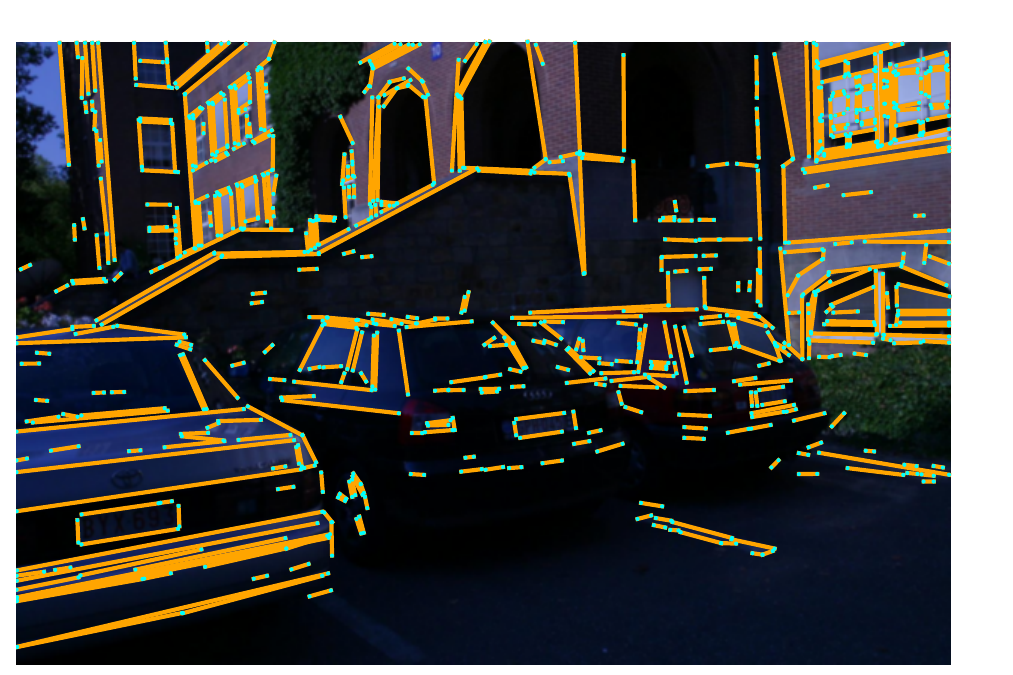}
        \\
        \includegraphics[width=0.19\linewidth]{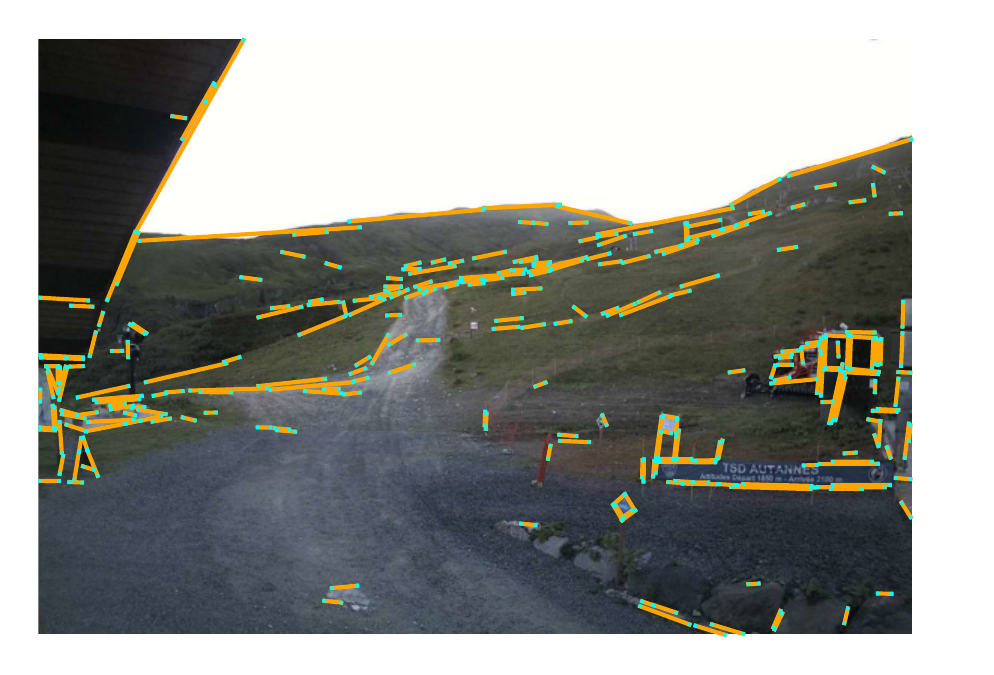} &
        \includegraphics[width=0.19\linewidth]{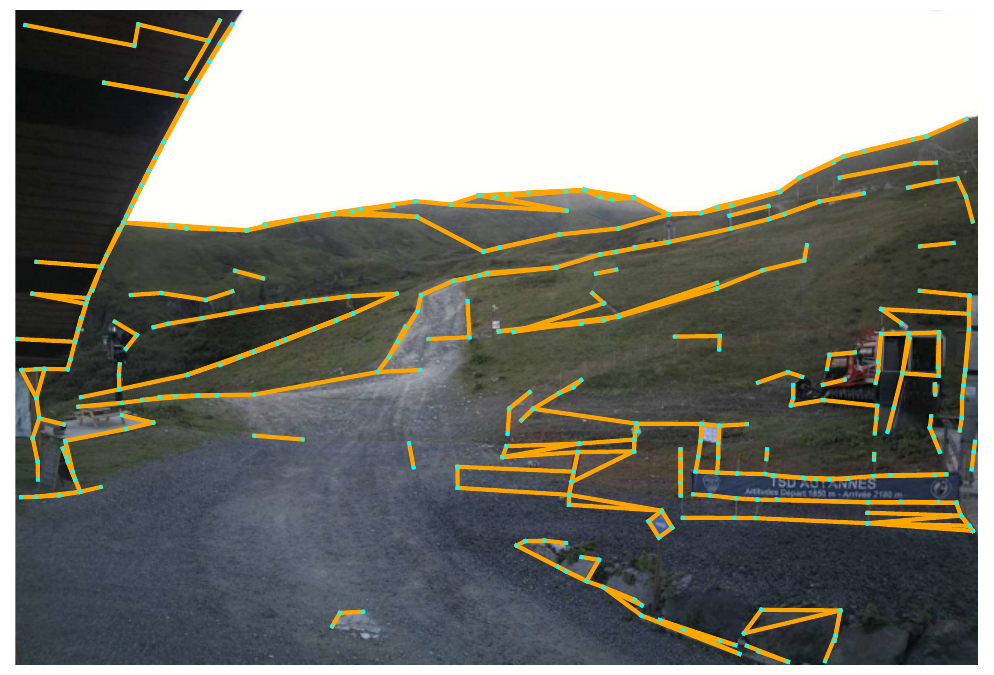} &
        \includegraphics[width=0.19\linewidth]{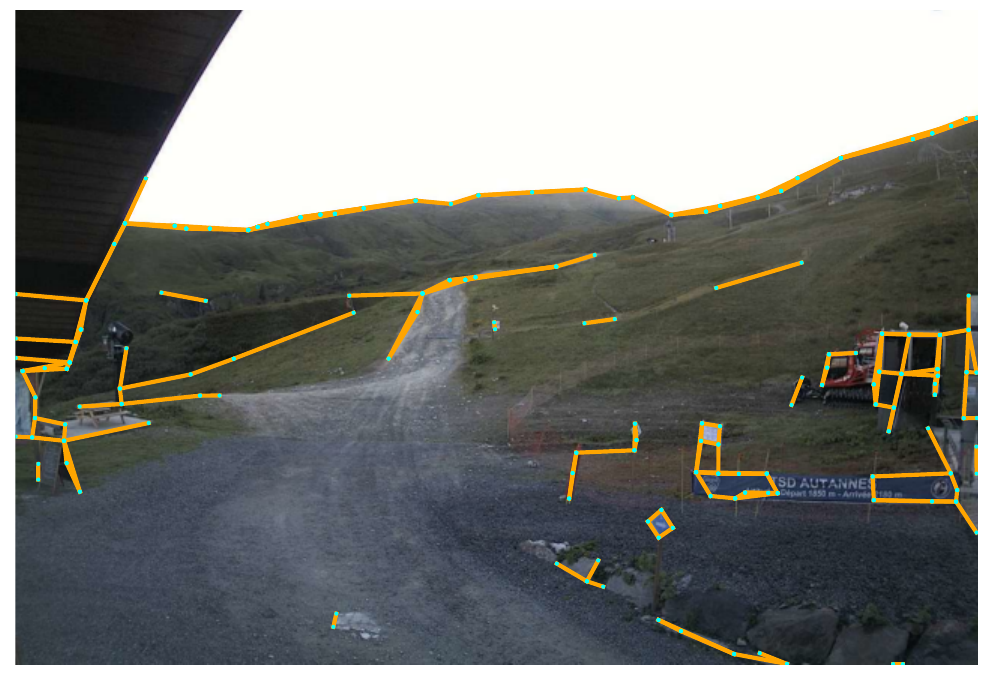} &
        \includegraphics[width=0.19\linewidth]{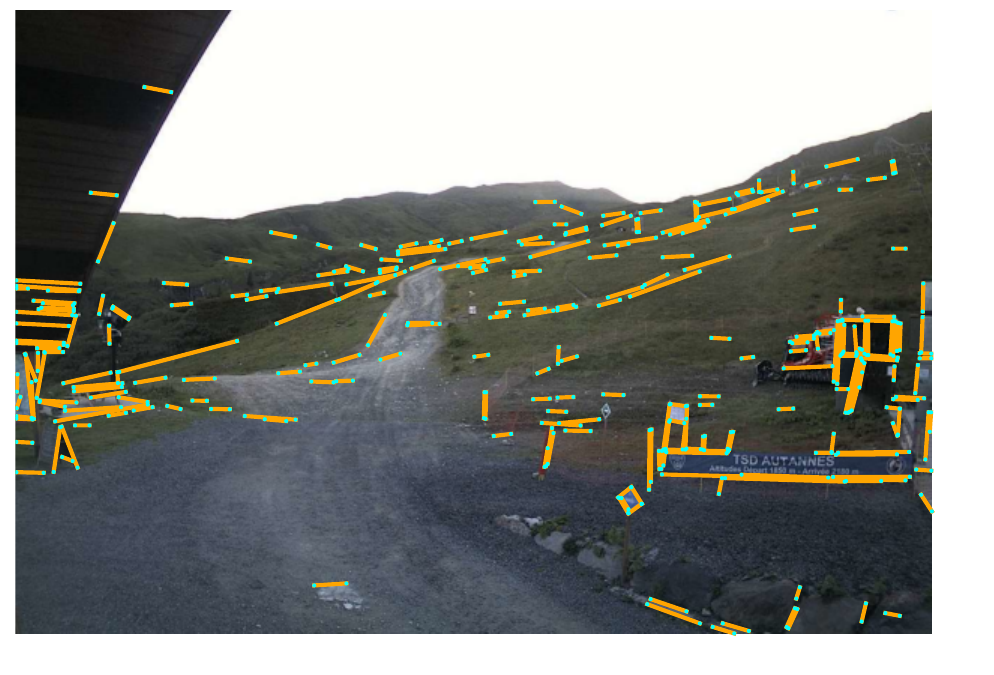} &
        \includegraphics[width=0.19\linewidth]{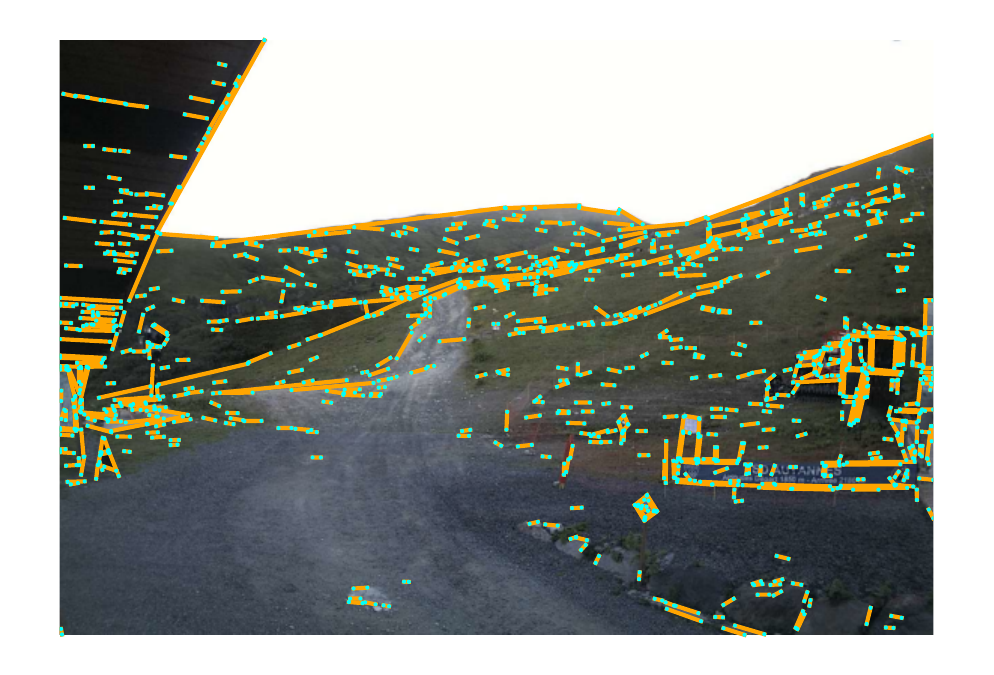}
        \\
        \includegraphics[width=0.19\linewidth]{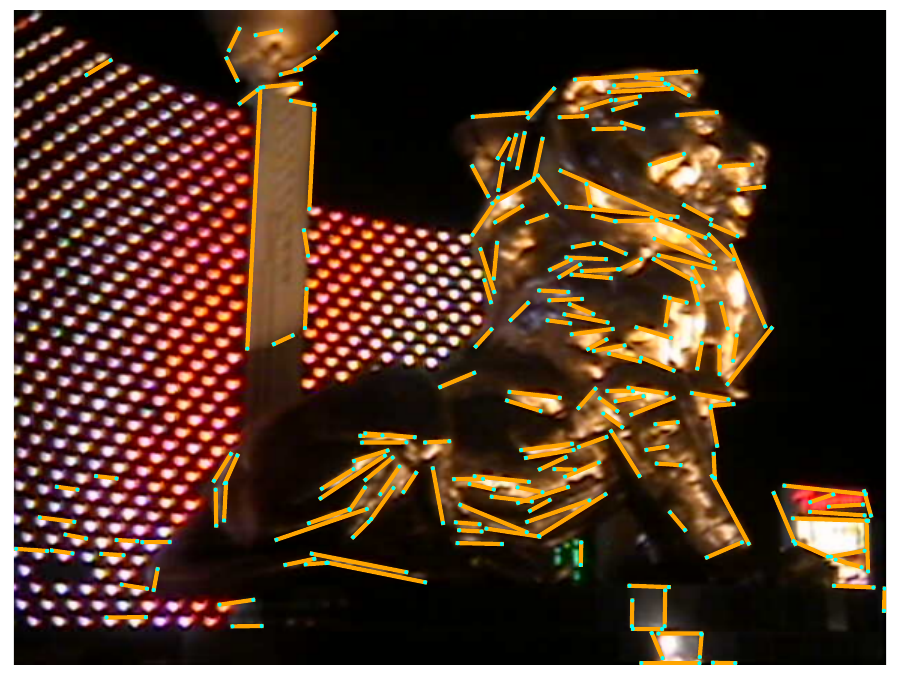} &
        \includegraphics[width=0.19\linewidth]{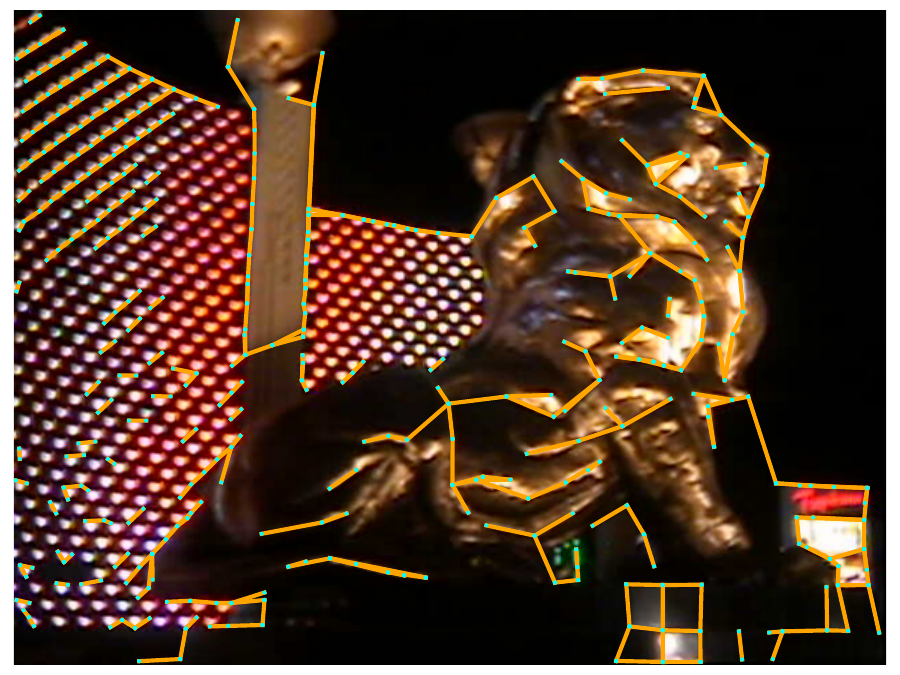} &
        \includegraphics[width=0.19\linewidth]{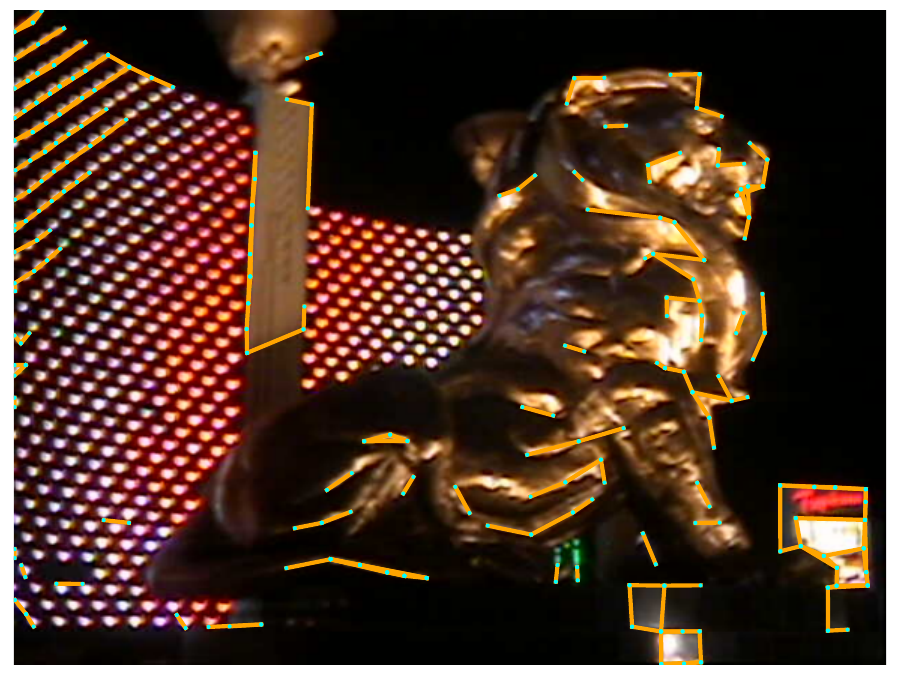} &
        \includegraphics[width=0.19\linewidth]{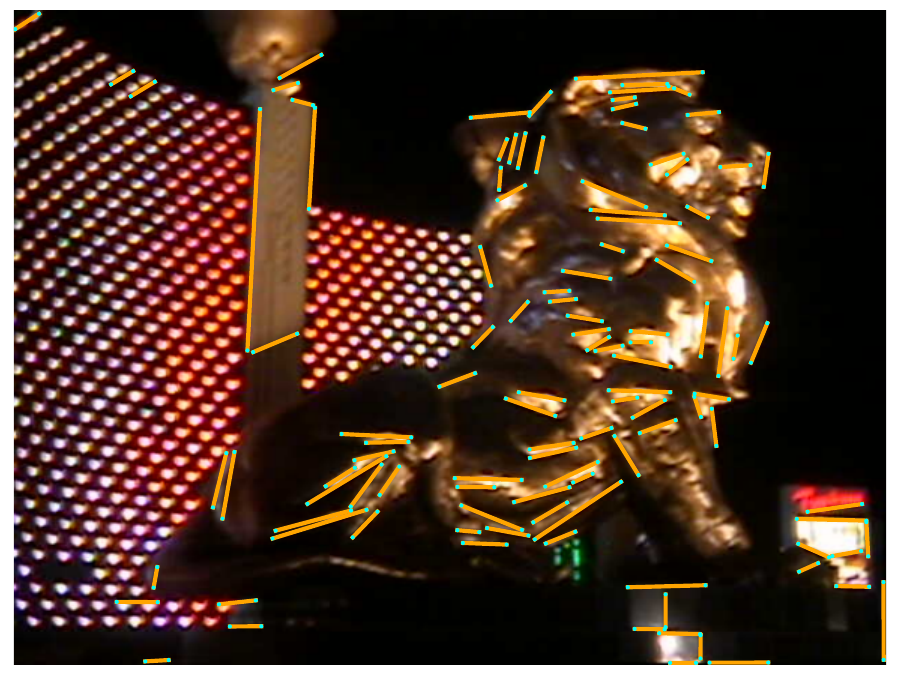} &
        \includegraphics[width=0.19\linewidth]{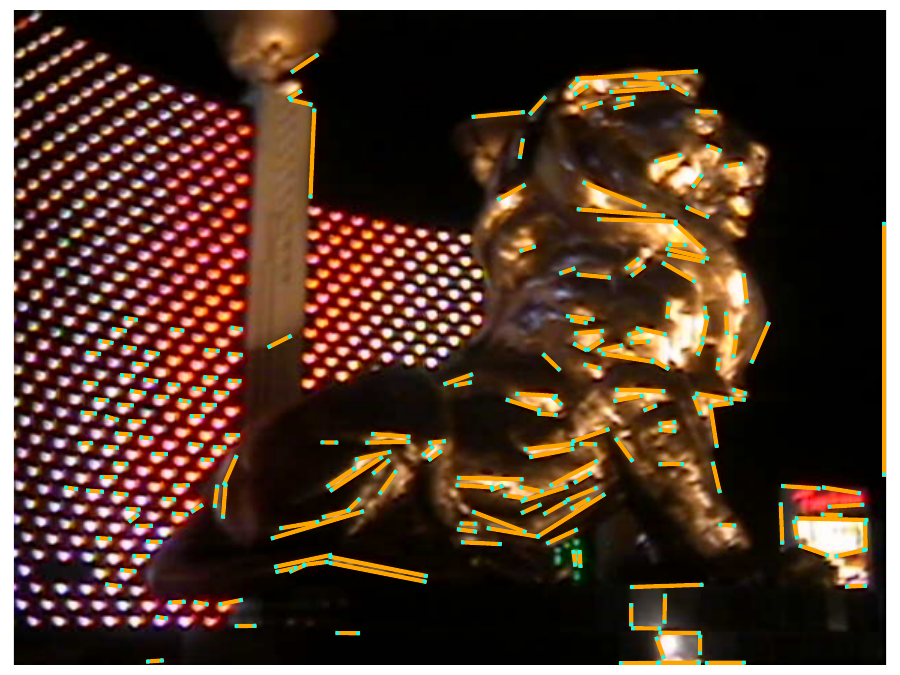}
    \end{tabular}
    \vspace{-0.2cm}
    \caption{\textbf{Line detection examples.} Unlike wireframe-based methods \cite{ScaleLSD,ferre2025wireframe} that focus solely on structural lines, \OURS\ produces more general and repeatable line detections that better support downstream tasks, while remaining significantly more lightweight. The last four images are challenging examples for \OURS\ and other detectors.}
    \label{fig:supp_example_detection}
\end{figure*}

%% file: figures/supp_matching_ex.tex
% Point and line matching qualitative examples (supplementary).
% Source: glue-factory JPL predictions on HPatches, top-300 matches drawn.
% Colour: yellow/green = correct, red = wrong (<3 px error under the GT
% homography). One combined figure per modality (success + failure), laid
% out as a 2-column grid; per-panel labels give the scene and the variation
% (viewpoint or illumination). The near-square stained-glass montages are
% shown at reduced width so the grids stay even.
%
% ============================ POINT MATCHING ==============================
\begin{figure*}[tbp]
    \centering
    \setlength{\tabcolsep}{2pt}
    \begin{tabular}{cc}
        \includegraphics[width=0.48\linewidth]{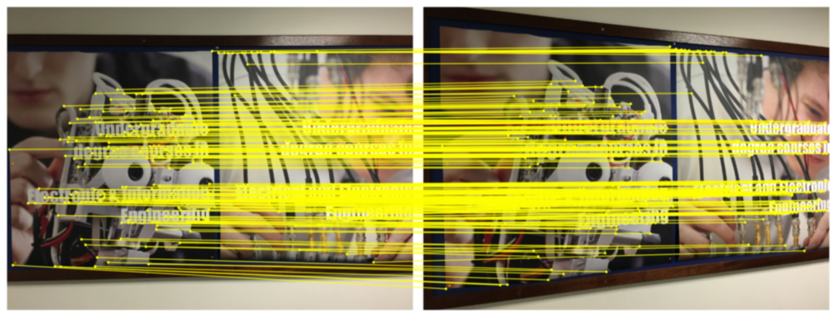} &
        \includegraphics[width=0.48\linewidth]{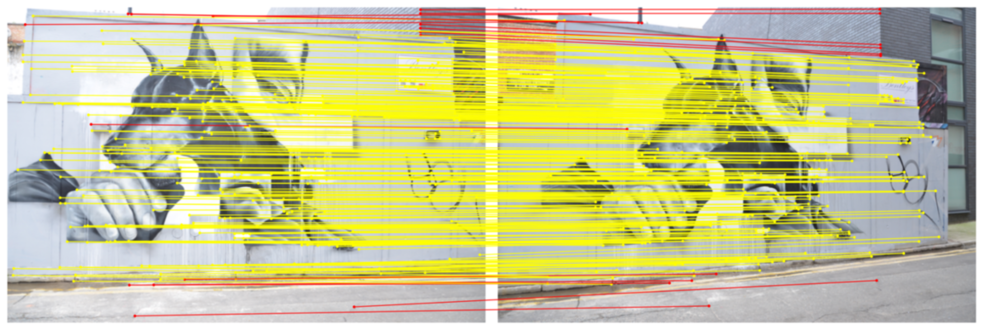} \\[-4pt]
        {\scriptsize artwork, viewpoint} & {\scriptsize sculpture, viewpoint} \\[3pt]
        \includegraphics[width=0.48\linewidth]{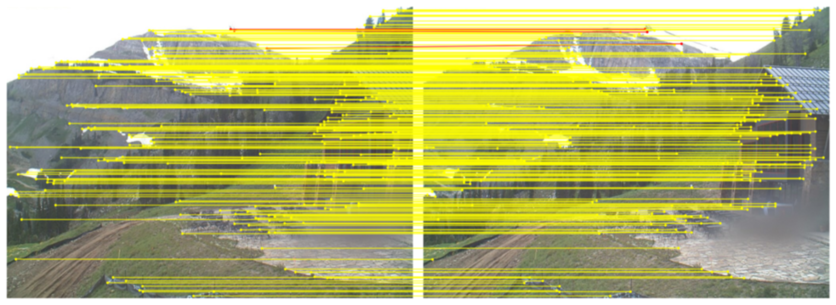} &
        \includegraphics[width=0.48\linewidth]{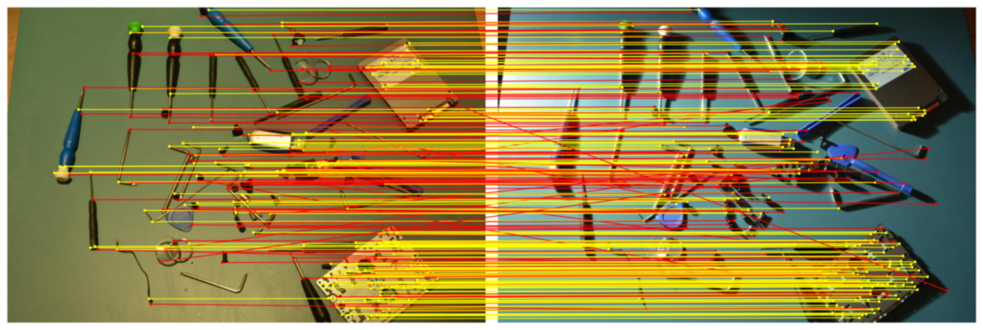} \\[-4pt]
        {\scriptsize resort, illumination} & {\scriptsize tools, illumination} \\[3pt]
        \multicolumn{2}{c}{\includegraphics[width=0.42\linewidth]{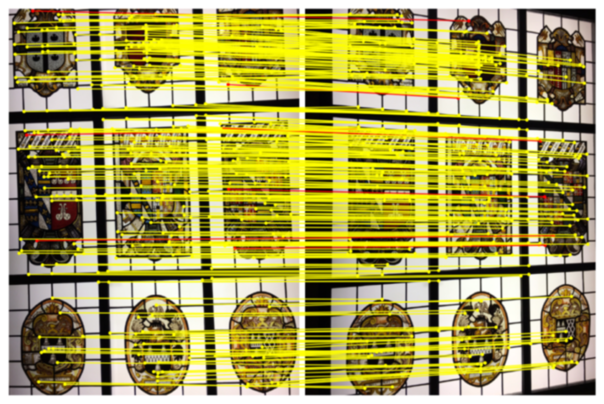}} \\[-4pt]
        \multicolumn{2}{c}{\scriptsize stained glass, viewpoint} \\[3pt]
        \includegraphics[width=0.48\linewidth]{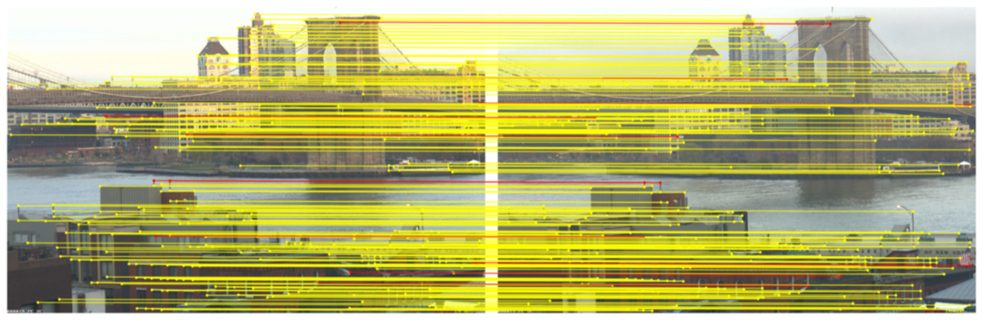} &
        \includegraphics[width=0.48\linewidth]{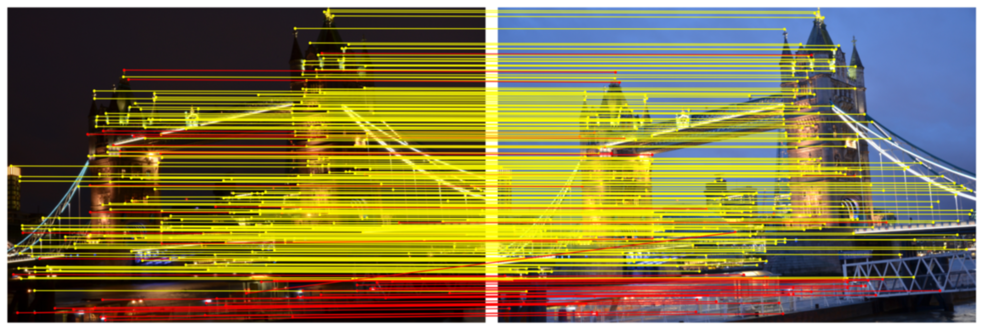} \\[-4pt]
        {\scriptsize brooklyn bridge, illumination} & {\scriptsize tower bridge, illumination} \\[3pt]
        \includegraphics[width=0.48\linewidth]{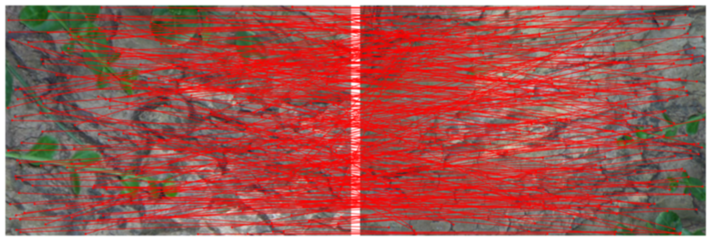} &
        \includegraphics[width=0.48\linewidth]{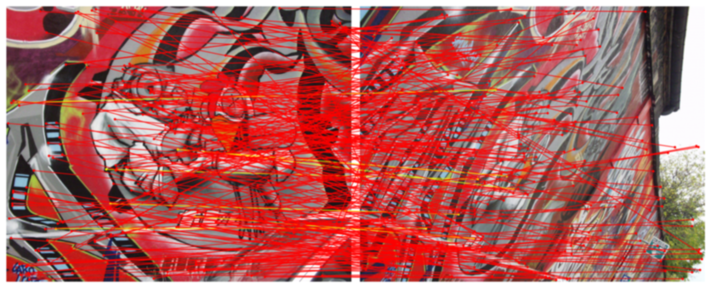} \\[-4pt]
        {\scriptsize bark, viewpoint} & {\scriptsize graffiti, viewpoint} \\
    \end{tabular}
    \caption{\textbf{Point matching on HPatches} (yellow $=$ correct, red $=$ wrong with respect to the ground-truth homography; only the most salient matches are shown). \OURS\ generally matches keypoints reliably under both viewpoint and illumination changes. On repetitive structures the salient corners still match, with errors confined to weaker detections, whereas specular surfaces and especially self-similar texture under large viewpoint changes are harder and yield many incorrect matches.}
    \label{fig:supp_point_matching}
\end{figure*}

% ============================ LINE MATCHING ===============================
\begin{figure*}[tbp]
    \centering
    \setlength{\tabcolsep}{2pt}
    \begin{tabular}{cc}
        \includegraphics[width=0.48\linewidth]{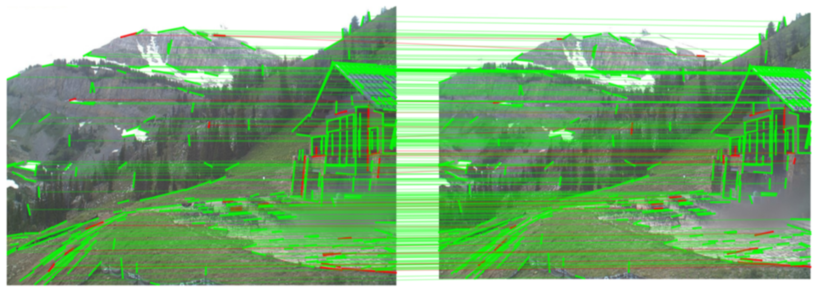} &
        \includegraphics[width=0.48\linewidth]{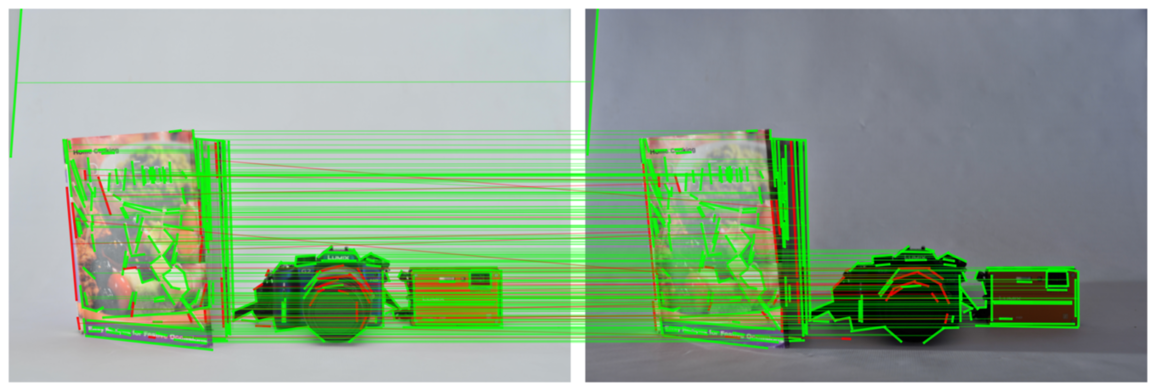} \\[-4pt]
        {\scriptsize resort, illumination} & {\scriptsize product, illumination} \\[3pt]
        \includegraphics[width=0.48\linewidth]{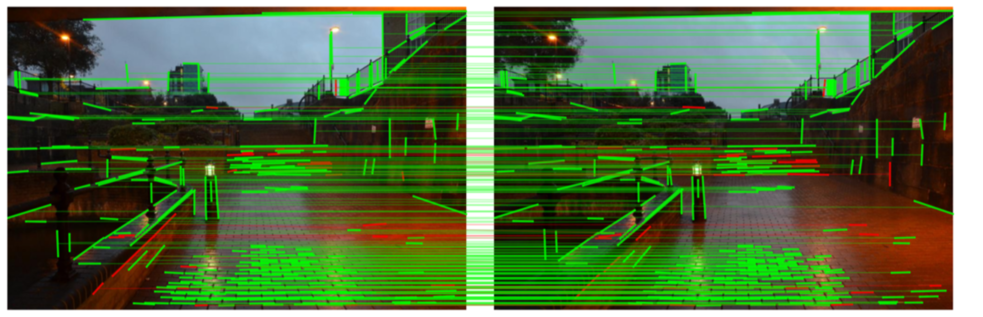} &
        \includegraphics[width=0.48\linewidth]{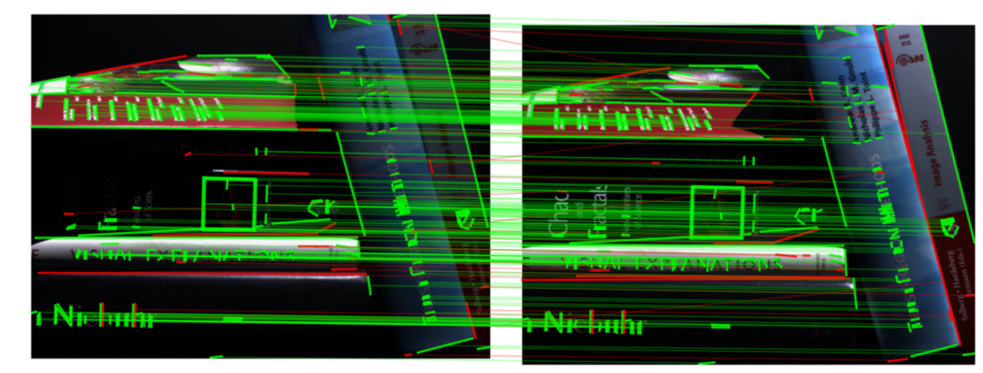} \\[-4pt]
        {\scriptsize stairs, illumination} & {\scriptsize books, illumination} \\[3pt]
        \includegraphics[width=0.48\linewidth]{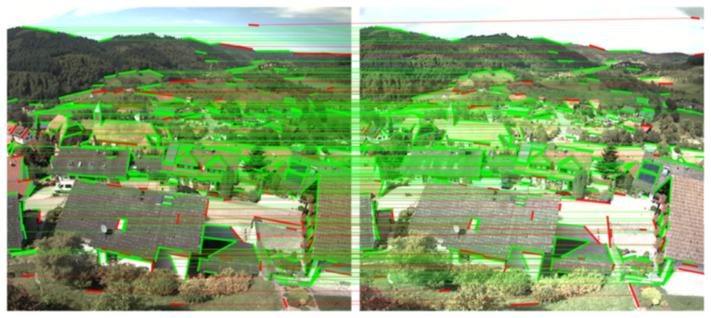} &
        \includegraphics[width=0.48\linewidth]{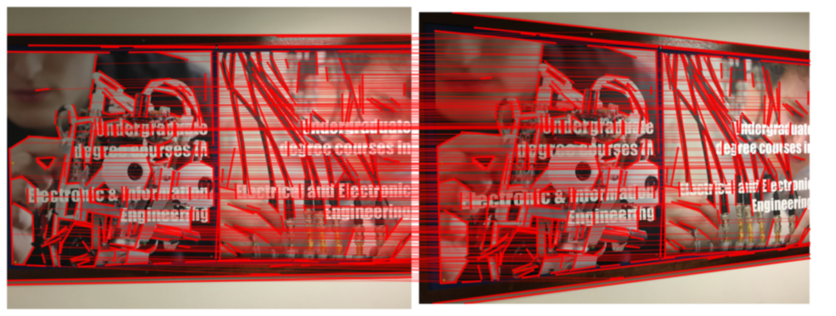} \\[-4pt]
        {\scriptsize village, illumination} & {\scriptsize artwork, viewpoint} \\[3pt]
        \includegraphics[width=0.48\linewidth]{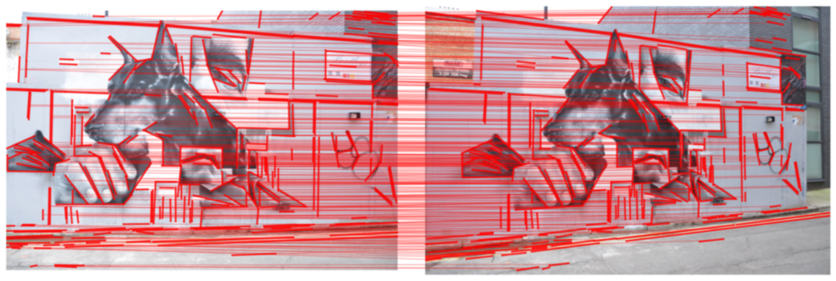} &
        \includegraphics[width=0.42\linewidth]{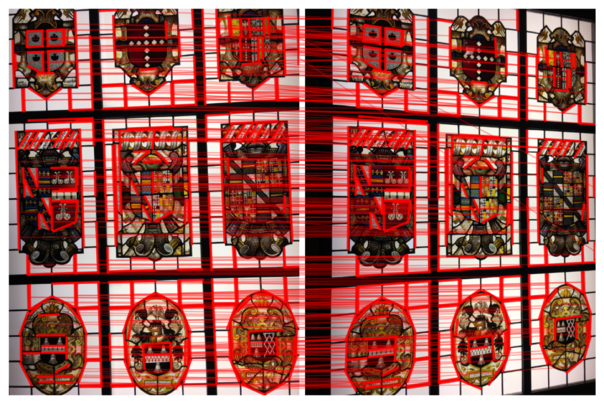} \\[-4pt]
        {\scriptsize sculpture, viewpoint} & {\scriptsize stained glass, viewpoint} \\
    \end{tabular}
    \caption{\textbf{Line matching on HPatches} (green $=$ correct, red $=$ wrong with respect to the ground-truth homography; only the longest matched segments are shown). \OURS\ matches lines well under illumination changes on structured man-made scenes, but its line branch is not viewpoint-robust: the same viewpoint pairs that are matched reliably by points are almost entirely mismatched as lines.}
    \label{fig:supp_line_matching}
\end{figure*}